\documentclass[lettersize,journal]{IEEEtran}
\usepackage{amsmath,amsfonts}
\usepackage{array}
\usepackage[caption=false,font=normalsize,labelfont=sf,textfont=sf]{subfig}
\usepackage{textcomp}
\usepackage{stfloats}
\usepackage{url}
\usepackage{verbatim}
\usepackage{graphicx}
\usepackage{cite}
\usepackage{graphicx}
\usepackage{amsmath,amssymb,amsfonts}
\usepackage{pifont}
\newcommand{\cmark}{\ding{51}}%
\newcommand{\xmark}{\ding{55}}%
\usepackage{xspace}
\usepackage{subfig}
\usepackage{tikz} %
\usetikzlibrary{svg.path}
\usetikzlibrary{petri,positioning,backgrounds}
\usepackage{threeparttable}
\usepackage{multirow} 
\usepackage{booktabs}
\usepackage{makecell}
\usepackage{algorithm2e}
\SetKwComment{Comment}{// }{}%

\RestyleAlgo{ruled}

\makeatletter
\DeclareRobustCommand\onedot{\futurelet\@let@token\@onedot}
\def\@onedot{\ifx\@let@token.\else.\null\fi\xspace}
\def\eg{\emph{e.g}\onedot}
\def\ie{\emph{i.e}\onedot}

\def\etc{\emph{etc}\onedot} 

\def\etal{\emph{et al}\onedot}
\makeatother
\begin{document}

\title{From Algorithm to Hardware: A Survey on Efficient and Safe Deployment of Deep Neural Networks}


\author{Xue Geng, Zhe Wang, Chunyun Chen, Qing Xu, Kaixin Xu, Chao Jin, Manas Gupta, \\ Xulei Yang, Zhenghua Chen, Mohamed M. Sabry Aly, Jie Lin, Min Wu, Xiaoli Li~\IEEEmembership{Fellow, IEEE}
\thanks{Xue Geng, Zhe Wang, Qing Xu, Kaixin Xu, Chao Jin, Manas Gupta, Xulei Yang, Zhenghua Chen, Min Wu and Xiaoli Li are with the Institute for Infocomm Research, A*STAR, Singapore. (email: geng\_xue, wang\_zhe, xu\_qing, Xu\_Kaixin, jin\_chao, manas\_gupta, yang\_xulei, chen\_zhenghua, wumin, xlli@i2r.a-star.edu.sg).}
\thanks{Chunyun Chen and Mohamed M. Sabry Aly are with the Nanyang Technological University (email: chunyun001@e.ntu.edu.sg, msabry@ntu.edu.sg).}
\thanks{Dr. Jie Lin, Dr. Min Wu and Dr. Xiaoli Li are the corresponding authors.}}

\markboth{Journal of \LaTeX\ Class Files,~Vol.~14, No.~8, August~2021}%
{Shell \MakeLowercase{\textit{et al.}}: A Sample Article Using IEEEtran.cls for IEEE Journals}

\IEEEpubid{0000--0000/00\$00.00~\copyright~2021 IEEE}

\maketitle

\begin{abstract}
Deep neural networks (DNNs) have been widely used in many artificial intelligence (AI) tasks. However, deploying them 
brings significant challenges due to 
the huge cost of memory, energy, and computation. 
To address these challenges, researchers have developed various model compression techniques such as model quantization and model pruning. 
Recently, there has been a surge in research of 
compression methods 
to achieve model efficiency while retaining the performance. 
Furthermore, more and more works focus on customizing the DNN hardware accelerators to better leverage the model compression techniques. 
In addition to efficiency, preserving security and privacy is critical for deploying DNNs.
However, the vast and diverse body of related works can be overwhelming.
This inspires us to conduct a comprehensive survey on recent research toward the goal of high-performance, cost-efficient, and safe deployment of DNNs.  
Our survey first covers the mainstream model compression techniques such as model quantization, model pruning, knowledge distillation, and optimizations of non-linear operations. We then introduce recent advances in designing hardware accelerators that can adapt to efficient model compression approaches. Additionally, we discuss how homomorphic encryption can be integrated to secure DNN deployment. Finally, we discuss several issues, such as hardware evaluation, generalization, and integration of various compression approaches. Overall, we aim to provide a big picture of efficient DNNs, from algorithm to hardware accelerators and security perspectives. 
\end{abstract}


\section{Introduction}
\label{sec:introduction}
\IEEEPARstart{D}{eep} neural networks (DNNs) are currently the foundation of many modern artificial intelligence (AI) applications with superior performance.  However, with millions or even billions of parameters \cite{chung2022scaling}, 
deploying DNNs to devices presents fundamental challenges due to high costs such as computational and energy costs. 
In particular, in addition to the performance accuracy~\cite{sze2017efficient}, it is also essential to consider 
the latency or throughput, which determines if the DNN can run fast in real-time. Besides, we need to 
take the following hardware-related factors into account: 1) energy and power, which determines the actual running cost; 2) area, which determines the actual design cost; and 3) storage, which determines if the DNN can be stored, or run with intermediate data. Finally, when deploying DNNs, privacy is critical to ensure secure deployment. 
To deploy DNNs successfully, various computing architectures have been proposed to meet the above criteria. 
Different computing architectures might need to address various challenges when deploying DNNs. For example, cloud computing aims to centralize the computing resources and lets the cloud not directly interfere with the end user. However, it faces the latency issue and the tremendous cost of maintaining large-scale cloud centers. Meanwhile, edge computing helps reduce the latency by running DNNs at the edge (IoT devices) with direct interference from the user. It has been widely used in many areas, including autonomous driving, smart agriculture, surveillance, etc. \cite{varghese2016challenges}. However, edge devices provide a limited resource for computation. Model compression techniques have been widely explored to meet the above criteria among various computing architectures.

\begin{table*}[tbp]
\centering
\scriptsize
\caption{summary of this article}
\vspace{-5pt}
\label{tab:summary}
\begin{tabular}{l|cc|l}
\hline
Preliminaries                           & \multicolumn{2}{c|}{Brief Descriptions of Efficient and Safe DNNs}                                                                                                                                      & \multicolumn{1}{l}{Sections} \\ \hline
\multirow{16}{*}{Model compression}     & \multicolumn{1}{c|}{\multirow{5}{*}{Quantization and Entropy Encoding}}                 
& Uniform and Non-uniform Quantization                                                                                      &        Section \ref{sec:uniform_nonuniform}                      \\ \cline{3-4}
& \multicolumn{1}{c|}{}    
                & Mixed-precision Quantization                                                                                              &  Section \ref{sec:mixed_precision}                            \\ \cline{3-4}
& \multicolumn{1}{c|}{}                                              & Transform Quantization                                                                                                    &  Section \ref{sec:transform_quantization}                            \\ \cline{3-4}
& \multicolumn{1}{c|}{}                                              & Quantization for Transformers                                                                                                     &  Section \ref{sec:quant_vi}                            \\ \cline{3-4}
& \multicolumn{1}{c|}{}                                              & Entropy Encoding                                                                                                          &     Section \ref{sec:entropy_encoding}                         \\ \cline{2-4} 
                                        & \multicolumn{1}{c|}{\multirow{2}{*}{Network Pruning}}                       
                                        &  Pruning Strategies - Before, During, and After Training                                                                                                         &     Section \ref{subsec:pruning_schemes}                         \\ \cline{3-4} 
                                        & \multicolumn{1}{c|}{} &
                                        Pruning Paradigms - Unstructured, Structured, and Semi-Structured Pruning                                                                                                      &  Section \ref{subsec:pruning_paradigms}                            \\ \cline{2-4}
                                        & \multicolumn{1}{c|}{\multirow{2}{*}{Knowledge Distillation}}       
                                        & Structured Knowledge       &    Section \ref{sec:structured_knowledge}                          \\ \cline{3-4}
                                        & \multicolumn{1}{c|}{}                                              & Distillation Schemes                                                                                                      &    Section \ref{subsec:distillation_schemes}      \\ \cline{2-4} 
                                        & \multicolumn{1}{c|}{\multirow{3}{*}{Non-linear Operations}}        & Non-maximum Suppression                                                                                                                       & Section \ref{sec:nms}                             \\ \cline{3-4}
                                        & \multicolumn{1}{c|}{}                                              & Softmax                                                                                                                   &   Section \ref{sec:softmax}                           \\ \cline{3-4}
                                        & \multicolumn{1}{c|}{}                                              & Activation functions                                                                                                      &       Section \ref{sec:activation_functions} \\ \cline{2-4}
                                        & \multicolumn{1}{c|}{\multirow{2}{*}{NAS \& TinyML}}
                                        & NAS & Section~\ref{subsec:nas}\\ \cline{3-4}
                                        & \multicolumn{1}{c|}{} & TinyML & Section~\ref{subsec:tinyml}
                                        \\ \hline
\multirow{7}{*}{Hardware Optimization}  & \multicolumn{1}{c|}{\multirow{4}{*}{Accelerators on Linear Operations}}            & Hardware Accelerating on CNNs and Transformers                                & Section \ref{sec:cnns_transformers}         \\ \cline{3-4}
                                        & \multicolumn{1}{c|}{}                                              & \begin{tabular}[c]{@{}c@{}}Hardware Accelerating after Quantization: \\ Mixed-Precision and Entropy Encoding\end{tabular} & Section \ref{sec:accelerator_quantization}         \\ \cline{3-4}
                                        & \multicolumn{1}{c|}{}                                              & Hardware Accelerating after Pruning:  Sparse Architecture              & Section \ref{sec:accelerator_pruning}         \\ \cline{2-4} 
                                        & \multicolumn{1}{c|}{\multirow{2}{*}{Accelerators on Non-linear Operations}}        & Non-maximum Suppression                                                                                                                       & Section \ref{sec:accelerator_nms}         \\ \cline{3-4}
                                        & \multicolumn{1}{c|}{}                                              & Softmax                                                                                                                   & Section \ref{sec:hardware_softmax}         \\ \cline{2-4}
                                        & \multicolumn{1}{c|}{Stochastic Computing Architecture }
                                        & Stochastic Computing Architecture                                                                                                                       & Section \ref{subsec:stochastic_architecture}         \\ \cline{1-4}
\multirow{3}{*}{Homomorphic Encryption} & \multicolumn{1}{c|}{\multirow{3}{*}{Homomorphic Neural Networks}} & Homomorphic Neural Networks                                                                                    & Section \ref{sec:ahe_hnn}        \\ \cline{3-4}
& \multicolumn{1}{c|}{}                                              & Compression on Homomorphic Neural Networks                                                                                                     & Section \ref{sec:ahe_compression}  \\ \cline{3-4}

& \multicolumn{1}{c|}{}                                              & Hardware on Homomorphic Neural Networks                                                                                                     & Section \ref{sec:ahe_accelerator}         \\ \hline
\end{tabular}
\vspace{-10pt}
\end{table*}

Model compression aims at the compact model to reduce the hardware cost. A smaller model requires fewer arithmetic operations and thus improves the inference speed, requires less memory bandwidth to fetch data and thus saves energy consumption, and requires less on-chip memory to store and, therefore, is less expensive. In particular, we review the mainstream approaches to compressing the DNNs, including model quantization, network pruning, and knowledge distillation, and the optimization of non-linear operations such as softmax and non-maximum suppression (NMS) as they play an essential role in deploying DNNs. 

In addition, effectively deploying a compressed model poses a challenge as current hardware accelerators are optimized for uncompressed DNNs, leading to significant wastage of computation cycles and memory bandwidth when running on compressed DNN models~\cite{deng2020model}.
Therefore, we need to co-design the algorithm and the hardware. This paper also presents an overview of the custom hardware accelerators, \eg, Field Programmable Gate Arrays (FPGAs) and Application-Specific Integrated Circuits (ASICs), which are designed for the compressed DNNs in terms of various network operations. 
\IEEEpubidadjcol
In addition to efficient computation, privacy is crucial in maintaining the integrity, reliability, and availability of third-party computing, particularly in cloud computing. In the past, encryption of confidential information was the conventional approach before using the cloud model. While it may safeguard user data privacy from an untrustworthy third party, it cannot facilitate effective ciphertext computing. Conventional cryptosystems effectively protect stored data and data in transit, but they fail to secure the data while it is decrypted for processing.
Therefore, many researchers have been exploring Fully Homomorphic Encryption (FHE). FHE is a valuable capability in distributed computation and heterogeneous networking. In this survey, we present a comprehensive review of the latest FHE techniques for DNNs and how model compression helps in these scenarios.

Few surveys exist on deep neural compression\cite{deng2020model, mishra2023transforming}. However, these works mainly focus on existing models' compression and acceleration algorithms. For example, \cite{cheng2018model, mishra2020survey, neill2020overview, alqahtani2021literature} mainly focused on compression algorithms of existing models, including quantization and knowledge distillation. In addition to the compression algorithms, \cite{berthelier2021deep} also discussed neural architecture optimization.   \cite{deng2020model} also discussed the hardware accelerators, especially on mixed-precision data and sparse architectures. In addition to these topics, our survey also discusses non-linear operation accelerators. \cite{mittal2021survey} mainly focused on hardware accelerators for Recurrent Neural Networks (RNNs). \cite{nan2019deep} explored the performances of various model compression techniques on the mobile platform.  \cite{xu2022survey} surveyed many works in model compression and acceleration for pre-trained language models.  \cite{gupta2022compression} discussed the compression algorithms for NLP tasks.  \cite{mishra2023transforming} talked about the model compression techniques for IoT applications. Besides, some surveys exist on specific network compression techniques such as quantization \cite{guo2018survey, gholami2021survey} or network pruning \cite{xu2020convolutional, liu2020pruning, reed1993pruning}, architecture search \cite{elsken2019neural, ren2021comprehensive, wistuba2019survey}, knowledge distillation \cite{gou2021knowledge, wang2021knowledge}. 

In summary, while some surveys focus on specific aspects of deep neural compression, there is no survey as our article on model compression from the perspective of algorithm-efficient, hardware-accelerating, and deployment-securing DNNs. These three aspects are critical for many real-world applications that require small model sizes, hardware acceleration, and secure environments to save memory, energy, and computation costs. Our contributions are summarised as follows:
\begin{itemize}
    \item We present a comprehensive overview on neural network compression techniques. Efficient compression algorithms are essential for reducing memory and storage footprint of deep learning models.
    They enable faster, more energy-efficient inference, reduce bandwidth requirements, and offer adaptability to different deployment scenarios by facilitating efficient fine-tuning, thus optimizing model deployment.
    \item We also compile the literature on accelerating DNNs considering the hardware constraints. Hardware-aware accelerators play a crucial role in optimizing the use of specialized hardware for model inference, ensuring efficient hardware utilization, scalability, energy efficiency, and low latency. 
    \item To secure computations on encrypted data, we review Homomorphic Encryption (HE) for applications like Homomorphic Neural Networks (HNNs) in privacy-sensitive domains. In particular, we focus on model compression and hardware accelerators, specifically for HE.
    \item We summarize the challenges for compressing, accelerating and securing DNNs and discuss several future directions in the promising field.
\end{itemize}

The organization of this article is shown in Table \ref{tab:summary} and is summarized as follows. In Section \ref{sec:compression}, we describe the model compression techniques, including quantization, pruning, knowledge distillation, and optimization on non-linear operations. Section \ref{sec:compression_hardware} discusses how the hardware accelerates the compressed model. Section \ref{sec:compression_ahe} presents an overview of the privacy-preserving technique with Homomorphic Encryption (HE). Finally, Section \ref{sec:challenges} presents the challenges and opportunities of model compression and hardware-software co-design in real-world applications, before concluding the paper.

\section{Neural Network Compression}
\label{sec:compression}
\noindent We briefly introduce the general deep neural networks (DNNs) in Appendix A and then we introduce advanced neural network compression techniques. We classify existing approaches into four main categories:
\begin{itemize}
\item Quantization and Entropy Encoding: Quantization attempts to reduce the bit width of the data 
in a DNN, aiming at reducing the model size for memory saving and simplifying the operations for accelerating computation.
\item Network Pruning: 
This category of methods aims to explore redundancy in the model parameters, eliminating non-critical and redundant parameters. This results in a more efficient model.
\item Knowledge Distillation (KD): The KD methods learn a distilled model by training a more compact neural network to reproduce the output of a larger network.

\item Hardware-friendly Non-linear Operations: 
This category of methods aims to design hardware-friendly procedures (such as linear functions) to substitute non-linear operations, shortening the hardware design cycle and speeding up inference, which is better suited for deployment on hardware platforms.

\end{itemize}






\subsection{Quantization and Entropy Coding}
\label{subsec:quantization}
Benefiting from discretizing the parameters, quantized DNNs have 
reduced memory cost and lower computation, as discretized parameters, after quantization, can be stored and calculated more efficiently.
Moreover, 
by taking advantage of the peaky distribution of discretized values, entropy coding can compress the size of quantized parameters, leading higher compression ratio~\cite{han16iclr}.
This subsection will review the main quantization and entropy coding approaches. In particular,
first, we discuss the quantization approaches built by uniform quantizer and non-uniform quantizer.
Second, we introduce specific quantization schemes developed for neural network compression, including mixed-precision and transform quantization.
Finally, we discuss the entropy coding techniques. 

\subsubsection{Uniform Quantizer and Non-Uniform Quantizer}
\label{sec:uniform_nonuniform}
\begin{figure}[tbp]
\centering
\includegraphics[width=0.45\textwidth]{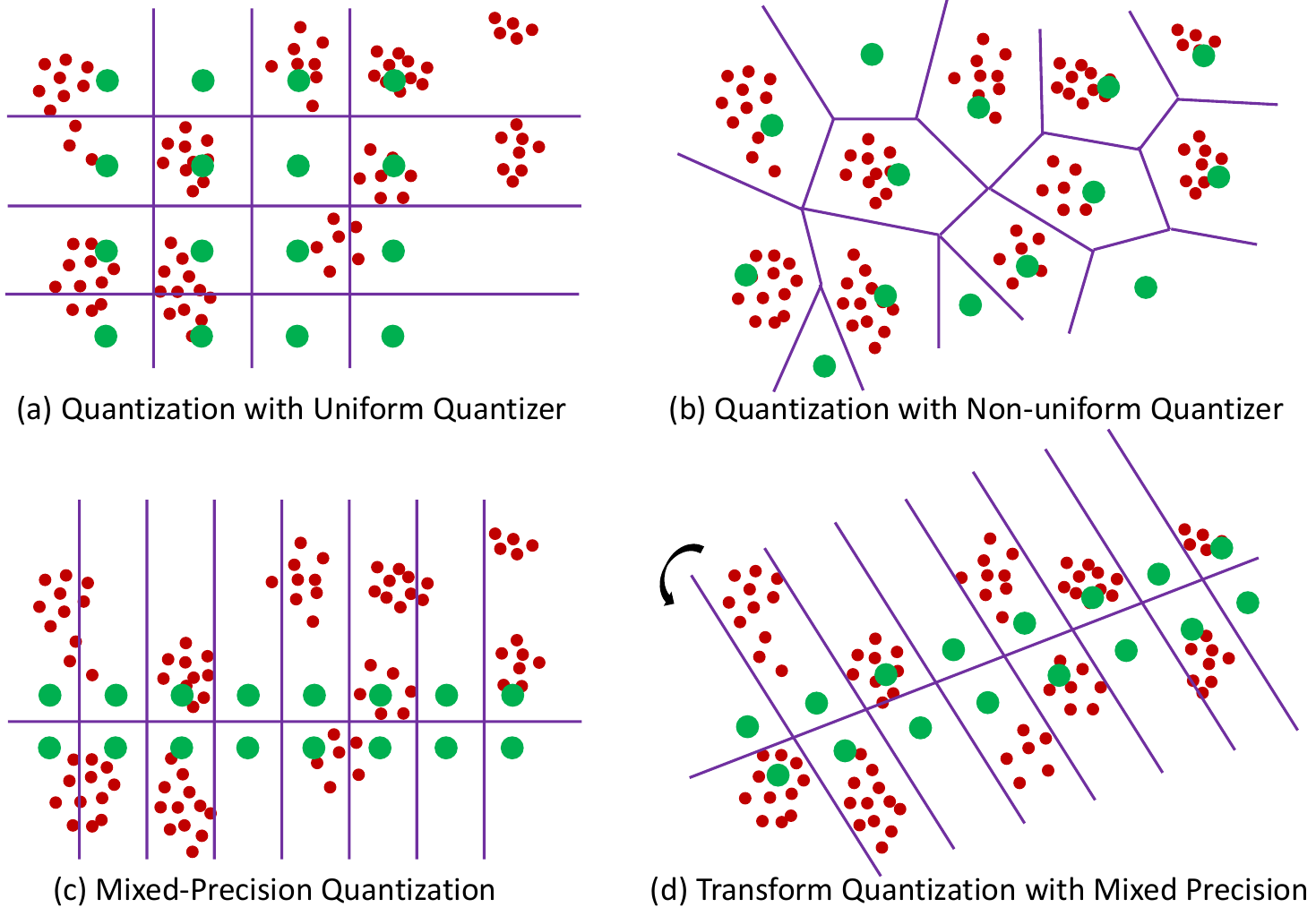}
\caption{(a) Quantization with uniform quantizer: the space is divided into equidistant quantization centroids. (b) Quantization with non-uniform quantizer: the space can be divided into polygonal cells based on data distribution. (c) Mixed-precision quantization: different dimensions (layers, channels, or kernels in the case of neural network quantization) are quantized with different bit widths. (d) Transform quantization with mixed precision: the space is first rotated and then quantized with mixed precision.}
\label{fig:quant}
\vspace{-10pt}
\end{figure}
Quantization approaches are ususally built based on uniform quantizers and non-uniform quantizers.
A uniform quantizer discretizes parameters onto equidistant quantization values as shown in Fig.~\ref{fig:quant}(a), while a non-uniform quantizer maps parameters onto irregular values decided by the distribution as shown in Fig.~\ref{fig:quant}(b).
A non-uniform quantizer can reduce quantization error by offering greater flexibility in selecting quantization values. Nevertheless, the need for look-up tables to store discretized quantization values introduces additional implementation overheads.

Uniform quantizers are commonly employed in various studies due to their hardware-friendly nature. For instance, LQ-Nets \cite{Zhang18eccv} proposed the joint training of a quantized DNN and its associated quantizers, ensuring compatibility with bit operations. INQ \cite{inq} focused on converting pre-trained neural networks into low-precision versions, restricting weights to power of two or zero. DOREFA-NET \cite{dorefa} suggested training neural networks with low bit width weights, activations, and gradients. Additionally, numerous works utilize uniform quantizers to quantize DNNs to very low bit widths (1-bit or 2-bits). 
For example, Li \textit{et al.} \cite{Li16nips} introduced Ternary Weight Networks (TWNs) with weights constrained to 1, 0, and -1 (2-bits). Additional relevant literature on uniform quantizers can be explored in Appendix B.

On the other hand, quantization with a non-uniform quantizer involves k-means or hashing tricks and may additionally involve the Discrete Cosine Transform (DCT) and residual quantization.
In particular, Gong \textit{et al.} \cite{pq} investigated a vector quantization method to compress the weights by applying k-means clustering and conducting product quantization.
HashNets \cite{hash} used a low-cost hash function to randomly group connection weights into hash buckets, and all links within the same hash bucket are quantized into a single parameter value.
Chen \textit{et al.} \cite{frequency} proposed to convert weights to the frequency domain with a DCT and used hash functions to group frequency parameters into hash buckets, where parameters assigned to the same hash bucket share a single quantization value.
Stock \textit{et al.} \cite{revisit} introduced a vector quantization method that aims at preserving the quality of the reconstruction of the network outputs rather than its weights, with only a set of unlabelled data needed at quantization time.
APoT \cite{APoT} proposed an efficient nonuniform quantization scheme for the bell-shaped and long-tailed distribution of weights and activations in neural networks. 
SYQ \cite{Faraone2018arxiv} introduced a quantization method to reduce information loss by learning a symmetric codebook for particular weight subgroups.
Table I in Appendix B provides a comprehensive summary and comparative analysis of various quantization methods, including those utilizing both uniform and non-uniform quantizers.

\subsubsection{Mixed-Precision Quantization}
\label{sec:mixed_precision}
Mixed-precision quantization as shown in Fig.~\ref{fig:quant}(c) uses 
distinct bit widths to quantize weights and activations of different layers, channels, or kernels.
Because weights and activation in various layers, channels, or kernels 
exhibit different levels of sensitivity to quantization, 
and thus allocating different bit widths 
is more reasonable and 
can deliver higher accuracy than uniform quantization.
Mixed-precision quantization can be categorized as layer-wise, channel-wise, and kernel-wise approaches based on the minimal elements 
used to determine the bit width.

ReLeQ~\cite{releq} introduced an end-to-end deep reinforcement learning (RL) framework to automate the process of discovering quantization bit width.
Alternatively, HAQ~\cite{wang19cvpr} leveraged RL to determine quantization bit width but employed a hardware simulator to generate direct feedback signals to the RL agent.
DNAS~\cite{dnas} presented a differentiable neural architecture search framework to explore the hyper-parameter space of quantization bit width.
Differentiable Quantization (DQ)~\cite{dq} learned quantizer parameters, such as step size and range, through training with straight-through gradients and then inferred quantization bit width based on the learned step size and range.
Hessian AWare Quantization (HAWQ)~\cite{HAWQ} introduced a second-order quantization method to select the quantization bit width for each layer based on the Hessian spectrum.
All the above methods deploy a layer-wise bit allocation scheme.
\begin{table*}[!htbp]
\centering
\scriptsize
\begin{threeparttable}
\caption{A summary of mixed-precision quantization approaches for the results on ImageNet.
}
\vspace{-5pt}
\begin{tabular}{l | c  | c |  c |  c |  c }
\hline
Methods & Bit Allocation Scheme & Model & Weight & Activation & Accuracy (Original) \\
\hline
\multirow{2}{*}{ReLeQ \cite{releq}} & \multirow{2}{*}{Layer-wise} & AlexNet &  5 Bits MP & 5 Bits MP & --- (---) \tnote{a} \\ 
& & MobileNet-V2 &  6 Bits MP & 6 Bits MP & --- (---)  \tnote{a} \\ 
\hline
\multirow{2}{*}{HAQ \cite{wang19cvpr}} & \multirow{2}{*}{Layer-wise} & MobileNet-V2 &  4 Bits MP & 4 Bits MP & 67.0\% (71.8\%) \\
& & ResNet-50 &  2 Bits MP & 32 Bits & 70.6\% (76.2\%) \\
\hline
DNAS \cite{dnas} & Layer-wise & ResNet-18 &  1.5 Bits MP & 32 Bits & 69.6\% (71.0\%) \\
\hline
\multirow{2}{*}{DQ \cite{dq}} & \multirow{2}{*}{Layer-wise} & MobileNet-V2 &  4 Bits MP & 4 Bits MP & 70.6\% (70.2\%) \\
& & ResNet-18 &  4 Bits MP & 4 Bits MP & 70.7\% (70.3\%) \\
\hline
\multirow{3}{*}{HAWQ \cite{HAWQ}} & \multirow{3}{*}{Layer-wise} & GoogleNet &  2 Bits MP & 4 Bits MP & 75.5\% (77.5\%) \\
 & & ResNet-50 &  2 Bits MP & 4 Bits MP & 75.5\% (77.4\%) \\
 & & SqueezeNet &  3 Bits MP & 8 Bits MP & 68.0\% (69.4\%) \\
\hline
ALQ \cite{ALQ} & Kernel-wise & ResNet-18 &  1 Bit MP & 2 Bits MP & 63.2\% (69.8\%) \\  
\hline
\multirow{3}{*}{Banner \textit{et al.} \cite{post4bits}} & \multirow{3}{*}{Channel-wise} & VGG &  4 Bits MP & 4 Bits MP & 70.5\% (71.6\%) \\
& & GoogleNet &  4 Bits MP & 4 Bits MP & 66.4\% (77.2\%) \\ 
& & ResNet-50 &  4 Bits MP & 4 Bits MP & 73.8\% (76.1\%) \\
\hline
\multirow{2}{*}{FracBits \cite{FracBits}} & \multirow{2}{*}{Kernel-wise} & MobileNet-V2 &  3 Bits MP & 3 Bits MP & 68.2\% (71.8\%) \\
& & ResNet-18 &  3 Bits MP & 3 Bits MP & 69.8\% (70.2\%) \\
\hline
DMBQ \cite{DMBQ} & Channel-wise & ResNet-18 &  1 Bit MP & 2 Bits MP & 63.5\% (70.3\%) \\
\hline\multirow{3}{*}{AutoQ \cite{autoq}} & \multirow{3}{*}{Kernel-wise} & ResNet-50 &  4 Bits MP & 4 Bits MP & 74.5\% (74.8\%) \\
& & SqueezeNet &  4 Bits MP & 4 Bits MP & 56.5\% (56.9\%) \\
& & MobileNet-V2 &  4 Bits MP & 4 Bits MP & 70.8\% (71.1\%) \\
\hline
\end{tabular}\label{table:mixed}
\begin{tablenotes}
    \item[a] ReLeQ \cite{releq} did not report the detailed accuracy and only reported the accuracy gap between original and compressed models (\textit{i.e.}, -0.1\% and -0.5\%) on AlexNet and MobileNet-V2 respectively.
\end{tablenotes}
\end{threeparttable}
\vspace{-10pt}
\end{table*}
\begin{table*}[!htbp]
\scriptsize
\centering
\caption{A summary of compression approaches with entropy coding for the results on ImageNet.
}
\vspace{-5pt}
\begin{tabular}{ l c | c |  c |  c }
\hline
Methods & Entropy Coding & Model & Compression Ratio  &  Accuracy (Original) \\
\hline
\multirow{2}{*}{Deep Compression~\cite{han16iclr}} & \multirow{2}{*}{Huffman Coding (FVL)} & AlexNet & $35 \times$ & 57.2\% (57.2\%) \\
& & VGG & $49 \times$ & 68.2\% (68.5\%) \\
\hline
\multirow{2}{*}{Coreset \cite{coresets}} & \multirow{2}{*}{Huffman Coding (FVL)} & AlexNet & $55 \times$ & 57.1\% (57.2\%) \\
& & VGG & $238 \times$ & 68.1\% (68.9\%) \\
\hline
\multirow{3}{*}{DeepCABAC \cite{DeepCABAC}} & \multirow{3}{*}{Arithmetic Coding (FVL)} & VGG & $64 \times$ & 69.4\% (69.4\%) \\
& & ResNet-50 & $17 \times$ & 74.1\% (76.1\%) \\
& & MobileNet-V1 & $8 \times$ & 66.2\% (70.7\%) \\
\hline
EPR \cite{EPR} & Arithmetic Coding (FVL) & ResNet-50 & $19 \times$ & 74.0\% (75.0\%) \\
\hline
\multirow{2}{*}{RDOC \cite{zhe2021rate}} & \multirow{2}{*}{Tunstall Coding (VFL)} & ResNet-50 & $20 \times$ & 75.0\% (75.5\%) \\
& & MobileNet-V2 & $10 \times$ & 70.2\% (71.0\%) \\
\hline
\end{tabular}
\label{table:coding}
\vspace{-5mm}
\end{table*}

Compared with layer-wise approaches, channel-wise or kernel-wise approaches can achieve higher accuracy because of their fine-grained quantization.
To this end, Khoram \textit{et al.} \cite{khoram18iclr} proposed Adaptive Quantization (AQ) which simplifies a trained DNN model by finding a unique, optimal precision for each network parameter such that the increase in loss is minimized.
Qu \textit{et al.} \cite{ALQ} proposed Adaptive Loss-aware Quantization (ALQ) which achieves an average bit width below one-bit without significant loss in inference accuracy.
Meanwhile, Banner \textit{et al.} \cite{post4bits} adopted channel-wise bit allocation to improve quantization precision and provided an analytic solution for quantization bit width, 
assuming certain distributions of parameters.
FracBits \cite{FracBits} generalized quantization bit width to an arbitrary real number to make it differentiable and learned channel-wise bit allocation during training.
Distribution-aware Multi-Bit Quantization (DMBQ) \cite{DMBQ} proposed a loss-guided bit width allocation strategy to adjust the bit width of weights and activations channel-wisely.
Lastly, AutoQ \cite{autoq} presented a hierarchical deep reinforcement learning approach to find the quantization bit width of channels 
while simultaneously optimizing hardware metrics such as latency and energy.
Table \ref{table:mixed} summarizes the mixed-precision quantization approaches discussed above.
In Table~\ref{table:mixed}, the bit width of mixed-precision quantization approaches is the average bit width across all the layers. For example, if a two-layer model quantizes one layer to 4 bits and the other one to 8 bits, then the average bit width of this mixed-precision quantization is (4+8)/2 = 6 bits. The variance in reported accuracies for CNN architectures like MobileNet-V2 across literature~\cite{dq, autoq} complicates the evaluation of quantization's impact on model performance. This inconsistency could be due to the differences in dataset pre-processing, training protocols, or minor architectural modifications. To accurately evaluate quantization effects, it requires a standardized approach towards reporting the performance metrics of both original and quantized models. This would facilitate a more precise isolation of quantization's effects from other influencing factors. As a result, emphasizing the relative changes in model accuracy resulting from quantization—rather than focusing solely on absolute accuracy figures—offers a more robust measure for evaluating the effectiveness of quantization techniques. 
Such an approach would help in better understanding the trade-offs between model efficiency and performance in the context of quantization, thereby contributing to more informed decisions in model optimization and deployment. 
Therefore, we present the accuracies of the original and compressed models to demonstrate the relative change in model accuracy.

\subsubsection{Transform Quantization}
\label{sec:transform_quantization}
Transform quantization decorrelates and quantizes the weights, which first rotates the space and then quantizes the data after rotation with mixed precision as shown in Fig.~\ref{fig:quant}(d).
Because the redundancy between the dimension of weights can be removed after decorrelation, transform quantization could quantize weights to lower bit width in the transformed space.
A unified framework was proposed \cite{young2021transform} to enable low bit-rate quantization of deep neural networks by combining quantization and dimension reduction (decorrelation) techniques. The framework introduces a theory of rate and distortion for neural network quantization, which formulates optimum quantization as a rate-distortion optimization problem. Optimal End-to-end Learned Transform (ELT) was then derived to solve optimal bit width allocation following decorrelation. Experimental results demonstrated that transform quantization significantly improved the state-of-the-art in neural network quantization.

A similar idea was explored in other neural network compression works.
Wang \textit{et al.}~\cite{dct} proposed a technique to compress neural networks that involved spatially decorrelating the weights using the DCT, followed by vector quantization on the decorrelated weights. Alongside the DCT approach, other methods aimed to reduce the dimension of neural network layers using principal component analysis (PCA) and related techniques. \cite{svd} employed singular value decomposition (SVD) to perform low-rank projection across kernels in convolution layers or across the rows and columns of weight matrices for fully-connected layers. Similarly, \cite{svd_p} extended this method using the generalized SVD. \cite{svd_p2} further applied SVD on all four axes of convolutional tensors. Additionally, Li \textit{et al.}~\cite{svd_p3} partitioned the kernels into subsets and performed SVD on each subset of kernels separately.

\subsubsection{Quantization for Transformers}
\label{sec:quant_vi}
Transformers~\cite{li2021bossnas,chen2021autoformer} now stand as the leading architecture for a range of computer vision tasks, encompassing image classification, object detection, and semantic segmentation.
However, achieving high-accuracy performance in transformers comes with the trade-off of substantial computational complexity, often involving tens of millions or even more parameters in a model.
The substantial number of parameters poses a significant challenge for deploying transformers on mobile devices and hinders their practical applications.
As a result, the quantization of transformers becomes more important and has received attentions.

\cite{bondarenko2021understanding} observed that activations of the transfer have high dynamic activation ranges and structured outliers and thus propose a per-embedding-group quantization scheme.  
\cite{dettmers2022gpt3} introduced a mixed-precision decomposition scheme, which isolates the outlier feature dimensions into a 16-bit matrix multiplication.  
Liu \textit{et al.}~\cite{liu2021post} recently introduced post-training quantization for transformers, integrating ranking-aware loss, bias correction, and mixed-precision techniques.
In particular, the inclusion of a ranking-aware loss aimed to preserve the relative order of quantized attention maps within transformers. Additionally, a bias correction method was implemented to adjust the distribution of quantized weights and activations. The exploration of mixed-precision quantization involved assigning more bit widths to layers with higher sensitivity, thus ensuring optimal performance retention.
Ding \textit{et al.} \cite{ding2022towards} went a step further in enhancing by refining the calibration process following linear layers.

\subsubsection{Entropy Coding}
\label{sec:entropy_encoding}
Quantized values can be further compressed using entropy coding in a lossless manner. 
There are two types of entropy coding methods
\cite{taubmanjpeg2000}: Fixed-to-Variable Length (FVL) entropy coding and Variable-to-Fixed Length (VFL) entropy coding. 
FVL coding maps a fixed number of symbols to variable-length codes, 
using arithmetic or Huffman coding.
FVL entropy coding methods achieve high coding efficiency (high compression ratio), but their decoding process is inefficient as it decodes codes bit by bit, leading to slower processing.
It has comparable coding efficiency as Huffman coding, with $10\times$ to $15\times$ faster decoding speed.
On the other hand, VFL coding maps a variable number of symbols to fixed-length codes, enabling byte-by-byte decoding, resulting in much faster decoding compared to FVL coding. Tunstall coding \cite{Tunstall1067phd} is a popular VFL coding method that achieves comparable coding efficiency as Huffman coding but offers decoding speeds that are $10\times$ to $15\times$ faster.

Several studies in the literature 
have utilized entropy coding to compress DNNs.
One such approach is the Deep Compression framework \cite{han16iclr}, which includes
pruning, quantization, and Huffman coding.
Similarly, Coreset-Based \cite{coresets} Compression coupled quantization with Huffman coding and exploited the weight redundancies to improve the compression ratio.
As mentioned above, the decoding of Huffman coding is inefficient, 
potentially slowing down the inference stage. 
To address this, DeepCABAC \cite{DeepCABAC} proposed a context-adaptive binary arithmetic coding to compress DNNs.
EPR \cite{EPR} represented the weights of neural networks in a latent space and used arithmetic coding to compress the representation. To address this, 
DeepCABAC and EPR employed arithmetic coding to obtain a higher compression ratio, although arithmetic coding has the most increased computational complexity, which is very difficult to implement.
RDOC \cite{zhe2021rate} adopted Tunstall coding to obtain superior compression capability and fast decoding speed.
Table \ref{table:coding} summarizes the works that utilize entropy coding for neural network compression.

In summary, different quantization approaches have advantages and disadvantages regarding the computational complexity, quantization loss, hardware implementation, etc. In uniform quantization, the space is divided
into equidistant quantization centroids. Such quantization scheme is straight-forward for implementation without additional overhead involved, but the quantization loss is relatively high. Non-uniform quantization is able to achieve less quantization loss because the quantizer is more flexible and can divide the space into different clusters based on the distributions of data. The drawbacks of non-uniform quantization is that it requires multiple code-books to store the quantization centroids. Mixed-precision quantization uses different bit widths to quantize different parts of parameters. It is more reasonable than uniform quantization because different parts of parameters may react distinctively to quantization. However, specific hardware is needed to support the computation with mixed precision (\textit{e.g.}, 4-bit vs. 8-bit). Different with all the others, transform quantization firstly rotates the space before quantization to remove redundancy between the dimensions. As a result, lower bit widths are able to be achieved after the space transform. The problem is that a large rotation matrix is usually needed for the space transform.

\vspace{-2.5mm}
\subsection{Network Pruning}
\label{subsec:pruning}
Pruning is a widely used technique for compressing neural networks by removing neurons from networks. 
This helps reducing the storage size required for the network parameters, as well as the memory access during inference to load the surviving weights (neurons) from off-chip to on-chip memory.
Over the years, numerous algorithms for pruning have been developed. 
\subsubsection{Pruning Strategies - Before, During, and After Training}
\label{subsec:pruning_schemes} 
Pruning solutions can be categorized according to multiple schemes. One scheme is when pruning is done with respect to the training of the model. Three choices exist here - pruning before training, during-training (pruning with training), or post-training (pruning after training).  

\textbf{Pruning before training} It aims to identify sparse structures in dense network at initialization and directly train the sparse network, with the merit of accelerating training and back-propagation. 
SNIP~\cite{lee2018snip} set the pruning criterion as the normalized magnitudes of gradients while GraSP~\cite{Wang2020Picking} pruned those weights whose removal will result in least decrease in the gradient norm after pruning.
FORCE~\cite{jorge2021progressive} presented a modified saliency metric based on \cite{lee2018snip}. 

\begin{table*}[ht]
\centering
\scriptsize
\caption{Classification of different pruning algorithms by their type, stage of doing pruning, and whether they are structured (denoted by `S') or unstructured (denoted by `U'). More comprehensive summary of Pruning at initialization scheme can be found in~\cite{ijcai2022p786}.}
\vspace{-5pt}
\resizebox{\textwidth}{!}{
\begin{tabular}{l|l}
\hline
Pruning methods & Pruning with training  \\
\hline
Gradient based & Optimal Brain Damage \cite{NIPS1989_250} [U], WoodFisher \cite{woodfisher} [U], MFAC \cite{mfac} [U], HALP~\cite{shen2022structural}, Taylor~\cite{molchanov2019importance}, RD~\cite{xu2023efficient}  \\
\hline
Customized criteria based
& DPF \cite{Lin2020Dynamic} [S,U], DST \cite{LIU2020Dynamic} [U],  Molchanov \textit{et al.} \cite{molchanov2016pruning} [U], SCOP~\cite{tang2020scop} [S]  \\
\hline
Regularization based & STR \cite{pmlr-v119-kusupati20a} [S,U], Savarese \textit{et al.} \cite{ContinuousSparsification2020} [U], Wang \textit{et al.} \cite{wang2021neural} [S,U] \\
\hline
Magnitude based &  RigL \cite{pmlr-v119-evci20a} [S,U], LAP \cite{Park2020LookaheadAF} [S,U], GMP \cite{Zhu2018ToPO} [U], LAMP \cite{lee2021layeradaptive} [U], Global MP \cite{globalmp} [U], DSR~\cite{ICML-2019-MostafaW} [U], SM\cite{sparse_momentum} [U]
 \\ 
\hline
Reinforcement learning based & PuRL \cite{gupta2020learning} [U], AMC \cite{he2018amc} [S,U], RNP \cite{NIPS2017_6813} [S]   \\
\hline
\end{tabular}
}
\label{table:pruning_methods}
\vspace{-5mm}
\end{table*}

\textbf{Pruning with training} It starts with a dense network from scratch and gradually prunes out more connections from the networks during training. 
Researchers proposed different pruning criteria (\textit{e.g.}, estimation of inverse Hessian in WoodFisher~\cite{woodfisher} and MFAC~\cite{mfac}, Taylor-based criterion in Molchanov \textit{et al.}~\cite{molchanov2016pruning}, $L2$ regularization based pruning~\cite{wang2021neural}, magnitude-based criterion~\cite{pmlr-v119-evci20a} and reinforcement learning based methods~\cite{gupta2020learning}),  to remove less important parameters. 
Some approaches either finished the pruning in one training pass (\eg, DPF~\cite{Lin2020Dynamic}) or multiple training passes~\cite{frankle2018the}. 
Another important technique is \textbf{prune-and-regrow}~\cite{pmlr-v119-evci20a}. Specifically, instead of removing a neuron once and for all during a pruning cycle in the training, neurons are given chances to grow back later. 
RigL~\cite{pmlr-v119-evci20a} droped parameters using parameter magnitudes and grow
the connections with the highest magnitude gradients. 
DSR~\cite{ICML-2019-MostafaW} randomly selected pruned neurons to grow back during training but keeps the active neurons in the whole network the same throughout. 
SM\cite{sparse_momentum} pruned weights
with small magnitude and growed new weights to fill in missing connections with the highest momentum magnitude. 
A comprehensive summary of the above pruning methods during training can be found in Tab.~\ref{table:pruning_methods}.
In summary, pruning before or during training offers the advantage of reducing the computational requirements throughout training, specifically in terms of floating-point operations per second (FLOPs).

\textbf{Post-training Pruning} 
 Different from pruning with training, post-training pruning typically conducts one-shot pruning after the model has been trained on a dataset~\cite{banner2019post, frantar2022optimal, kwon2022fast}.
 Post-training pruning in the context of Transformers has drawn significant interest due to growing concerns about the substantial resources and time needed for re-training.
 \cite{frantar2022optimal} proposed a post-training compression framework which covers both weight pruning and quantization in a unified setting based on an efficient realization of the
classical Optimal Brain Surgeon (OBS) framework~\cite{NIPS1989_250}. 
\cite{kwon2022fast} introduced a fast post-training framework that automatically
prunes the Transformer model using structured sparsity methods.

\subsubsection{Pruning Paradigms - Unstructured, Structured, and Semi-Structured Pruning}
\label{subsec:pruning_paradigms}
Another essential consideration is whether pruning occurs at the weight or channel level. It can be categorized into three types: unstructured pruning, structured pruning, and semi-structured pruning. 

\textbf{Unstructured Pruning}
Unstructured pruning refers to remove individual weights. 
Table~\ref{table:resnet50_imagenet} summarizes the top-performing algorithms for unstructured pruning. Among them, 
global magnitude pruning (Global MP)~\cite{globalmp} achieved the top performance in terms of sparsifying parameters. This method ranks the weights by their absolute magnitude and prunes the smallest ones, resulting in high parameter sparsity. 
On the other hand, STR~\cite{pmlr-v119-kusupati20a} achieved the top performance in terms of sparsifying FLOPs. It reparameterizes the weights using a soft threshold operator and learns the sparsity rates using gradient descent. 
GMP~\cite{Zhu2018ToPO} developed a simple gradual pruning approach that requires minimal tuning. 
DNW~\cite{DNW} provided an effective mechanism for discovering sparse subnetworks of predefined architectures in a single training run. 

\begin{table}[!ht]
\vspace{-2.5mm}
\scriptsize
\begin{threeparttable}
\caption{Results of unstructured pruning on ResNet-50 on ImageNet. 
\vspace{-2.5mm}
}
\begin{tabular}[t]{l|c|c|c|c}
\toprule
\multirow{1}{*}{Methods} & Top-1 & Params & Sparsity & FLOPs pruned\\
\midrule
ResNet-50 & 77.0\% & 25.6M\ & 0.00\% &  0.0\%\\
\midrule
GMP~\cite{Zhu2018ToPO} & 75.60\% & 5.12M\ & 80.00\% & 80.0\%\\
DSR\tnote{*\#} \ \cite{ICML-2019-MostafaW} & 71.60\% & 5.12M\ & 80.00\% & 69.9\%\\
DNW~\cite{DNW} & 76.00\% & 5.12M\ & 80.00\% & 80.0\%\\
SM~\cite{sparse_momentum} & 74.90\% & 5.12M\ & 80.00\% & -\\
SM(ERK)~\cite{pmlr-v119-evci20a} & 75.20\% & 5.12M\ & 80.00\% & 58.0\%\\
RigL\tnote{*} \cite{pmlr-v119-evci20a} & 74.60\% & 5.12M\ & 80.00\% & 77.5\%\\
RigL(ERK)~\cite{pmlr-v119-evci20a} & 75.10\% & 5.12M\ & 80.00\% & 58.0\%\\
DPF~\cite{Lin2020Dynamic} & 75.13\% & 5.12M\ & 80.00\% & 80.0\%\\
STR~\cite{pmlr-v119-kusupati20a} & 76.19\% & 5.22M\ & 79.55\% & 81.3\%\\
\textbf{Global MP (One-shot)}~\cite{globalmp} & \textbf{76.84\%} & 5.12M & 80.00\% & 72.4\%\\
\underline{Global MP (Gradual)}~\cite{globalmp} & \underline{76.12\%} & 5.12M & 80.00\% & 76.7\%\\
\midrule
\end{tabular}
\begin{tablenotes}
    \item[*] The first layer of the network architecture is dense.
    \item[\#] The last layer of the network architecture is dense.
\end{tablenotes}
\end{threeparttable}
\label{table:resnet50_imagenet}
\end{table}
\textbf{Structured Pruning}
Structured pruning targets the removal of organized neurons, such as channel-wise or block-wise pruning. Certain studies~\cite{8953212, lin2020hrank} established metrics, excluding $l1$-norm~\cite{Wang2020PruningFS}, to decide specific channel pruning. HRank~\cite{lin2020hrank} mathematically proved that filters with lower-rank feature maps contain less informative content, suggesting their suitability for removal. Another research direction explores automatic pruning without meticulous hyper-parameter adjustments, with DHP \cite{10.1007/978-3-030-58598-3_36} introduced a differentiable pruning technique using hypernetworks. This approach allows for automatic network pruning by using hypernetworks that take latent vectors as input and generate weight parameters for the backbone network. In parallel, a research line focuses on simultaneously pruning diverse redundant structures \cite{8953628, CHL00L21}. For instance, GAL~\cite{8953628} achieved simultaneous end-to-end pruning of diverse structures, such as channels, filters, and blocks, using label-free generative adversarial learning. GAL employs a sparse soft mask to scale specific structures' outputs to zero.
3D~\cite{CHL00L21} pruned a model by determining the optimal value of a pruned network along three dimensions, \textit{i.e.}, layers, filters, and image resolution. It did this by conducting multiple pruning runs to build a database of pruning rates along each dimension and the associated accuracy. It then used the Lagrangian theorem to fit the data to an optimal pruning function.
Table \ref{table:resnet50_structured} presents the top-performing algorithms for structured pruning. 
\vspace{-5mm}
\begin{table}[!ht]
\centering
\scriptsize
\caption{Structured pruning outcomes for ResNet-50 on ImageNet.}
\vspace{-5mm}
\resizebox{0.45\textwidth}{!}{
\begin{tabular}[t]{lcccc}
\toprule
\multirow{1}{*}{Methods} & Top-1 Acc & Params & Sparsity & FLOPs pruned\\
\midrule
ResNet-50 & 76.15\% & 25.6M\ & 0.00\% & 0.0\%\\
\midrule
GAL~\cite{8953628} & 71.95\% & 21.25M\ & 17.00\% & 43.0\% \\ 
FPGM~\cite{8953212} & 74.83\% & - & - & 54.0\% \\ 
HRank~\cite{lin2020hrank} & 74.98\% & 16.13M\ & 37.00\% & 44.0\% \\ 
DHP~\cite{10.1007/978-3-030-58598-3_36} & 75.45\% & 11.78M\ & 54.00\% & 49.9\% \\ 
\textbf{PScratch}\cite{Wang2020PruningFS} & 75.45\% & \textbf{9.22M}\ & \textbf{64.00\%} & \textbf{49.9\%} \\ 
\textbf{3D}~\cite{CHL00L21} & \textbf{75.90\%} & 12.03M\ & 53.00\% & \textbf{49.9\%} \\ 
\bottomrule
\end{tabular}
}
\label{table:resnet50_structured}
\end{table}

\textbf{Semi-structured Pruning}
Aside from the two major sparsity patterns, semi-structured pruning scheme is also adopted by few literature, which aligns the sparsity patterns with the GEMM (GEneric Matrix Multiplication) designs to achieve handy hardware acceleration while still benefits from relatively low accuracy drop. \cite{lagunas2021block} pruned feedforward weights in Transformer-based language models (\textit{e.g.}, BERT) in block-sparsity pattern and achieved a trade-off between accuracy retaining and acceleration. 
Apart from the regular block shaped type of semi-structure, another N:M sparsity~\cite{zhou2021learning} arises recently available on NVIDIA's A-series GPUs. Several Vision Transformers pruning works~\cite{fang2022algorithm,yang2023global,yu2023boost} explored using such sparsity scheme to achieve fine-grained feed-forward and attention pruning.

By examining Table~\ref{table:resnet50_imagenet} and Table~\ref{table:resnet50_structured}, it becomes evident that structured pruning is conducive to hardware constraints. Nevertheless, it prunes fewer neurons and FLOPs, resulting in lower memory costs when compared to fine-grained unstructured pruning while maintaining similar levels of accuracy.  More memory analysis of pruning can be found in~\cite{cheng2023survey}. 

There are some interesting works on exploring network pruning to meet specific computational constraints. One important technique is \textbf{slimmable network}~\cite{yu2018slimmable,yu2019universally,li2021dynamic}. Instead of optimizing a fixed and unified sparse network for inference, slimmable network aims to dynamically determine the network widths (\eg, layers, channels, or blocks) at runtime, to predict the sample with minimal performance sacrifice. 
\cite{yu2018slimmable} was one of the pioneer works in slimmable network, which proposed networks with variable width (number of channels) and an altered BatchNorm strategy to train the network with switchable widths by sharing network parameters. 
\cite{li2021dynamic} adopted learnable gating layers to decide the layer widths. 
Another important research line is \textbf{Once-for-All} networks~\cite{cai2019once}. Instead of training multiple models for various resource constraints, Once-for-All networks learn a shared set of parameters that can be applied to different subnetworks. These subnetworks vary in depth, width, or other architectural aspects, allowing the model to adapt to diverse computational requirements.   

In summary, network pruning has a long history and its effectiveness has been verified by numerous research works. 
Unstructured pruning is the de-facto scheme known for best performance preservation, but limited by the difficulties to deploy on real hardware to achieve speedups. The new N:M sparsity scheme on NVIDIA's Ampere platform offers a great fine-grained sparsity pattern similar unstructured pruning and with remarkable acceleration, but the restriction of row-wise sparsity constraint and the limited availability are still outstanding issues to address. Block-structured and Structured pruning look at larger structures in the networks as elements for pruning which is more hardware friendly. However, the dependency of neurons within a pruning group (\emph{e.g.} A 2d block of neurons for block-structured or all the neurons in a channel in structured pruning) poses challenge to evaluate the impact of removing the whole pruning group. The optimal pruning criteria for minimizing accuracy loss while removing redundancy remains an open question, especially considering the continuous evolution of novel neural network architectures. Classic criteria such as magnitude-based and derivative-based importance are originally proposed under post-train-pruning, where a prune-then-finetune procedure is required. These criteria are further borrowed into more advanced schemes such as prune-at-initialization and prune-during-training, where pruning are interleaved with the training. 
\vspace{-5pt}
\subsection{Knowledge Distillation}
\label{subsec:kd}
Knowledge Distillation (KD) compresses the learned knowledge from a large model (Teacher) into a smaller model (Student), which is designed to meet the latency and footprint requirements on resource-limited devices. This compression results in fewer trainable parameters and FLOPs while maintaining similar accuracy to the teacher model. The student architecture is flexible and can be optimized for hardware efficiency.
Due to its flexibility and efficiency in model compression, it has drawn a lot of interest and has also been verified in many research areas such as computer vision~\cite{wang2017model, lee2018teacher, zhu2021student, park2021learning}, neural language processing~\cite{kim2016sequence, gordon2019explaining,wang2021joint}, speech recognition~\cite{chebotar2016distilling, kwon2020adaptive,li2021mutual}, time series regression~\cite{xu2021kdnet, xu2022contrastive} and so on.
Two key challenges in KD research are defining what constitutes knowledge and developing effective methods for transferring that knowledge to the student model.
In this subsection, we first provide an overview of the structured knowledge in the literature, before delving into a discussion of the popular distillation schemes. 

\subsubsection{Structured Knowledge}
\label{sec:structured_knowledge}
Generally, knowledge can be broadly categorized into three types: knowledge from logits, features, and relations. 

\textbf{Knowledge from logits}
The terminology of ``knowledge distillation'' was first introduced by Hinton \textit{et al.} \cite{hinton2015distilling}. The logits from the softmax layer of a complex teacher model, softened by a temperature factor, were employed as the ``dark knowledge'' to guide the training process of a compact student model. Compared with ground truth labels, the softened logits serve as a regularization term to improve the generalization capability of the student model. Zhang \textit{et al.} \cite{zhang2018better} formalized the logits from multiple teachers as a graph and presented a logit graph distillation approach for video classification. More recently, Zhao \textit{et al.} \cite{Zhao_2022_CVPR} reformulated teacher's logits into two parts: TCKD and NCKD, which are a binary probability vector for target class and an independent probability vector for non-target classes, respectively. They empirically showed that the NCKD is critical to the success of distillation and therefore proposed to decouple them by introducing two independent hyper-parameters.


Many KD works have shown the effectiveness of using logits-based knowledge \cite{chen2017learning, yuan2019obtain, hegde2020variational}. However, this type of knowledge is originally designed for classification tasks but not for regression tasks where the output is a scalar. Despite this, recent works such as \cite{saputra2019distilling, xu2021kdnet, xu2022contrastive} have demonstrated the effectiveness of logits-based knowledge on time series regression tasks. By directly minimizing the distance between the teacher and student's scalar predictions, the generalization capability of the compact student can be improved.

\textbf{Knowledge from feature representations}
Utilizing the feature representations from the teacher's intermediate layers as the knowledge has received significant attention as they contain more meaningful information than logits. Directly aligning teacher and student's features via a specific distance metric can force a student to learn similar representations to its teacher. However, direct regression of teacher and student's feature maps can lead to suboptimal performance, resulting in less effective knowledge transfer.
Therefore, other well-structured knowledge derived from feature representations was further presented, such as attention maps~\cite{zagoruyko2016paying}, the flow of solution procedure~\cite{yim2017gift}, factor transferring~\cite{kim2018paraphrasing}, mutual information~\cite{ahn2019variational} and more. 
Furthermore, Liu \textit{et al.}~\cite{liu2021exploring} 
proposed using the spatial correlations between different channels of features as the knowledge to align the diversity and homology of feature space between the teacher and student.
direct regression teacher and student's feature maps. 
Lin \textit{et al.}~\cite{lin2022knowledge} addressed the sub-optimal issue of direct regression of teacher and student's feature maps by reconstructing the student's feature maps with a novel target-aware transformer.
Recently, Shang \textit{et al.}~\cite{shang2021lipschitz} formulated the knowledge as Lipschitz constants via a transmitting matrix derived from the input and output feature maps of each network block. Huang \textit{et al.}~\cite{huang2021revisiting} argued that we should encourage the student not only to inherit the knowledge from the teacher but also to explore more diverse feature representations via designing a novel dis-similarity loss among feature representations. Finally, Ji \textit{et al.}~\cite{ji2021show} explored the similarity relations between teacher and student's feature maps from different layers via an attention network to avoid manual layer pairing.

\textbf{Knowledge from instance relation}
Instead of simply mimicking teachers' responses via logits or feature representations on individual training instances, the relationships between different instances could also serve as knowledge. For example, Passalis and Tefas ~\cite{passalis2018learning} modeled teacher's knowledge as a probability distribution among a batch of samples and transferred it by minimizing the distribution divergence between teacher and student. In a similar vein, Park \textit{et al.}~\cite{park2019relational} proposed transferring the mutual relations instead of actual representations by introducing two novel distillation losses, allowing student to learn the structured relations among different samples. Liu \textit{et al.}~\cite{liu2019knowledge} formulated the instance relationship graph as the knowledge source, and  Huang \textit{et al.}~\cite{huang2022knowledge} aligned the inter-class and intra-class relation simultaneously for instances from different classes and same class, allowing for knowledge transfer from stronger teachers or more robust training strategies. Similarly, Yun \textit{et al.}~\cite{yun2020regularizing} minimized the intra-class variance among different instances having the same labels. Moreover, contrastive learning has been introduced for student representation learning and sample relation optimizing~\cite{tian2019contrastive, zhu2021complementary,xu2022contrastive}. Specifically, the teacher and student-generated outputs from the same sample are generally considered positive pairs, while the outputs of teachers generated from different samples are regarded as negative pairs. Contrastive learning minimizes the distance of positive pairs and maximizes the distance of negative pairs. 

To sum up, although the effectiveness of the aforementioned structured knowledge has been verified by many existing works, they have their advantages and disadvantages. The logit-based knowledge is the most widely-used knowledge due to its superior simplicity. However, the limited information provided by the logits severely hinders the efficacy of knowledge transformation. Meanwhile, its performance on other tasks (\textit{e.g.}, regression task) still requires more exploration. On the contrary, feature representations can offer more informative knowledge than logits for classification and regression tasks. However, they have a drawback that task-specific knowledge from feature representations may not be useful for other tasks. When distilling knowledge between different network architectures, pre-defined feature-based knowledge can harm the performance of students \cite{xu2021kdnet}. Choosing the right intermediate layers for distillation is challenging, especially when the teacher and student models differ significantly in capacity. Combining logit-based and feature-based knowledge is a common strategy in many KD-related works. Lastly, although some works have demonstrated that relation-based knowledge distillation works well, it does involve more complex training processes \cite{wang2021knowledge,peng2019correlation}, like extra memory for data storage and graph construction. Generally, it is also very challenging to integrate with the other two types of structured knowledge.

\subsubsection{Distillation Schemes}
\label{subsec:distillation_schemes}
Distillation schemes pertain to efficiently transferring knowledge from the teacher model to the student model, categorized into three types briefly: a) metric-specified vs. metric-free distillation; b) teacher-specified vs. Teacher-free distillation; and c) data-available vs. data-free distillation. More details can be found in Appendix C.
\subsection{Non-linear operations}
\label{subsec:nonlinear}
Most existing works focus on linear matrix operations in convolutional layers or fully connected layers which account for over 99\% of the total operations in modern DNNs \cite{geng2018hardware}. 
The implementation of non-linear blocks, such as Softmax and Sigmoid, presents significant difficulties for hardware. This is due to two main reasons. Firstly, convolutional operations are much simpler and cheaper to perform using a multiplier and adder, whereas non-linear functions like exponential require a more complex hardware unit. Secondly, the compute ratio per unit area for non-linear blocks is currently an order of magnitude lower than that for convolutional layers in existing accelerators. However, there has been a recent increase in research efforts to optimize non-linear operations for hardware implementations.
In this section, we review the literature which aims cost-efficient inference for non-linear operations.

\subsubsection{NMS}
\label{sec:nms}
 \IncMargin{0.66em}
The greedy non-maximum suppression (NMS) algorithm~\cite{greedynms} (Algorithm in Appendix D) sorts bounding boxes by class confidence scores and removes boxes with significant overlap (IoU) iteratively.

Table \ref{tab:nms_comparing} provides an overview of various NMS algorithms. In particular, except greedy NMS, we summarize them into six types. The first three types intend to optimize the sub-module in the greedy algorithm, whereas the last three seek an entire algorithm replacement.
The six types are: \textbf{1) greedy-sorting}:
In the greedy NMS, the boxes are sorted based on the categorical confidence but not the location confidence, and thus the ranking list might not be reliable. This type of methods~\cite{bodla2017soft, tychsen2018improving, jiang2018acquisition} focuses on producing a more reliable ranking list by 
incorporating extra 
location-based information 
into the ranking. For example, SoftNMS~\cite{bodla2017soft} decayed the confidence score of the remaining detection boxes using a function with the IoU as an argument, thereby affecting the sorting. 
\textbf{2) greedy-finalizing box}: it focuses on finalizing the box coordination~\cite{gahlert2020visibility, he2019bounding}. \textbf{3) greedy-duplicate check}: these methods focus on duplicate bounding box check criteria~\cite{liu2019adaptive, salscheider2021featurenms}. For example, \cite{liu2019adaptive} dynamically modified the IoU threshold in the duplicate check using object density. \textbf{4) end-to-end}: these methods replace NMS with a neural network sub-module and train in an end-to-end manner without manual intervention~\cite{hosang2016convnet,hosang2017learning, hu2018relation}. \textbf{5) clustering}: this kind of methods focuses on parallel clustering predicted bounding boxes to speed up the NMS~\cite{rothe2014non, gpunms2}. \textbf{6) pooling}: it focuses on reformulating NMS as a max pooling operation which is inherently
parallel and hardware-friendly
~\cite{shapoolnms, maxpoolnms, psrrmaxpool}. 


\begin{table*}[t]\centering
\caption{Comparing NMS algorithms in the literature}\label{tab:nms-alg-lit}
\vspace{-5pt}
\scriptsize
\begin{tabular}{lccccc}\toprule
Methods &Types & Network Replacement & Parallelable? & Two-stage detector? &Time Complexity \\\cmidrule{1-6}
GreedyNMS~\cite{greedynms} & greedy &\xmark &\xmark &\cmark &$\mathcal{O}(n\log(n)) + \mathcal{O}(nm)$ \\\cmidrule{1-6}
SoftNMS~\cite{bodla2017soft} & 1)  &\xmark &\xmark &\cmark &$\mathcal{O}(n\log(n)) + \mathcal{O}(nm)$ \\\cmidrule{1-6}
AdaptiveNMS~\cite{liu2019adaptive} &3) &\xmark &\xmark &\cmark &$\mathcal{O}(n\log(n)) + \mathcal{O}(nm)$ \\\cmidrule{1-6}
TNetNMS~\cite{hosang2016convnet} &4) &\cmark &\cmark &\cmark &Nework architecture related \\\cmidrule{1-6}
FeatureNMS~\cite{salscheider2021featurenms} &3) &\xmark &\xmark &\cmark &$\mathcal{O}(nm)$ \\\cmidrule{1-6}
vg-NMS~\cite{gahlert2020visibility} &2) &\xmark &\xmark &\cmark &$\mathcal{O}(n\log(n)) + \mathcal{O}(nm)$ \\\cmidrule{1-6}
MaxpoolNMS~\cite{maxpoolnms} &6) &\xmark &\cmark &\xmark &$\mathcal{O}(n)$ \\\cmidrule{1-6}
PSRR-MaxpoolNMS~\cite{psrrmaxpool} &6)  &\xmark &\cmark &\cmark &$\mathcal{O}(n)$ \\\cmidrule{1-6}
ClusteringNMS~\cite{rothe2014non} &5) &\xmark &\cmark &\xmark &$\mathcal{O}(nm)$ \\\cmidrule{1-6}
GPUNMS~\cite{gpunms2} &5)&\xmark &\cmark &\xmark &$\mathcal{O}(nm)$ \\\cmidrule{1-6}
FitnessNMS~\cite{tychsen2018improving} &1) &\xmark &\xmark &\cmark &$\mathcal{O}(n\log(n)) + \mathcal{O}(nm)$ \\\cmidrule{1-6}
SofterNMS~\cite{he2019bounding} &2) &\xmark &\xmark &\cmark &$\mathcal{O}(n\log(n)) + \mathcal{O}(nm)$ \\\cmidrule{1-6}
GnetNMS~\cite{hosang2017learning} &4)  &\cmark &\cmark &\cmark &Network architecture related \\\cmidrule{1-6}
IoU-Net~\cite{jiang2018acquisition} &1)  &\xmark &\cmark &\cmark &$\mathcal{O}(n\log(n)) + \mathcal{O}(nm)$ \\\cmidrule{1-6}
RelationNetNMS~\cite{hu2018relation} &4) &\cmark &\cmark &\cmark & Nework architecture related \\\midrule
\end{tabular}
\label{tab:nms_comparing}
\vspace{-5mm}
\end{table*}

\begin{table}[!htp]\centering
\caption{Comparing Softmax algorithms in the literature}\label{tab:softmax-alg-lit }
\vspace{-5pt}
\scriptsize
\begin{tabular}{lcl}\toprule
Methods &Types & Stage \\\cmidrule{1-3}
GPUSoftmax~\cite{joulin2017efficient} & 1)  & train \\\cmidrule{1-3}  
SvdSoftmax~\cite{shim2017svd} & 3)   & inference \\\cmidrule{1-3}
AdaptiveSoftmax~\cite{blanc2018adaptive} & 2) & train  \\\cmidrule{1-3}
SOFT~\cite{lu2021soft} & 5) & train \& inference  \\\cmidrule{1-3}
ANNSoftmax~\cite{zhao2021ann} & 2) &train  \\\cmidrule{1-3}
Sigsoftmax~\cite{chen2018learning} & 5) & train \& inference  \\\cmidrule{1-3}
L2S~\cite{chen2018learning} & 1) &inference  \\\cmidrule{1-3}
DS~\cite{liao2019doubly} & 1) &inference  \\\cmidrule{1-3}
RF-Softmax~\cite{rawat2019sampled} & 2)  & train  \\\cmidrule{1-3}
RegularizerSoftmax~\cite{andreas2015accuracy} & 4) & inference\\\cmidrule{1-3}
\end{tabular}
\label{tab:softmax}
\vspace{-5mm}
\end{table}

\subsubsection{Softmax}
\label{sec:softmax}
Softmax is a general operation to predict the class probability in many tasks. Its equation is $\sigma(z_i) = \frac{e^{z_i}}{\sum_{j=1}^{K} e^{z_j}}$ 
where $z_i$ denotes the  $i$-th element of the input vector $\mathbf{z}$. $K$ denotes the number of classes. Softmax involves expensive non-linear operations such as exponentials and divisions, and its computational cost can become prohibitively expensive when dealing with a large number of classes.

Researchers have developed methods to speed up softmax computation in Natural Language Processing (NLP), as it can dominate the complexity of neural networks with large vocabularies.
Table~\ref{tab:softmax} summarizes 
various softmax approximation methods in the literature, which can be categorized into five types: 
1) \textbf{class-based hierarchical softmax}~\cite{joulin2017efficient}: . A tree structure is constructed to estimate the softmax value along the tree depth, avoiding the need to traverse the whole vocabulary. 
\textbf{2) sampling-based softmax}: it chooses a small subset of possible outputs and trains only with those. \textbf{3) differentiated softmax}: this restricts the effective parameters, using the fraction of the total output matrix. The matrix allocates a higher dimensional representation to frequent words and only a lower dimensional vector to rare words. \textbf{4) self-normalization}: it employs an additional training loss term~\cite{andreas2015accuracy}, which leads to a normalization factor close to 1. \textbf{5) softmax replacement}: a new probability computation function is proposed for accuracy improvement~\cite{chen2018learning} or computational saving~\cite{lu2021soft}. 




\subsubsection{Activation functions}
\label{sec:activation_functions}

Activation functions like ReLU~\cite{krizhevsky2012imagenet}, PReLU~\cite{he2015delving}, and Sigmoid are crucial in neural networks and should be nonlinear, differentiable, continuous, bounded, and zero-centered. The implementation of activation functions with linear operations is easy, while exponential or complex nonlinear operations may require linear approximations for faster inference. Choosing an appropriate activation function depends on the specific neural network requirements and the available computational resources.

To enhance the efficiency and performance of neural networks, the optimization of non-linear blocks, including activation functions like softmax and sigmoid, is a crucial focus of research. The challenges arise from the computational complexity of these functions, particularly in hardware implementations where simpler operations, such as those in convolutional layers, are more cost-effective. Recent efforts have been directed towards devising hardware-friendly activation functions, employing quantization techniques, utilizing look-up tables, and exploring fixed-point arithmetic. 
These optimization endeavors aim to strike a balance between computational efficiency and model accuracy, facilitating the deployment of neural networks in diverse hardware environments, including edge devices and real-time systems.
\subsection{NAS and TinyML}
\label{sec:nas_tinyml}
\vspace{-2pt}
\subsubsection{NAS}
\label{subsec:nas}
The research on Neural Architecture Search (NAS) for efficient model design has yielded significant advancements in automating the creation of compact yet high-performing neural networks. Notable works such as EfficientNet~\cite{tan2019efficientnet}, ProxylessNAS~\cite{cai2018proxylessnas}, MnasNet~\cite{tan2019mnasnet}, FBNet~\cite{wu2019fbnet}, DARTS~\cite{liu2018darts} and Once-for-All~\cite{cai2019once}, showcase various approaches to NAS that consider efficiency in terms of model size, computational resources, and hardware constraints. These methods leverage techniques like gradient-based optimization~\cite{liu2018darts}, platform-awareness~\cite{wu2019fbnet}, parameter sharing~\cite{pham2018efficient, chen2021autoformer}, various network operations~\cite{li2021bossnas},  and specialized network training~\cite{cai2019once} to find architecture configurations that strike a balance between efficiency and performance.


\subsubsection{TinyML}
\label{subsec:tinyml}
The study of Tiny Machine Learning (TinyML) has garnered significant attention due to its focus on deploying deep learning models on edge devices with limited resources, such as IoT devices. There is a growing body of literature and available resources in this field. These methods make use of various techniques, taking into account hardware constraints, including factors like energy consumption, Random Access Memory (RAM), and  multiply–accumulate (MAC) operations. For example, Ancilotto \textit{et al.} discussed considerations related to hardware constraints in their work~\cite{ancilotto2023xinet}. Additionally, Lin \textit{et al.} addressed the issue of imbalanced memory distributions in Convolutional Neural Networks (CNNs) by implementing patch-by-patch scheduling~\cite{lin2021mcunetv2}. Cai \textit{et al.} proposed a method to reduce memory usage during training by freezing weights in transfer learning~\cite{cai2020tinytl}.

\section{When NN compression meets Hardware}
\label{sec:compression_hardware}
\subsection{Overview}

Performance, power, and area (PPA) are the three most important metrics in evaluating hardware cost.
Table II in Appendix E summarizes the quantitative and relative energy and area consumption of various operations such as adding and multiplication in different data formats.
We observe that energy efficiency can be achieved through good data addressing, as data reading is relatively expensive. 
It costs 166.7$\times$ and 1,666.7$\times$ more energy to read 32-bit data from an 8KB static random-access memory (SRAM) and 1MB SRAM than perform an 8-bit integer addition, respectively. 
In addition, quantizing float-point data to integers, from high precision (\eg, 32-bit) to low precision (\eg, 8-bit), reduces energy consumption. For instance, a 32-bit float-point addition 
requires 30$\times$ more energy than an 8-bit integer addition. 
Finally, the trend of the area costs of different operations is 
similar to their energy costs.

For the linear operations in DNNs, quantization, and pruning~\etc can be applied to reduce the number of weights or even activations, hence optimizing the PPA of the hardware.
For the non-linear operations, however, hardware-friendly approximation algorithms are needed to save the hardware cost.

Hardware accelerators generally target either Field Programmable Gate Arrays (FPGAs) or Application-Specific Integrated Circuits (ASICs). One accelerator can either be mapped into the FPGA flexibly for quick validation and further design space explorations or be fabricated as a high-performance ASIC chip when its functionalities and architecture are frozen and it is well-validated.





\subsection{Hardware Accelerating on Linear Operations}

\subsubsection{Hardware Accelerating on CNNs and Transformers}
\label{sec:cnns_transformers}

With the rapid advancements of DNNs, the corresponding hardware accelerators have become increasingly important.
Most CNN accelerators use direct convolution as their primary computation method, which accumulates the product of the inputs and corresponding weights in a certain dataflow.
Members of DianNao family~\cite{chen2014diannao,chen2014dadiannao,liu2015pudiannao} computes convolutions directly.
DianNao~\cite{chen2014diannao} exploited the locality properties of large-scale layers, achieving impressive results with only 3 mm$^2$ at 65 nm. 
However, the memory wall becomes the bottleneck when computing classifier and convolutional layers with private kernels~\cite{chen2014dadiannao}.
 To address this, DaDianNao~\cite{chen2014dadiannao} processed convolution with the data from nearby SRAM buffers and embedded dynamic random access memory (eDRAM) banks.
NVDLA~\cite{nvdla}, an open-source hardware inference accelerator, provided direct convolution mode. A wide multiply-accumulate (MAC) pipeline was implemented to parallel process the convolutional operations with memory bandwidth optimization.
Another accelerator, Maeri~\cite{kwon2018maeri} supported arbitrary dataflows by utilizing fine-grained reconfigurable tree-based interconnection network typologies to shape different sizes and numbers of virtual neurons.

By transforming the regular convolution into Generalized Matrix-matrix Multiplication (GEMM) or Generalized Matrix-vector Multiplication (GEMV), some matrix multiplication optimization algorithms, such as Strassen algorithm~\cite{strassen1969gaussian}, Winograd algorithm~\cite{winograd1980arithmetic} and Fast Fourier Transform (FFT)~\cite{brigham1967fast} can be applied to speed up the computation. For example,
Cong~\etal\cite{cong2014minimizing} applied the Strassen algorithm to the convolution transformed in GEMM and reduced the number of multiplications from $\mathcal{O}(N^3)$ to $\mathcal{O}(N^{2.807})$.
Lavin~\cite{lavin2016fast} and NVDLA~\cite{nvdla} applied Winograd convolution to reduce the number of multiplications while increasing the adders for the transformation.
The FFT algorithm converts the data to the more computationally efficient Fourier domain.
Previous work~\cite{ben1997fast} explored the application of FFT to accelerate inference, while
more recent works have attempted to map FFT on GPU platforms~\cite{mathieu2013fast,vasilache2014fast}.
The architecture proposed by Liang~\etal\cite{liang2019evaluating} also supported the FFT algorithm.

Over the past few years, the Transformer model has replaced Recurrent Neural Networks (RNNs)  and CNNs in the Natural Language Processing (NLP) and Computer Vision (CV) domains. 
However, Transformer models have different computation patterns than traditional neural networks, with more weights and computations, necessitating careful hardware accelerator design.
The hardware accelerating solutions for Transformers are reported in~\cite{lu2020hardware,ham20203,hu2021vis,sun2022vaqf,yu2022nn,marchisio2023swifttron,nag2023vita}. 
The accelerator for the multi-head attention (MHA) ResBlock and the position-wise feed-forward network (FFN) ResBlock in the NLP Transformer was first reported by Lu~\etal~\cite{lu2020hardware}.
In A$^3$~\cite{ham20203}, an approximate method was applied for efficient attention computation.
Vis-TOP~\cite{hu2021vis} focused on accelerating Visual Transformers (ViTs) and presented a customized hardware architecture for a three-layer, two-level basic Transformer. 
VAQF~\cite{sun2022vaqf} supported quantized ViTs with binary weights and low-bit activations on FPGA for inference.
SwiftTron~\cite{marchisio2023swifttron} directly mapped the ViT on ASIC with an area of 273 mm$^2$ and power of 33.64 W, adopting approximated non-linear functions proposed by NN-LUT~\cite{yu2022nn}, where they were computed via Piece-wise Linear Functions.
The accelerator introduced by Nag~\etal~\cite{nag2023vita} targeted FPGAs with dedicated computation blocks for the GEMMs in the MHA.

In summary, hardware accelerators for DNNs can use direct convolution or optimization algorithms like Strassen or Winograd to accelerate computation. Furthermore, specialized accelerators have been created for Transformer models.

\subsubsection{Accelerating Networks after Quantization: Mixed-Precision and Entropy Coding}
\label{sec:accelerator_quantization}

The bit width of weights in different layers or channels can be different to balance model precision and size. 
Table~\ref {tab:hw-mix-pre} summarizes the popular hardware accelerators that support multiple precision data formats. In~\cite{yin20171}, 8-bit and 16-bit fixed-point NN layers were supported.
Zhang~\etal~\cite{zhang2019new} proposed a new flexible unit that supports five different precision in training and inference. 
LNPU~\cite{lee20197} was a 
highly energy-efficient
DNN accelerator that supports training and inference, 
achieving up to 25.3 TFLOPS/W energy efficiency 
when processing a highly sparse weight layer in 8-bit float-point data format.
Zhou~\etal~\cite{zhou2020convolutional} proposed a CNN accelerator that supports mixed precision computations both within-layer and layer-wise. 
By introducing a fine-grained mixed precision unit, 
the hardware enables within-layer mixed-precision operations, which reduces computation area by nearly 50\% and dynamic power by around 12.1\% in AlexNet and VGG16 compared to the baseline.
Some works exploit the low-bit precision such as 4-bit or sub-4-bit.
The accelerator proposed by Fleischer~\etal~\cite{fleischer2018scalable} supported very low-precision data formats such as binary and ternary for aggressive inference performance. 
In addition, the 16-bit float-point data format is supported for high accuracy in training, hence achieving 24 Tera Operations Per Second (TOPS) inference performance in a $9~mm^2$ chip. 
Wang~\etal~\cite{wang2018design} introduced a flexible design flow that supports mixed-precision computations of hybrid neural networks with quantized activation and quantized weights targeting FPGAs.
With parameterized computation engines, the introduced AccELB performs binary and ternary convolution without multiplications.
The flexible 4-core AI chip introduced by Agrawal~\etal~\cite{agrawal20219} supported both training and inference with multiple precisions, \ie, INT2, INT4, hybrid-FP8, FP16, and FP32 with a performance of up to 102.4 TOPS.

\begin{table*}[!t]
\centering
\scriptsize
\caption{A Summary of mixed-precision hardware and their compute density (TOPS/W or TOPS/mm$^2$).}
\vspace{-5pt}
\label{tab:hw-mix-pre}
\resizebox{\textwidth}{!}{%
\begin{threeparttable}
\begin{tabular}{l|c|c|c|c|c|c|c|c}
\hline
Methods &
  \begin{tabular}[c]{@{}c@{}}Supporting\\      Format\end{tabular} &
  Platform &
  Area &
  Technology &
  \begin{tabular}[c]{@{}c@{}}Inference\\      Performance\end{tabular} &
  Frequency &
  \begin{tabular}[c]{@{}c@{}}Training or   \\      Inference\end{tabular} &
  \begin{tabular}[c]{@{}c@{}}Compute   Density\\      (TOPS/W or TOPS/mm$^2$)\end{tabular}\\ \hline
  Yin~\etal~\cite{yin20171} &
  INT8, INT16 &
  ASIC &
  19.36 mm$^2$ &
  65 nm lp &
  368.4 GPOS &
  200 MHz &
  Inference &
  1.28 TOPS/W \\ \hline
\multirow{5}{*}{\begin{tabular}[c]{@{}l@{}}Zhang~\etal\\      ~\cite{zhang2019new}~\tnote{a}\end{tabular}}&
  INT4 &
  \multirow{5}{*}{ASIC} &
  \multirow{5}{*}{2943 $\mu\text{m}^2$} &
  \multirow{5}{*}{28 nm} &
  29.6   GOPS / unit &
  \multirow{5}{*}{3.7 GHz} &
  \multirow{5}{*}{Both} &
  4.81   TOPS/W \\ \cline{2-2} \cline{6-6} \cline{9-9} 
 &
  INT8 &
   &
   &
   &
  14.8   GOPS / unit &
   &
   &
  2.09   TOPS/W \\ \cline{2-2} \cline{6-6} \cline{9-9} 
 &
  INT16 &
   &
   &
   &
  7.4   GOPS / unit &
   &
   &
  1.01   TOPS/W \\ \cline{2-2} \cline{6-6} \cline{9-9} 
 &
  FP8 &
   &
   &
   &
  14.8   GFLOPS / unit &
   &
   &
  1.10   TFLOPS/W \\ \cline{2-2} \cline{6-6} \cline{9-9} 
 &
  FP16 &
   &
   &
   &
  7.4   GFLOPS / unit &
   &
   &
  0.55   TFLOPS/W \\ \hline
\multirow{2}{*}{LNPU~\cite{lee20197}} &
  FP8 &
  \multirow{2}{*}{ASIC} &
  \multirow{2}{*}{16 mm$^2$} &
  \multirow{2}{*}{65 nm} &
  600   GFLOPS &
  \multirow{2}{*}{200 MHz} &
  \multirow{2}{*}{Both} &
  3.48   TFLOPS/W (FP8, 0\% sparsity) \\ \cline{2-2} \cline{6-6} \cline{9-9} 
 &
  FP16 &
   &
   &
   &
  300   GFOPS &
   &
   &
  25.3   TFLOPS/W (FP8, 90\% sparsity) \\ \hline
Zhou~\etal~\cite{zhou2020convolutional} &
  INT8, INT16 &
  FPGA &
  -- &
  -- &
  -- &
  -- &
  Inference &
  -- \\ \hline
\multirow{3}{*}{\begin{tabular}[c]{@{}l@{}}Fleischer~\etal\\      ~\cite{fleischer2018scalable}~\tnote{b}\end{tabular}} &
  Binary &
  \multirow{3}{*}{ASIC} &
  \multirow{3}{*}{9 mm$^2$} &
  \multirow{3}{*}{14 nm} &
  24   TOPS &
  \multirow{3}{*}{1.5 GHz} &
  \multirow{3}{*}{Both} &
  2.67   TOPS/mm$^2$ \\ \cline{2-2} \cline{6-6} \cline{9-9} 
 &
  Ternary &
   &
   &
   &
  12   TOPS &
   &
   &
  1.33   TOPS/mm$^2$ \\ \cline{2-2} \cline{6-6} \cline{9-9} 
 &
  FP16 &
   &
   &
   &
  1.5   TFLOPS &
   &
   &
  0.17   TOPS/mm$^2$ \\ \hline
AccELB~\cite{wang2018design} &
  \begin{tabular}[c]{@{}c@{}}Hybrid-INT1, \\ INT2, INT4, INT8\end{tabular} &
  FPGA &
  -- &
  -- &
  0.49-10.3 TOPS &
  200 MHz &
  Inference &
  -- \\ \hline
\multirow{5}{*}{\begin{tabular}[c]{@{}l@{}}Agrawal~\etal\\~\cite{agrawal20219}\end{tabular}} &
  INT2 &
  \multirow{5}{*}{ASIC} &
  \multirow{5}{*}{16 mm$^2$} &
  \multirow{5}{*}{7 nm} &
  -- &
  \multirow{5}{*}{1.0-1.6 GHz} &
  \multirow{5}{*}{Both} &
  -- \\ \cline{2-2} \cline{6-6} \cline{9-9} 
 &
  INT4 &
   &
   &
   &
  64-102.4 TOPS &
   &
   &
  8.9-16.5 TOPS/W \\ \cline{2-2} \cline{6-6} \cline{9-9} 
 &
  Hybrid-FP8 &
   &
   &
   &
  16-25.6 TFLOPS &
   &
   &
  1.9-3.5 TFLOPS/W \\ \cline{2-2} \cline{6-6} \cline{9-9} 
 &
  FP16 &
   &
   &
   &
  8-12.8 TFLOPS &
   &
   &
  0.98-1.8 TFLOPS/W \\ \cline{2-2} \cline{6-6} \cline{9-9} 
 &
  FP32 &
   &
   &
   &
  -- &
   &
   &
  -- \\ \hline
\end{tabular}%

\begin{tablenotes}
\scriptsize
\item[a] This work proposes one computation unit. Hence the frequency is computed by $1/\text{delay}$, where the delay is reported. The inference performance is computed by $2 \times \text{\#parallel op } \times \text{frequency}$, where the number of parallel operations in one unit is reported.  The energy efficiency is computed by $1/\text{energy \#op}$, where the energy per operation is reported.
\item[b] The energy efficiency is not reported; area efficiency is reported instead. 
\end{tablenotes}

\end{threeparttable}
}
\vspace{-5mm}
\end{table*}

Besides quantization, entropy coding can lower hardware costs while maintaining accuracy by encoding weights in a more condensed representation, resulting in fewer bits per variable~\cite{recanatesi2019dimensionality}. This is accomplished by taking advantage of the peaked distribution of quantized values.

The coding schemes are classified into Fixed-to-Variable (F2V) and Variable-to-Fixed (V2F) entropy coding methods. 
F2V coding methods, such as arithmetic coding~\cite{oktay2019scalable,wiedemann2019deepcabac} and Huffman coding~\cite{dubey2018coreset,han2015deep}, encode a fixed number of symbols into variable-length codewords. 
Hence, it is challenging for the F2V coding methods to decode parallelly.
F2V coding methods have very high computational complexity for decoding. 
The decoding complexity of F2V coding methods is $O(n \cdot k)$, where $n$ is the number of codewords, and $k$ is the reciprocal compression ratio. Conversely, V2F coding methods, such as Tunstall coding~\cite{chen2021efficient}, encode multiple symbols to a fixed number of bits. 
During the decoding stage, it is feasible to process multiple bits simultaneously and decode multiple symbols per clock cycle. Additionally, parallel decoding of the encoded string is possible by dividing it into bit chunks of fixed length according to the codeword length.


Huffman coding algorithm first lists symbols in decreasing order of their probabilities. Then, it constructs a Huffman tree by adding two symbols with the smallest probabilities at each step and removing them from the list. Finally, an auxiliary symbol is added to represent the two removed original symbols. Arithmetic coding assigns one code to the entire input instead of coding individual symbols like Huffman coding. It starts with a specified interval and symbolically reads the input while narrowing down the gap.

In Deep Compression~\cite{han2015deep} and Coreset-base Compression~\cite{dubey2018coreset}, Huffman coding~\cite{cormen2009introduction} was used to compress the quantized weights.
However, in~\cite{oktay2019scalable} and~\cite{wiedemann2019deepcabac}, another Fixed-to-Variable (F2V)\cite{rabbani2002jpeg2000} coding method, arithmetic coding\cite{witten1987arithmetic}, was used. 
Nevertheless, 
it is challenging to develop parallel implementations for decoding, as F2V codewords are of variable length and cannot be indexed efficiently, making in-efficient indexing and preventing the decoding of multiple symbols per clock cycle. 
In~\cite{chen2021efficient}, the Tunstall coding method was to compress ResNet and MobileNet after compression, where two Tunstall hardware decoders are designed and compared. Table~\ref{tab:hw-coding} summarizes these coding methods. Table III in Appendix E~\cite{chen2021efficient} reports the 256 entries Huffman hardware decoder consumes $3.14\times$ more hardware resources and costs $6.23\times$ more time to decode the same amount of weights.

\begin{table}[t!]
\scriptsize
\centering
\caption{The hardware implementations of various coding methods and the size of the ResNet-50 after compression.}
\vspace{-5pt}
\label{tab:hw-coding}
\resizebox{\columnwidth}{!}{%
\begin{tabular}{l|c|c|c|c|c}
\hline
Methods &
  \begin{tabular}[c]{@{}c@{}}Coding \\ Method\end{tabular} &
  \begin{tabular}[c]{@{}c@{}}Coding \\ Schemes\end{tabular} &
  Parallelable &
  \begin{tabular}[c]{@{}c@{}}Memory \\ Aligned\end{tabular} &
  \begin{tabular}[c]{@{}c@{}}Size After\\ Compression (MB)\end{tabular} \\ \hline
\begin{tabular}[c]{@{}l@{}}Deep \\ Compression ~\cite{han2015deep}\end{tabular}              & Huffman    & F2V & \xmark & \xmark & 6.06 \\ \hline
\begin{tabular}[c]{@{}l@{}}Coreset-base \\ Compression~\cite{dubey2018coreset}\end{tabular} & Huffman    & F2V & \xmark & \xmark & 6.46 \\ \hline
Oktay~\etal~\cite{oktay2019scalable}             & Arithmetic & F2V & \xmark & \xmark & 5.49 \\ \hline
Deepcabac~\cite{wiedemann2019deepcabac}          & Arithmetic & F2V & \xmark & \xmark & 6.06 \\ \hline
Chen~\etal~\cite{chen2021efficient}             & Tunstall   & V2F & \cmark & \cmark & 5.85 \\ \hline
\end{tabular}%
}
\vspace{-5mm}
\end{table}

In summary, mixed-precision and entropy coding are adopted by hardware accelerators after the quantization of the DNNs for higher compute density and performance. However, aggressive quantization would introduce significant accuracy loss.
Instead, entropy coding maintains accuracy while compressing the model.
However, F2V coding methods challenge the parallel decoding on the hardware, which introduces extra overhead to the throughput, while the decoding process of the V2F coding methods, such as the Tunstall coding, is hardware-friendly and preferred.

\subsubsection{Accelerating Networks after Pruning: Sparse Architecture}
\label{sec:accelerator_pruning}

To reduce the energy efficiency significantly and boost computation speed with fewer computation units, recently works~\cite{lee20197,chen2016eyeriss,han2016eie,parashar2017scnn,chen2019eyeriss} proposed accelerators that support sparse neural networks.
As summarized in~\cite{dave2021hardware}, there are five hardware accelerating methods for sparse neural networks:
\begin{itemize}
    \item Storing the compressed data in off-chip memory to optimize the memory capacity requirement and improve energy efficiency;
    \item Storing the compressed data in on-chip memory to optimize the memory capacity requirement and improve energy efficiency;
    \item Skipping zero elements to improve energy efficiency;
    \item Reducing ineffectual computation cycles to improve performance and energy efficiency;
    \item Balancing the workloads of different processing elements to increase performance;
\end{itemize}

Lee~\etal~\cite{lee20197} proposed an architecture called LNPU that supports sparse DNNs with fine-grained mixed precision of FP8-FP16.
Sparsity is exploited with intra- and inter-channel accumulation with the input load balancer (ILB) to alleviate the imbalanced workload problem caused by irregular sparsity, improving PE utilization.
Only non-zero weights are kept in the internal buffers in Cambricon-X~\cite{zhang2016cambricon}, while SCNN~\cite{parashar2017scnn} compressed both non-zero weights and activations in both dynamic random-access memory (DRAM) and internal buffers.

Sparse encoding methods can reduce memory access, increase energy efficiency, and accelerate computation time for accelerators by adapting them to sparse tensors. This is because most sparsity encoding methods only store non-zero elements.
These encoding methods are Coordinate (COO), COO-1D, Run-length Coding (RLC), Bitmap, Compressed Sparse Row (CSR), Compressed Sparse Column (CSC)~\cite{gustavson1972some} and Compressed Sparse Fiber (CSF)~\cite{smith2015splatt}.
Table~\ref{tab:sparse-encodings-overhead} summarizes the storage overhead and decoding taxonomy for standard encoding methods suitable for sparse data~\cite{dave2021hardware} while Table~\ref{tab:hw-sparsity-encoding} summarizes the area and energy efficiency of the hardware accelerators that use sparsity encoding methods.

\begin{table}[t!]
\centering
\scriptsize
\caption{Storage overhead and decoding taxonomy for common encoding methods. Vector $d$ stores $n$ dimensions of a tensor that contains $NNZ$ non-zero elements~\cite{dave2021hardware}.}
\vspace{-5pt}
\label{tab:sparse-encodings-overhead}
\resizebox{\columnwidth}{!}{%
\begin{tabular}{l|c|c}
\hline
Format &
Storage Overhead (bits) &
Decoding Taxonomy
\\ \hline
COO   & $NNZ \times \sum_1^n { \lceil \log_2{d_i} \rceil}$                                          & Direct                     \\ \hline
COO-1D & $NNZ \times \lceil \log_2{\prod_1^n d_i} \rceil$                                            & Direct                     \\ \hline
RLC    & $NNZ \times B$                                                                              & Single-step                \\ \hline
Bitmap & $\prod_1^n d_i$                                                                             & Single-step                \\ \hline
CSR    & $NNZ \times \lceil \log_2{d_1} \rceil$ + $(d_0 + 1) \times \lfloor \log_2{NNZ} + 1 \rfloor$ & Double-step                \\ \hline
CSC    & $NNZ \times \lceil \log_2{d_0} \rceil$ + $(d_1 + 1) \times \lfloor \log_2{NNZ} + 1 \rfloor$ & Double-step                \\ \hline
\end{tabular}%
}
\vspace{-4.5mm}
\end{table}

The COO sparsity encoding method was adopted in~\cite{hegde2019extensor,yuan2018sticker}.
COO-1D was utilized in~\cite{zhang2019snap, kang2019accelerator, lu2019efficient}.
RLC was used in~\cite{chen2016eyeriss, zhang2019compact, parashar2017scnn, lee20197}.
Bitmap was used in~\cite{zhang2019compact, yuan2018sticker, zhou2018cambricon, nvdla}.
Mishra~\etal~\cite{mishra2017fine} introduced CSR into the accelerator.
Data in~\cite{han2016eie, chen2019eyeriss, mishra2017fine, han2017ese} was compressed in CSC format.
Extensor~\cite{hegde2019extensor} utilized CSF encoding format.

More specifically, Eyeriss v2~\cite{chen2019eyeriss} encoded the weights in CSC data format for both on- and off-chip accessing to reduce the bandwidth requirements and energy consumption. Hence $1.2\times$ and $1.3\times$ improvement in speed and energy consumption are obtained by introducing CSC into Eyeriss v2. 
To best utilize the benefit of CSC data format, the PE in Eyeriss v2 is specially designed, where there are registers to store both address vectors and data vectors in the CSC compressed data, which is drawn in Fig. 1 within Appendix E. 

Sticker~\cite{yuan2018sticker} compressed the sparse weight into CSC format offline and decodes weight on-chip. In addition, there are COO encoders and decoders in Sticker to adopt the COO format for activations online.
Zhang~\etal~\cite{zhang2019snap} designed an associative index matching (AIM) module in their hardware accelerator SNAP before ending the data into a multiplier array to find the non-zero partial sum position, \textit{i.e.}, both the input activation and the corresponding weight are non-zero. 
A similar module was integrated in Cambricon-S~\cite{zhou2018cambricon} where the bitmap encoding method was deployed; hence the comparators in SNAP are replaced by AND gates, depicted in Fig. 2 within Appendix E. 
Envision~\cite{moons201714} exploited sparsity by representing the data in bitmap format, storing those flags in a GRD memory, hence guarding both memory fetches and MAC operations. 
SCNN~\cite{parashar2017scnn} employed a run-length encoding scheme, encoding the number of zeros between elements into the index vector, thus processing only non-zero weights and activations.

\begin{table*}[!t]
\centering
\caption{The hardware accelerators that use sparsity encoding methods}
\vspace{-5pt}
\label{tab:hw-sparsity-encoding}
\resizebox{\textwidth}{!}{%
\begin{threeparttable}
\begin{tabular}{l|c|c|c|c|c|c|c}
\hline
Methods &
Technology &
Area &
Frequency &
Energy Efficiency (TOPS/W)
&
Zero Elements Ratio &
Encoding Format &
Data Format\\ \hline
\textit{Eyeriss   v2~\cite{chen2019eyeriss}} &
  65 nm &
  2695k gates (NAND-2) &
  200 MHz &
  0.96 &
  68.98\% &
  CSC &
  16b \\ \hline
\textit{Envision~\cite{moons201714}}         & 28 nm & 1.87 mm$^2$        & 200 MHz & 1.3   & 5-82\%    & Bitmap & 16b \\ \hline
\textit{Sticker~\cite{yuan2018sticker}} &
  65 nm &
  7.8 mm$^2$ &
  200 MHz &
  62.1 &
  both 5\% &
  \begin{tabular}[c]{@{}c@{}}COO for activation, CSC for weight\end{tabular} &
  8b \\ \hline
\textit{SNAP~\cite{zhang2019snap}}          & 16 nm & 2.4 mm$^2$         & 260 MHz & 21.55 & both 10\% & COO-1D & 16b \\ \hline
\end{tabular}%

\end{threeparttable}
}
\vspace{-2.5mm}
\end{table*}

In summary, some works store the pruned weights in the memory to reduce the memory footprint and bandwidth requirements.
Sparse encoding methods are introduced in the hardware accelerators for better performance and higher efficiency while introducing additional computational complexity to the decoding process.

\vspace{-5pt}
\subsection{Hardware Accelerating on Non-linear Operations}
\subsubsection{NMS}
\label{sec:accelerator_nms}
Although there are lots of algorithms that optimize NMS, hardware-accelerated NMS methods are explored less.
The CPU-NMS~\cite{rothe2014non} and GPU-NMS~\cite{gpunms, gpunms2} techniques are designed to operate on CPU and GPU platforms. GPU-NMS V2~\cite{gpunms2} can process 1027 bounding boxes in just 0.324 ms on a GeForce GTX 1060 running at 1.70 GHz. Shi \textit{et al.}\cite{powernms} were motivated by GPU-NMS\cite{gpunms} to develop a power-efficient NMS accelerator that merges 1000 bounding boxes in 12.79 microseconds at 400 MHz. Despite this, the accelerated NMS still takes considerable time compared to the execution time of convolution operations. Moreover, these solutions lack support for high-resolution images.

MaxpoolNMS~\cite{maxpoolnms} and PSRR-MaxpoolNMS~\cite{psrrmaxpool} reformulate NMS as MaxPool operation, which is inherently parallel and friendly to the hardware. 
Inspired by these two NMS algorithms, ShapoolNMS~\cite{shapoolnms} is proposed, which is a salable NMS hardware accelerator that supports high image resolution and both one- and two-stage object detectors. These works are summarized in Table~\ref{tab:alg-nms}. 
As the number of boxes increases, ShapoolNMS~\cite{shapoolnms} surpasses GreedyNMS hardware, GPU-NMS V2~\cite{gpunms2}, and Shi~\etal's~\cite{powernms} solutions in terms of performance. This is because the time complexity of GPU-NMS and Shi~\etal's method is at least $\mathcal{O}(nm)$\cite{powernms}, whereas ShapoolNMS has a time complexity of $\mathcal{O}(n)$, as outlined at Table IV in Appendix E.

\begin{table*}[!t]
\centering
\scriptsize
\caption{Comparing the NMS algorithms and hardware implementations}
\vspace{-5pt}
\label{tab:alg-nms}
\resizebox{0.93\textwidth}{!}{%
\begin{tabular}{l|c|c|c|c|c|c}
\hline
NMS Algorithm/HW &
Platform &
Hardware Friendly?
  &
High Resolution?
  &
Parallelable? &
Two-stage detectors?
  &
Time Complexity \\ \hline
GreedyNMS~\cite{greedynms} &
  CPU &
  \xmark &
  \cmark &
  \xmark &
  \cmark &
  $\mathcal{O}(n\log(n)) + \mathcal{O}(nm)$ \\ \hline
SoftNMS~\cite{bodla2017soft} &
  CPU &
  \xmark &
  \cmark &
  \xmark &
  \cmark &
  $\mathcal{O}(n\log(n)) + \mathcal{O}(nm)$ \\ \hline
MaxpoolNMS~\cite{maxpoolnms} &
  CPU &
  \cmark &
  \cmark &
  \cmark &
  \xmark &
  $\mathcal{O}(n)$ \\ \hline
PSRR-MaxpoolNMS~\cite{psrrmaxpool}&
  CPU &
  \cmark &
  \cmark &
  \cmark &
  \cmark &
  $\mathcal{O}(n)$ \\ \hline
CPU-NMS~\cite{rothe2014non} &
  CPU &
  -- &
  \xmark &
  \cmark &
  \xmark &
  $\mathcal{O}(nm)$ \\ \hline
GPU-NMS~\cite{gpunms2}&
  GPU &
  -- &
  \xmark &
  \cmark &
  \xmark &
  $\mathcal{O}(nm)$ \\ \hline
Shi~\etal~\cite{powernms} &
  ASIC &
  -- &
  \xmark &
  \cmark &
  \xmark &
  $\mathcal{O}(nm)$ \\ \hline
ShapoolNMS~\cite{shapoolnms} &
  ASIC &
  -- &
  \cmark &
  \cmark &
  \cmark &
  $\mathcal{O}(n)$ \\ \hline
\end{tabular}%
}
\vspace{-2.5mm}
\end{table*}

The structure of ShapoolNMS~\cite{shapoolnms} is depicted in Fig. 3 within Appendix E. 
According to\cite{shapoolnms}, ShapoolNMS outperformed existing state-of-the-arts by an $8.54\times$ speedup and a $42,713\times$ speedup compared to GreedyNMS.

\subsubsection{Softmax}
\label{sec:hardware_softmax}
Softmax 
involves expensive operations such as exponentiation and division, which is challenging for hardware platforms dealing with large input vectors. The exponential operations' wide range can also cause overflow problems with limited hardware resources~\cite{yuan2016efficient}. 
Unfortunately, the Softmax layer can be a bottleneck for real-time applications as other convolutional layers can be hardware accelerated.

As summarized in Table~\ref{tab:hw-softmax}, the hardware acceleration of the softmax algorithm can be classified into three categories:
\begin{itemize}
    \item[A.] \textbf{Direct optimization}: optimize the hardware implementation of exponential and division units directly. 
    \item[B.] \textbf{Mathematical transformation}: Apply mathematical transforming (\eg, logarithmic transforming) to replace the exponential operations.
    \item[C.] \textbf{Mathematical reformulation}: Reformulate the de-facto softmax equation into other hardware-friendly equations, such as replacing the exponent base $e$ to $2$, new softmax layer based on integral stochastic computing.
\end{itemize}

\begin{table*}[t!]
\centering
\caption{The summary of softmax hardware implementations, optimizations, and average mean squared error (MSE) or accuracy loss.
\vspace{-2.5mm}
}
\label{tab:hw-softmax}
\resizebox{\textwidth}{!}{%
\begin{threeparttable}
\begin{tabular}{l|c|c|c|c|c}
\hline
\multicolumn{1}{l|}{Methods}&
Softmax Equation &
\begin{tabular}[c]{@{}c@{}}Exponential Function \\ Optimization Category\tnote{a}\end{tabular} &
\begin{tabular}[c]{@{}c@{}}Division Optimization\\ Category\tnote{a}\end{tabular} &
 Avg MSE &
\begin{tabular}[c]{@{}c@{}}Accuracy\\ Loss\end{tabular} \\ \hline
Yuan~\etal~\cite{yuan2016efficient} &
  $\ln(\sigma(z_i)) = (z_i-(z)_{\max})-\ln(\sum\nolimits_{L=1}^{K}e^{z_j-(z)_{\max}})$ &
  A (LUT-EXP) &
  B (logarithmic) &
  --  &
  -- \\ \hline
Geng~\etal~\cite{geng2018hardware} &
  $\sigma(z_i) = \frac{e^{z_i}}{\sum_{j=1}^{K} e^{z_j}}$
  &
  A (LUT-EXP, LUT-PLF) &
  A (bit shift) &
  -- &
  --  \\ \hline
Li~\etal~\cite{li2018efficient} &
  -- &
  A (LUT-EXP) &
  A (optimized divider) &
  $1.5 \times 10^{-5}$ &
  --   \\ \hline
Wang~\etal~\cite{wang2018high} &
  $\sigma {(z)_j} = \exp ({x_i} - ln(\sum\limits_{j = 1}^N   {{e^{{x_j}}}} ))$ that ${e^{{y_i}}} = {2^{{y_i} \cdot {{\log}_2}e}}$ &
  C (base 2) &
  B (exp-log) &
  $8 \times 10^{-5}$ &
  --  \\ \hline
Sun~\etal~\cite{sun2018high} &
  $\sigma {(z)_j} = {e^{{z_j} - \max ( z )}}/\sum\limits_{k = 1}^K {{e^{{z_k} - \max ( z )}}}$ &
  A (GLUT-EXP) &
  A (bit shift) &
  $10^{-8}$ &
  --  \\ \hline
Hu~\etal~\cite{hu2018efficient} &
  $\sigma {(z)_j} =\exp(-h_{j}')exp({-ln\sum _{j=1}^{m}e^{-h_{\acute {j}}}})$ &
  C (Integral SC) &
  B (logarithmic) &
  0.03 &
  --  \\ \hline
Du~\etal~\cite{du2019efficient} &
  $\sigma {(z)_j} = \exp((z_i-(z)_{\max})-\ln(\sum\nolimits_{L=1}^{K}e^{z_j-(z)_{\max}}))$ &
  A (GLUT-EXP) &
  B (logarithmic) &
  --  &
  0.24\% \\ \hline
Kouretas~\etal~\cite{kouretas2020hardware} &
  $\sigma {(z)_j} = e^{z_j-(z)_{\max}}$ &
  A (LUT-EXP) &
  B (logarithmic) &
  --  &
  0\% \\ \hline
Wang~\etal~\cite{wang2019customized} &
  $\sigma {(z)_j} = \frac{N^{z_j}}{\sum\nolimits_{i = 1}^j N^{x_i}}$ &
  C (dynamic learned) &
  -- &
  -- &
  --  \\ \hline
Cardarilli~\etal~\cite{cardarilli2021pseudo} &
  $\sigma {(z)_j} =\frac{2^{x_j}}{\sum _{k=1}^{N} 2^{x_k}}$ &
  C (base 2) &
  C (approximated) &
  $10^{-4}$ &
  --  \\ \hline
Spagnolo~\etal~\cite{spagnolo2021aggressive} &
  NA &
  A (accumulated) &
  A (bit shift) &
  --  &
  0\% \\ \hline
Stevens~\etal~\cite{stevens2021softermax} &
  NA &
  C (base 2) &
  -- &
  --  &
  0.5\% \\ \hline
\end{tabular}%

\begin{tablenotes}
\scriptsize
\item[a] The optimization categories are (A) Direct optimization, (B) Mathematical transformation, and (C) Mathematical reformulation. Detailed optimization methods are explained in brackets.
\end{tablenotes}

\end{threeparttable}
}
\vspace{-2.5mm}
\end{table*}

Yuan~\cite{yuan2016efficient}, for the first time, proposed an efficient hardware implementation of softmax that uses a logarithmic transformation to eliminate division operations. Low-complexity subtractors replace complex division units. To address overflow issues, down-scale parameters are introduced in exponential units, reducing data width. 
The $e^{x_i}$ operations are approximated by creating a lookup table (LUT) that stores the discrete results with the subset of $x_i$. 
The logarithmic unit is also based on LUT implementation.
As the final results in~\cite{yuan2016efficient} were based on a transformed version of the softmax equation, 
Du~\etal~\cite{du2019efficient} proposed a corrected version that the exponential function is applied to the final result. 
The exponential function is split into multiple basics based on the calculation rule. 
Hence a group of LUTs (GLUT) was introduced to store those basics. 
Sun~\etal~\cite{sun2018high} also proposed a similar architecture that utilizes GLUT to optimize the exponential function.

In addition to approximating $e^{x_i}$ directly based on LUT (LUT-EXP), Geng~\etal~\cite{geng2018hardware} also proposed another approximation technique using a Piecewise Linear Function (LUT-PLF).  Besides, Geng~\etal proposed six softmax operation combinations, including exponential functions, power-of-twos, and look-up tables.
In~\cite{cardarilli2021pseudo}, \cite{wang2018high} and \cite{stevens2021softermax}, the exponential function was optimized by replacing base exponent with 2. 
Hu~\etal~\cite{hu2018efficient} reformulated the softmax layer using integral stochastic computing for the exponent operations and logarithmic transformation to avoid the division operations. Spagnolo~\cite{spagnolo2021aggressive} proposed approach optimizes the softmax through aggressive optimization, replacing complex processes with hardware-friendly operations. Specifically, right-shifting operations replace division, and addition operations replace exponential operations.

As Table~\ref{tab:hw-softmax-ppa} summarized, most Softmax hardware accelerators target ASICs, with a relatively small area and power overhead compared with the convolution-based or Transformer-based hardware accelerators. However, none of them optimize the Softmax with a long input sequence or a lot of classes, crucial for Transformers. This gap presents a key opportunity to innovate Softmax accelerators for Transformer models. 

\begin{table*}[t!]
\scriptsize
\centering
\caption{The summary of softmax hardware accelerators.}
\vspace{-5pt}
\label{tab:hw-softmax-ppa}
\resizebox{0.92\textwidth}{!}{%
\begin{threeparttable}
\begin{tabular}{l|c|c|c|c|c|c}
\hline
\multicolumn{1}{l|}{
Methods} &
Platform &
  Technology &
  Frequency &
  Area (LUT) &
  Power &
  Performance \\ \hline
Yuan~\etal~\cite{yuan2016efficient}\tnote{a} &
  ASIC &
  90 nm &
  250 MHz &
  4.3576 $\times 10^4 \mu\text{m}^2$ &
  3228.2 mW &
  10 classes \\ \hline
Geng~\etal~\cite{geng2018hardware} &
  ASIC &
  65 nm &
  500 MHz &
  38 $\mu\text{m}^2$ &
  0.086 mW &
  R-FCN (ResNet-101) on COCO 2014 \\ \hline
Li~\etal~\cite{li2018efficient} &
  ASIC &
  45 nm &
  3.3 GHz &
  3.4348 $\times 10^4 \mu\text{m}^2$ &
  -- &
  16-bit input \\ \hline
Wang~\etal~\cite{wang2018high} &
  ASIC &
  28 nm &
  2.8 GHz &
  1.5 $\times 10^4 \mu\text{m}^2$ &
  51.60 mW &
  22.4 G/s (16-bit) \\ \hline
Sun~\etal~\cite{sun2018high} &
  ASIC &
  65 nm &
  1 GHz &
  0.44 mm $\mu\text{m}^2$ &
  333 mW &
  16-bit valid input \\ \hline
Hu~\etal~\cite{hu2018efficient} &
  FPGA &
  -- &
  -- &
  10746 LUTs &
  603 mW &
  -- \\ \hline
Du~\etal~\cite{du2019efficient} &
  ASIC &
  65 nm &
  500 MHz &
  0.64 mm$^2$ &
  0.82 mW &
  Target on binary and ternary NN \\ \hline
Kouretas~\etal~\cite{kouretas2020hardware} &
  ASIC &
  90 nm &
  255 MHz &
  2.5597 $\times 10^4 \mu\text{m}^2$ &
  1576.3 mW &
  5 classes \\ \hline
Cardarilli~\etal~\cite{cardarilli2021pseudo} &
  ASIC &
  90 nm &
  310 MHz &
  4.6816  $\times 10^4   \mu\text{m}^2$ &
  2.769 mW &
  8-bit integer, 10 classes \\ \hline
Spagnolo~\etal~\cite{spagnolo2021aggressive} &
  ASIC &
  28 nm &
  2.5 GHz &
  3595 $\mu\text{m}^2$ &
  NA &
  16-bit integer, 10 classes \\ \hline
\end{tabular}%

\begin{tablenotes}
\item[a] The data is reported in~\cite{kouretas2020hardware}.
\end{tablenotes}
\end{threeparttable}
}
\vspace{-5mm}
\end{table*}


\vspace{-5pt}
\subsection{Stochastic Computing Architecture}
\label{subsec:stochastic_architecture}
Power-efficient approximate computing techniques, Stochastic Computing (SC), is a random, non-weighted, bitstream-based unary computing technique.  An SC circuit necessitates minimal hardware, low power consumption, and high fault tolerance to computational errors~\cite{gaines1969stochastic}. 
It has found widespread application in various network architectures, including CNNs, RNNs, and MLPs~\cite{liu2020survey}. However, due to lengthy sequences and a high demand for stochastic number generators (SNGs), achieving competitive computational latency and energy consumption is challenging for SC neural networks.
Recent advancements in stochastic computing techniques have greatly enhanced the performance of SC Neural Networks (SC NNs), bringing them to a level of competitiveness with conventional binary designs while using significantly fewer hardware resources. 
Some designs focus on improve the mostly cost random number generators (RNGs), leading to better hardware and energy efficiency~\cite{salehi2020low}. Some research focus on designing efficient hardware architecture to implement SC NNs~\cite{frasser2022fully, moran2022digital}. \cite{moran2022digital} proposed a hardware implementation on stochastic computing for Radial Basis Function (RBF) neural networks.

\section{When NN compression meets Homomorphic Encryption}\label{sec:compression_ahe}

\subsection{Motivation}
\label{sec:ahe_hnn}
Homomorphic Encryption (HE) stands as a distinctive encryption paradigm enabling computations on encrypted data (ciphertexts) without the requirement of the decryption key. The decrypted results of these computations are equivalent to outcomes derived from standard arithmetic operations performed on unencrypted data. Homomorphic Encryption can be categorized into Partial Homomorphic Encryption (PHE) and Fully Homomorphic Encryption (FHE) based on the supported homomorphic arithmetic operations. PHE supports either homomorphic addition, as introduced by Paillier~\cite{paillier1999public}, or homomorphic multiplication, as outlined by Rivest~\cite{rivest1978method}, between two ciphertexts. In contrast, FHE, a more comprehensive variant pioneered by Gentry~\cite{gentry2009fully}, accommodates both homomorphic addition and multiplication on ciphertexts. Additionally, FHE incorporates the \textit{bootstrapping} primitive, although its practical performance remains a challenge, it holds the potential to reduce noise in ciphertexts, rendering them amenable for arbitrary depth of computations.

Given the powerful features of FHE, it is a natural choice of technology for protecting data privacy during computation in untrusted environment. Notably, the concept of Homomorphic Neural Networks (HNNs) involves the utilization of FHE to encrypt both neural network models and input data. This application ensures the confidentiality of models and data throughout training or inference processes conducted in untrusted settings, such as cloud computing or multi-party collaborations. This becomes particularly crucial in domains where data privacy holds paramount importance, as evidenced in sectors like medical or financial applications~\cite{isakov2019survey}.

\begin{figure}[!ht]
    \centering
    \includegraphics[width=.46\textwidth]{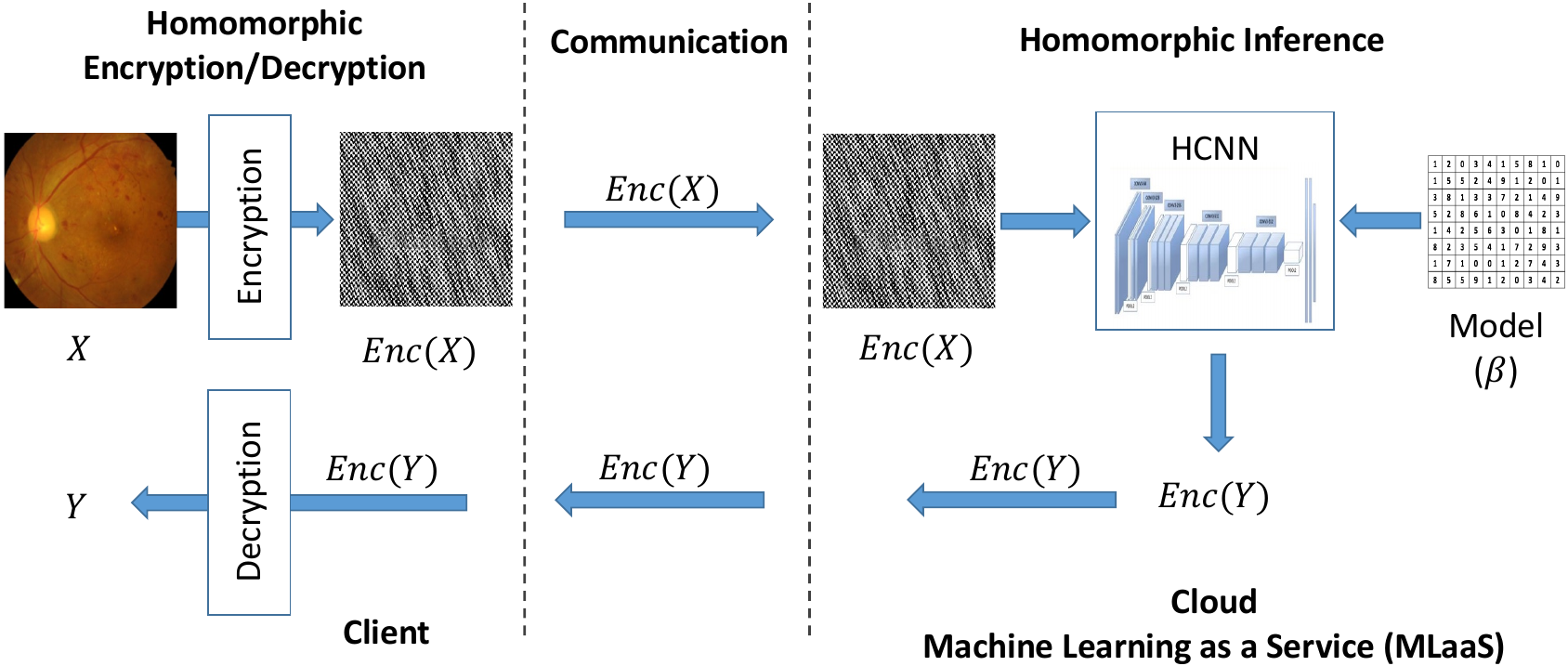}
    \caption{Homomorphic neural network for secure cloud computing in Secure Machine Learning as a Service scenario.}
    \label{fig:hcnn}
    \vspace{-5mm}
\end{figure}

Fig.~\ref{fig:hcnn} illustrates the application of Homomorphic Neural Networks (HNN) in preserving the privacy of medical images for disease diagnosis within a cloud deployment, which is also referred to as the Machine Learning as a Service scenario. The process begins with the client generating HE keys to encrypt her private images, transforming them into ciphertexts denoted as $Enc(X)$. Subsequently, these ciphertexts are transmitted to the cloud server to request disease diagnosis on the encrypted images. The cloud server then performs evaluations using the HNN-based disease diagnosis model on the image ciphertexts, yielding the encrypted diagnosis result, denoted as $Enc(Y)$. This encrypted result is relayed back to the client, who decrypts it into plaintext using her secret decryption key. Notably, throughout this entire process, the cloud server remains incapable of accessing any input data from the client or intermediate data during HNN evaluation in plaintext, as it lacks the decryption key.

However, it is non-trivial task to apply FHE to NNs to construct Homomorphic NNs, and the process faces multiple challenges. First, many mainstream FHE schemes, such as BGV~\cite{yagisawa2015fully} and BFV~\cite{fan2012somewhat} schemes, do not natively support float point arithmetic which are often found in neural network computations. Second, applying FHE to neural networks usually incur large performance and resource overheads in terms of computation and memory demands. This is attributed to the fact that plain data expands into large polynomials during homomorphic computation, and simple operations, like multiplications, translate into complex polynomial operations homomorphically~\cite{aikata2023reed}. Third, deeper networks involve deeper computations, which requires larger FHE parameters or even the expensive bootstrapping operations to control the noise growth in ciphertexts, which further increases the computationally intensity. Fortunately, neural network compression techniques can help to alleviate these challenges. This section reviews how NN compression helps on HNNs.

\subsection{NN Compression for HNNs}\label{sec:ahe_compression}
Various model compression techniques discussed in Section~\ref{sec:compression}, such as model quantization, pruning, and low-rank factorization, are commonly employed to enhance the inference speed of Homomorphic Neural Networks (HNNs). For instance, in~\cite{zhang2019encrypted}, Lloyd-max quantization was utilized to quantize weights based on their local density during training. In~\cite{lou2019she}, log-quantization was employed to transform real-valued data into power-of-two values, effectively converting matrix multiplication operations into faster shift operations to accelerate inference.
Some approaches concentrate on binarizing the network to expedite HNNs inference, as observed in works like~\cite{bourse2018fast} and~\cite{pmlr-v80-sanyal18a}. TAPAS~\cite{pmlr-v80-sanyal18a}, for instance, optimized FHE-based encrypted predictions by leveraging binary networks.
Furthermore, specific strategies adopt network pruning to diminish inference latency. For example, CryptoNets~\cite{chou2018faster} applied unstructured weight pruning to reduce the inference latency of Homomorphic neural networks, bypassing homomorphic operations related to pruned weights. Some techniques amalgamate quantization and pruning to expedite inference, as exemplified in~\cite{chou2018faster}, which proposes both pruning and quantization of models to minimize the required operations and enhance weight sparsity.
Another avenue explores low-rank factorization techniques. A recent approach presented in~\cite{lu2021ffconv} integrates low-rank factorization with FHE ciphertext packing, effectively mitigating rotation overhead and significantly accelerating the inference process.
\subsection{Hardware for HNNs}
\label{sec:ahe_accelerator}
Researchers have proposed acceleration techniques for HNNs on various hardware platforms, \textit{e.g.}, GPUs, Multi-core CPUs, FPGAs, and ASICs. Typically, hardware aids HNNs by speeding up HE operations, as demonstrated in various works~\cite{khalil2010hardware, cousins2016designing, lee2020optimizing, khalil2010hardware}. In particular, \cite{cousins2016designing} introduced an FPGA-based computation accelerator within a Homomorphic Encryption Processing Unit (HEPU) co-processor, mitigating computational bottlenecks in lattice encryption primitives to improve the practicality of computing on encrypted data. Another approach involved optimizing hardware costs, including memory usage and energy consumption, as explored in works such as~\cite{aikata2023reed, kim2023sharp, kim2022bts}. For instance, \cite{aikata2023reed} proposed a cost-effective chiplet-based FHE implementation with a non-blocking inter-chiplet communication strategy to reduce data exchange overhead. Moreover, \cite{doroz2014accelerating} achieved the first full realization of FHE in hardware, while \cite{reagen2021cheetah} presented a custom hardware accelerator architecture combining algorithmic optimizations to accelerate server-side HE inference toward plaintext speeds.
\subsection{Challenges in Compressing HNNs}
Performing compression on HNNs presents significant challenges due to the computational and operational complexity of homomorphic encryption, in which multiple factors needed to be taken into consideration, such as the homomorphic encryption schemes, and the ciphertext packing strategies. Notably, neural network quantization works well with bit-wise TFHE schemes~\cite{chillotti2020tfhe} or integer-based BFV schemes~\cite{fan2012somewhat}, but not with CKKS schemes~\cite{cheon2017homomorphic} which encodes float-point numbers. On the other hand, neuron or filter based pruning may not work well with packed weights or activations in ciphertexts. Compatibility issues with homomorphic encryption libraries and the absence of standardization create obstacles in achieving effective compression. Researchers are actively working on overcoming these challenges to make compression in HNNs more practical for real-world applications.

\section{Challenges, Future Trends and Conclusions}
\label{sec:challenges}
\noindent In this article, we have summarized recent works on compressing and accelerating DNNs.  
Next, we discuss some potential challenges and future trends.

\textbf{Generalization} While most state-of-the-art compression approaches are designed or verified on CNNs, their performance on other types of networks, especially larger and more complex models such as GPT-3 \cite{brown2020language} and LLaMA \cite{touvron2023llama} with billion parameters, and tiny models is not well established. For example, to reduce the cost of training complex models, post-quantization \cite{yao2022zeroquant, ding2022towards} has been explored to quantize the weights and activations after the model is trained.  \cite{xu2022etinynet}  further quantizes tiny models ($\le 1\text{MB}$) in aggressive low-bit while retaining the accuracy.

Besides, most latest approaches focus on 2D data, which limits their real applications to handle other data, such as 3D and time-series data. 3D models can be highly complex, with millions of polygons and texture maps. Compressing such models can be a challenging task that requires specialized algorithms and hardware. 

Furthermore, due to the existence of domain discrepancy between the model development and deployment stage, compressed models trained on historical data may suffer from performance degradation at the deployment stage. Although some existing works\cite{yang2020mobileda,ryu2022knowledge} have already attempted to simultaneously address domain shift while compressing the model with knowledge distillation, cross-domain model compression is still not well explored.




\textbf{Evaluation} Although evaluation metrics (FLOPs, Model size, Latency) have been proposed for evaluating model compression, two 
drawbacks still exist. First, no standard hardware exists to evaluate various compression techniques. Consequently, 
many studied 
focus on theoretical analysis using FLOPs and model size. However, the cost of custom hardware is also affected by its memory and computation capability. Second, 
some works use outdated or non-standard hardware for evaluation, making it difficult to perform a fair comparison when conducting experiments on different hardware.


\textbf{Hardware-software co-design}  Hardware constraints in various platforms, such as mobile and IoT chips, 
pose a significant challenge for the development of DNNs. 
Designing specific compression techniques for these platforms remains a critical bottleneck. Each platform would provide different operating conditions in energy per operation, memory-access latency, memory capacity, etc. Thus, selecting the compression mechanism should be \textit{energy aware} to maximize the gains of applying such techniques. 
Besides, fully using the existing compression approaches is critical. For example, efficiently deploying the sparse architecture of an unstructured pruning network to custom hardware remains a challenge.


\textbf{Integration} Many compression approaches, like model pruning, quantization and knowledge distillation (KD), can be integrated together as they compress the DNNs from different perspectives. For example, there is a growing trend to combine KD with quantization and pruning. 

\textbf{ Heterogeneous resource management} Considering the workload and network conditions, it is necessary to distribute DNN processing tasks across heterogeneous computational platforms, including CPU, GPU, the cloud, and smartphones. In this way, the system can remain responsive to near real-time needs. 
However, how to connect and collaborate with different devices/platforms remains unexplored. Overall, efficient heterogeneous resource management is critical for the successful deployment of DNNs on a range of platforms. Continued research in this area can help optimize the performance of DNNs while minimizing resource usage. 

Finally, our survey offers a comprehensive and unique perspective on neural network compression from three different angles: model compression, hardware accelerators, and homomorphic encryption. To the best of our knowledge, this is the first survey paper that comprehensively investigates this problem from software, hardware, and security perspectives. By exploring existing techniques and potential future improvements, we aim to encourage the broader adoption of these approaches in real-world applications. Such adoption can create significant economic and social benefits by making AI more sustainable and cost-efficient in secure environments that save memory, energy, and computation. We believe that our survey can provide valuable information for researchers looking to address critical research problems and further advance the field for the benefit of society.

\section{Acknowledgements}
\label{sec:acknowledgements}
\noindent This research is supported by the Agency for Science, Technology, and Research (A*STAR) under its Funds (AME Programmatic Funds A1892b0026, A19E3b0099, MTC Programmatic Fund M23L7b0021, NRF AME Young Individual Research Grant A2084c1067 and Career Development Fund C210812035). However, any opinions, findings, conclusions, or recommendations expressed in this material are those of the author(s) and do not reflect the views of the A*STAR.

\bibliographystyle{IEEEtran}
\bibliography{IEEEabrv, reference.bib}

\begin{thebibliography}{100}
\providecommand{\url}[1]{#1}
\csname url@samestyle\endcsname
\providecommand{\newblock}{\relax}
\providecommand{\bibinfo}[2]{#2}
\providecommand{\BIBentrySTDinterwordspacing}{\spaceskip=0pt\relax}
\providecommand{\BIBentryALTinterwordstretchfactor}{4}
\providecommand{\BIBentryALTinterwordspacing}{\spaceskip=\fontdimen2\font plus
\BIBentryALTinterwordstretchfactor\fontdimen3\font minus \fontdimen4\font\relax}
\providecommand{\BIBforeignlanguage}[2]{{%
\expandafter\ifx\csname l@#1\endcsname\relax
\typeout{** WARNING: IEEEtran.bst: No hyphenation pattern has been}%
\typeout{** loaded for the language `#1'. Using the pattern for}%
\typeout{** the default language instead.}%
\else
\language=\csname l@#1\endcsname
\fi
#2}}
\providecommand{\BIBdecl}{\relax}
\BIBdecl

\bibitem{chung2022scaling}
H.~W. Chung, L.~Hou, S.~Longpre, B.~Zoph, Y.~Tay, W.~Fedus, E.~Li, X.~Wang, M.~Dehghani, S.~Brahma \emph{et~al.}, ``Scaling instruction-finetuned language models,'' \emph{arXiv preprint arXiv:2210.11416}, 2022.

\bibitem{sze2017efficient}
V.~Sze, Y.-H. Chen, T.-J. Yang, and J.~S. Emer, ``Efficient processing of deep neural networks: A tutorial and survey,'' \emph{Proceedings of the IEEE}, 2017.

\bibitem{varghese2016challenges}
B.~Varghese, N.~Wang, S.~Barbhuiya, P.~Kilpatrick, and D.~S. Nikolopoulos, ``Challenges and opportunities in edge computing,'' in \emph{IEEE SmartCloud}, 2016.

\bibitem{deng2020model}
L.~Deng, G.~Li, S.~Han, L.~Shi, and Y.~Xie, ``Model compression and hardware acceleration for neural networks: A comprehensive survey,'' \emph{Proceedings of the IEEE}, 2020.

\bibitem{mishra2023transforming}
R.~Mishra and H.~Gupta, ``Transforming large-size to lightweight deep neural networks for iot applications,'' \emph{ACM Computing Surveys}, 2023.

\bibitem{cheng2018model}
Y.~Cheng, D.~Wang, P.~Zhou, and T.~Zhang, ``Model compression and acceleration for deep neural networks: The principles, progress, and challenges,'' \emph{IEEE Signal Processing Magazine}, 2018.

\bibitem{mishra2020survey}
R.~Mishra, H.~P. Gupta, and T.~Dutta, ``A survey on deep neural network compression: Challenges, overview, and solutions,'' \emph{arXiv preprint arXiv:2010.03954}, 2020.

\bibitem{neill2020overview}
J.~O. Neill, ``An overview of neural network compression,'' \emph{arXiv preprint arXiv:2006.03669}, 2020.

\bibitem{alqahtani2021literature}
A.~Alqahtani, X.~Xie, and M.~W. Jones, ``Literature review of deep network compression,'' in \emph{Informatics}.\hskip 1em plus 0.5em minus 0.4em\relax Multidisciplinary Digital Publishing Institute, 2021.

\bibitem{berthelier2021deep}
A.~Berthelier, T.~Chateau, S.~Duffner, C.~Garcia, and C.~Blanc, ``Deep model compression and architecture optimization for embedded systems: A survey,'' \emph{Journal of Signal Processing Systems}, 2021.

\bibitem{mittal2021survey}
S.~Mittal and S.~Umesh, ``A survey on hardware accelerators and optimization techniques for rnns,'' \emph{Journal of Systems Architecture}, 2021.

\bibitem{nan2019deep}
K.~Nan, S.~Liu, J.~Du, and H.~Liu, ``Deep model compression for mobile platforms: A survey,'' \emph{Tsinghua Science and Technology}, 2019.

\bibitem{xu2022survey}
C.~Xu and J.~McAuley, ``A survey on model compression for natural language processing,'' \emph{arXiv preprint arXiv:2202.07105}, 2022.

\bibitem{gupta2022compression}
M.~Gupta and P.~Agrawal, ``Compression of deep learning models for text: A survey,'' \emph{TKDD}, 2022.

\bibitem{guo2018survey}
Y.~Guo, ``A survey on methods and theories of quantized neural networks,'' \emph{arXiv preprint arXiv:1808.04752}, 2018.

\bibitem{gholami2021survey}
A.~Gholami, S.~Kim, Z.~Dong, Z.~Yao, M.~W. Mahoney, and K.~Keutzer, ``A survey of quantization methods for efficient neural network inference,'' \emph{arXiv preprint arXiv:2103.13630}, 2021.

\bibitem{xu2020convolutional}
S.~Xu, A.~Huang, L.~Chen, and B.~Zhang, ``Convolutional neural network pruning: A survey,'' in \emph{CCC}, 2020.

\bibitem{liu2020pruning}
J.~Liu, S.~Tripathi, U.~Kurup, and M.~Shah, ``Pruning algorithms to accelerate convolutional neural networks for edge applications: A survey,'' \emph{arXiv preprint arXiv:2005.04275}, 2020.

\bibitem{reed1993pruning}
R.~Reed, ``Pruning algorithms-a survey,'' \emph{IEEE TNNLS}, 1993.

\bibitem{elsken2019neural}
T.~Elsken, J.~H. Metzen, and F.~Hutter, ``Neural architecture search: A survey,'' \emph{JMLR}, 2019.

\bibitem{ren2021comprehensive}
P.~Ren, Y.~Xiao, X.~Chang, P.-Y. Huang, Z.~Li, X.~Chen, and X.~Wang, ``A comprehensive survey of neural architecture search: Challenges and solutions,'' \emph{CSUR}, 2021.

\bibitem{wistuba2019survey}
M.~Wistuba, A.~Rawat, and T.~Pedapati, ``A survey on neural architecture search,'' \emph{arXiv preprint arXiv:1905.01392}, 2019.

\bibitem{gou2021knowledge}
J.~Gou, B.~Yu, S.~J. Maybank, and D.~Tao, ``Knowledge distillation: A survey,'' \emph{IJCV}, 2021.

\bibitem{wang2021knowledge}
L.~Wang and K.-J. Yoon, ``Knowledge distillation and student-teacher learning for visual intelligence: A review and new outlooks,'' \emph{T-PAMI}, 2021.

\bibitem{han16iclr}
S.~Han, H.~Mao, and W.~J. Dally, ``Deep compression: Compressing deep neural networks with pruning, trained quantization and huffman coding,'' in \emph{ICLR}, 2016.

\bibitem{Zhang18eccv}
D.~Zhang, J.~Yang, D.~Ye, and G.~Hua, ``Lq-nets: learned quantization for highly accurate and compact deep neural networks,'' in \emph{ECCV}, 2018.

\bibitem{inq}
A.~Zhou, A.~Yao, Y.~Guo, L.~Xu, and Y.~Chen, ``Incremental network quantization: {T}owards lossless {cnn}s with low-precision weights,'' in \emph{arXiv preprint arXiv:1702.03044}, 2017.

\bibitem{dorefa}
S.~Zhou, Y.~Wu, Z.~Ni, X.~Zhou, H.~Wen, and Y.~Zou, ``Dorefa-net: {T}raining low bitwidth convolutional neural networks with low bitwidth gradients,'' in \emph{arXiv preprint arXiv:1606.06160}, 2016.

\bibitem{Li16nips}
F.~Li, B.~Zhang, and B.~Liu, ``Ternary weight networks,'' in \emph{NIPS Workshop}, 2016.

\bibitem{pq}
Y.~Gong, L.~Liu, M.~Yang, and L.~Bourdev, ``Compressing deep convolutional networks using vector quantization,'' in \emph{ICLR}, 2015.

\bibitem{hash}
W.~Chen, J.~T. Wilson, S.~Tyree, K.~Q. Weinberger, and Y.~Chen, ``Compressing neural networks with the hashing trick,'' in \emph{ICML}, 2015.

\bibitem{frequency}
W.~Chen, J.~Wilson, S.~Tyree, K.~Q. Weinberger, and Y.~Chen, ``Compressing convolutional neural networks in the frequency domain,'' in \emph{SIGKDD}, 2016.

\bibitem{revisit}
P.~Stock, A.~Joulin, R.~Gribonval, B.~Graham, and H.~Jegou, ``And the bit goes down: revisiting the quantization of neural networks,'' in \emph{ICLR}, 2020.

\bibitem{APoT}
Y.~Li, X.~Dong, and W.~Wang, ``Additive powers-of-two quantization: An efficient non-uniform discretization for neural networks,'' in \emph{ICLR}, 2020.

\bibitem{Faraone2018arxiv}
J.~Faraone, N.~Fraser, M.~Blott, and P.~H. Leong, ``Syq: Learning symmetric quantization for efficient deep neural networks,'' in \emph{arXiv}, 2018.

\bibitem{releq}
A.~T. Elthakeb, P.~Pilligundla, F.~Mireshghallah, A.~Yazdanbakhsh, and H.~Esmaeilzadeh, ``Releq: A reinforcement learning approach for deep quantization of neural networks,'' in \emph{NeurIPS Workshop on ML for Systems}, 2018.

\bibitem{wang19cvpr}
K.~Wang, Z.~Liu, Y.~Lin, J.~Lin, and S.~Han, ``H{A}{Q}: Hardware-aware automated quantization with mixed precision,'' in \emph{CVPR}, 2019.

\bibitem{dnas}
B.~Wu, Y.~Wang, P.~Zhang, Y.~Tian, P.~Vajda, and K.~Keutzer, ``Mixed precision quantization of convnets via differentiable neural architecture search,'' in \emph{ICLR}, 2019.

\bibitem{dq}
S.~Uhlich, L.~Mauch, F.~Cardinaux, K.~Yoshiyama, J.~A. Garcia, S.~Tiedemann, T.~Kemp, and A.~Nakamura, ``Mixed precision dnns: All you need is a good parametrization,'' \emph{arXiv preprint arXiv:1905.11452}, 2019.

\bibitem{HAWQ}
Z.~Dong, Z.~Yao, A.~Gholami, M.~W. Mahoney, and K.~Keutzer, ``Hawq: Hessian aware quantization of neural networks with mixed-precision,'' in \emph{arXiv}, 2019.

\bibitem{ALQ}
Z.~Qu, Z.~Zhou, Y.~Cheng, and L.~Thiele, ``Adaptive loss-aware quantization for multi-bit networks,'' in \emph{arXiv}, 2020.

\bibitem{post4bits}
R.~Banner, Y.~Nahshan, E.~Hoffer, and D.~Soudry, ``Post-training 4-bit quantization of convolution networks for rapid-deployment,'' \emph{arXiv preprint arXiv:1810.05723}, 2018.

\bibitem{FracBits}
L.~Yang and Q.~Jin, ``Fracbits: Mixed precision quantization via fractional bit-widths,'' \emph{arXiv preprint arXiv:2007.02017}, 2020.

\bibitem{DMBQ}
S.~Zhao, T.~Yue, and X.~Hu, ``Distribution-aware adaptive multi-bit quantization,'' in \emph{CVPR}, 2021.

\bibitem{autoq}
Q.~Lou, F.~Guo, M.~Kim, L.~Liu, and L.~Jiang, ``Autoq: Automated kernel-wise neural network quantizations,'' in \emph{ICLR}, 2020.

\bibitem{coresets}
A.~Dubey, M.~Chatterjee, and N.~Ahuja, ``Coreset-based neural network compression,'' in \emph{arXiv:1807.09810}, 2018.

\bibitem{DeepCABAC}
S.~W. et~al., ``Deepcabac: Context-adaptive binary arithmetic coding for deep neural network compression,'' in \emph{ICML Workshop}, 2019.

\bibitem{EPR}
D.~Oktay, J.~Ballé, S.~Singh, and A.~Shrivastava, ``Scalable model compression by entropy penalized reparameterization,'' in \emph{ICLR}, 2020.

\bibitem{zhe2021rate}
W.~Zhe, J.~Lin, M.~S. Aly, S.~Young, V.~Chandrasekhar, and B.~Girod, ``Rate-distortion optimized coding for efficient cnn compression,'' in \emph{DCC}, 2021.

\bibitem{khoram18iclr}
S.~Khoram and J.~Li, ``Adaptive quantization of neural networks,'' in \emph{ICLR}, 2018.

\bibitem{young2021transform}
S.~Young, Z.~Wang, D.~Taubman, and B.~Girod, ``Transform quantization for cnn compression,'' \emph{T-PAMI}, 2021.

\bibitem{dct}
Y.~Wang, C.~Xu, S.~You, D.~Tao, and C.~Xu, ``Cnnpack: packing convolutional neural networks in the frequency domain,'' in \emph{NIPS}, 2016.

\bibitem{svd}
E.~L. Denton, W.~Zaremba, J.~Bruna, Y.~LeCun, and R.~Fergus, ``Exploiting linear structure within convolutional networks for efficient evaluation,'' in \emph{NIPS}, 2014.

\bibitem{svd_p}
X.~Zhang, J.~Zou, X.~Ming, K.~He, and J.~Sun, ``Efficient and accurate approximations of nonlinear convolutional networks,'' \emph{CVPR}, 2015.

\bibitem{svd_p2}
Y.-D. Kim, E.~Park, S.~Yoo, T.~Choi, L.~Yang, and D.~Shin, ``Compression of deep convolutional neural networks for fast and low power mobile applications,'' in \emph{ICLR}, 2016.

\bibitem{svd_p3}
Y.~Li, S.~Gu, L.~V. Gool, and R.~Timofte, ``Learning filter basis for convolutional neural network compression,'' \emph{ICCV}, 2019.

\bibitem{li2021bossnas}
C.~Li, T.~Tang, G.~Wang, J.~Peng, B.~Wang, X.~Liang, and X.~Chang, ``Bossnas: Exploring hybrid cnn-transformers with block-wisely self-supervised neural architecture search,'' in \emph{ICCV}, 2021.

\bibitem{chen2021autoformer}
M.~Chen, H.~Peng, J.~Fu, and H.~Ling, ``Autoformer: Searching transformers for visual recognition,'' in \emph{ICCV}, 2021.

\bibitem{bondarenko2021understanding}
Y.~Bondarenko, M.~Nagel, and T.~Blankevoort, ``Understanding and overcoming the challenges of efficient transformer quantization,'' in \emph{EMNLP}, 2021.

\bibitem{dettmers2022gpt3}
T.~Dettmers, M.~Lewis, Y.~Belkada, and L.~Zettlemoyer, ``Gpt3. int8 (): 8-bit matrix multiplication for transformers at scale,'' \emph{BIPS}, 2022.

\bibitem{liu2021post}
Z.~Liu, Y.~Wang, K.~Han, W.~Zhang, S.~Ma, and W.~Gao, ``Post-training quantization for vision transformer,'' \emph{NIPS}, 2021.

\bibitem{ding2022towards}
Y.~Ding, H.~Qin, Q.~Yan, Z.~Chai, J.~Liu, X.~Wei, and X.~Liu, ``Towards accurate post-training quantization for vision transformer,'' in \emph{ACM MM}, 2022.

\bibitem{taubmanjpeg2000}
D.~S. Taubman and M.~W. Marcellin, ``Jpeg2000 image compression fundamentals, standards and practice,'' \emph{Journal of Electronic Imaging}, 2002.

\bibitem{Tunstall1067phd}
B.~P. Tunstall, ``Synthesis of noiseless compression codes,'' in \emph{PhD diss., Georgia Institute of Technology}, 1967.

\bibitem{lee2018snip}
N.~Lee, T.~Ajanthan, and P.~H. Torr, ``Snip: Single-shot network pruning based on connection sensitivity,'' in \emph{ICLR}, 2018.

\bibitem{Wang2020Picking}
C.~Wang, G.~Zhang, and R.~Grosse, ``Picking winning tickets before training by preserving gradient flow,'' in \emph{ICLR}, 2020.

\bibitem{jorge2021progressive}
P.~de~Jorge, A.~Sanyal, H.~Behl, P.~Torr, G.~Rogez, and P.~K. Dokania, ``Progressive skeletonization: Trimming more fat from a network at initialization,'' in \emph{ICLR}, 2021.

\bibitem{ijcai2022p786}
H.~Wang, C.~Qin, Y.~Bai, Y.~Zhang, and Y.~Fu, ``Recent advances on neural network pruning at initialization,'' in \emph{IJCAI}, 2022.

\bibitem{NIPS1989_250}
Y.~LeCun, J.~S. Denker, and S.~A. Solla, ``Optimal brain damage,'' in \emph{NIPS}, 1990.

\bibitem{woodfisher}
S.~P. Singh and D.~Alistarh, ``Woodfisher: Efficient second-order approximation for neural network compression,'' in \emph{NIPS}, 2020.

\bibitem{mfac}
E.~Frantar, E.~Kurtic, and D.~Alistarh, ``M-{FAC}: Efficient matrix-free approximations of second-order information,'' in \emph{NIPS}, 2021.

\bibitem{shen2022structural}
M.~Shen, H.~Yin, P.~Molchanov, L.~Mao, J.~Liu, and J.~M. Alvarez, ``Structural pruning via latency-saliency knapsack,'' \emph{NIPS}, 2022.

\bibitem{molchanov2019importance}
P.~Molchanov, A.~Mallya, S.~Tyree, I.~Frosio, and J.~Kautz, ``Importance estimation for neural network pruning,'' in \emph{CVPR}, 2019.

\bibitem{xu2023efficient}
K.~Xu, Z.~Wang, X.~Geng, M.~Wu, X.~Li, and W.~Lin, ``Efficient joint optimization of layer-adaptive weight pruning in deep neural networks,'' in \emph{ICCV}, 2023.

\bibitem{Lin2020Dynamic}
T.~Lin, S.~U. Stich, L.~Barba, D.~Dmitriev, and M.~Jaggi, ``Dynamic model pruning with feedback,'' in \emph{ICLR}, 2020.

\bibitem{LIU2020Dynamic}
J.~Liu, Z.~XU, R.~SHI, R.~C.~C. Cheung, and H.~K. So, ``Dynamic sparse training: Find efficient sparse network from scratch with trainable masked layers,'' in \emph{ICLR}, 2020.

\bibitem{molchanov2016pruning}
P.~Molchanov, S.~Tyree, T.~Karras, T.~Aila, and J.~Kautz, ``Pruning convolutional neural networks for resource efficient inference,'' in \emph{ICLR}, 2017.

\bibitem{tang2020scop}
Y.~Tang, Y.~Wang, Y.~Xu, D.~Tao, C.~Xu, C.~Xu, and C.~Xu, ``Scop: Scientific control for reliable neural network pruning,'' \emph{NIPS}, 2020.

\bibitem{pmlr-v119-kusupati20a}
A.~Kusupati, V.~Ramanujan, R.~Somani, M.~Wortsman, P.~Jain, S.~Kakade, and A.~Farhadi, ``Soft threshold weight reparameterization for learnable sparsity,'' in \emph{ICML}, 2020.

\bibitem{ContinuousSparsification2020}
P.~Savarese, H.~Silva, and M.~Maire, ``Winning the lottery with continuous sparsification,'' in \emph{NIPS}, 2020.

\bibitem{wang2021neural}
H.~Wang, C.~Qin, Y.~Zhang, and Y.~Fu, ``Neural pruning via growing regularization,'' in \emph{ICLR}, 2021.

\bibitem{pmlr-v119-evci20a}
U.~Evci, T.~Gale, J.~Menick, P.~S. Castro, and E.~Elsen, ``Rigging the lottery: Making all tickets winners,'' in \emph{ICML}, 2020.

\bibitem{Park2020LookaheadAF}
S.~Park, J.~Lee, S.~Mo, and J.~Shin, ``Lookahead: a far-sighted alternative of magnitude-based pruning,'' in \emph{ICLR}, 2020.

\bibitem{Zhu2018ToPO}
M.~Zhu and S.~Gupta, ``To prune, or not to prune: exploring the efficacy of pruning for model compression,'' in \emph{ICLR Workshop}, vol. abs/1710.01878, 2018.

\bibitem{lee2021layeradaptive}
J.~Lee, S.~Park, S.~Mo, S.~Ahn, and J.~Shin, ``Layer-adaptive sparsity for the magnitude-based pruning,'' in \emph{ICLR}, 2021.

\bibitem{globalmp}
M.~Gupta, E.~Camci, V.~R. Keneta, A.~Vaidyanathan, R.~Kanodia, C.-S. Foo, W.~Min, and L.~Jie, ``Is complexity required for neural network pruning? a case study on global magnitude pruning,'' \emph{arXiv preprint arXiv:2209.14624}, 2022.

\bibitem{ICML-2019-MostafaW}
H.~Mostafa and X.~Wang, ``{Parameter efficient training of deep convolutional neural networks by dynamic sparse reparameterization},'' in \emph{{ICML}}, 2019.

\bibitem{sparse_momentum}
T.~Dettmers and L.~Zettlemoyer, ``Sparse networks from scratch: Faster training without losing performance,'' \emph{CoRR}, vol. abs/1907.04840, 2019.

\bibitem{gupta2020learning}
M.~Gupta, S.~Aravindan, A.~Kalisz, V.~Chandrasekhar, and L.~Jie, ``Learning to prune deep neural networks via reinforcement learning,'' \emph{ICML AutoML Workshop}, 2020.

\bibitem{he2018amc}
{He, Yihui}, J.~Lin, Z.~Liu, H.~Wang, L.-J. Li, and S.~Han, ``Amc: Automl for model compression and acceleration on mobile devices,'' in \emph{ECCV}, 2018.

\bibitem{NIPS2017_6813}
J.~Lin, Y.~Rao, J.~Lu, and J.~Zhou, ``Runtime neural pruning,'' in \emph{NIPS}, 2017.

\bibitem{frankle2018the}
J.~Frankle and M.~Carbin, ``The lottery ticket hypothesis: Finding sparse, trainable neural networks,'' in \emph{ICLR}, 2019.

\bibitem{banner2019post}
R.~Banner, Y.~Nahshan, and D.~Soudry, ``Post training 4-bit quantization of convolutional networks for rapid-deployment,'' \emph{NIPS}, 2019.

\bibitem{frantar2022optimal}
E.~Frantar and D.~Alistarh, ``Optimal brain compression: A framework for accurate post-training quantization and pruning,'' \emph{NIPS}, 2022.

\bibitem{kwon2022fast}
W.~Kwon, S.~Kim, M.~W. Mahoney, J.~Hassoun, K.~Keutzer, and A.~Gholami, ``A fast post-training pruning framework for transformers,'' \emph{NIPS}, 2022.

\bibitem{DNW}
M.~Wortsman, A.~Farhadi, and M.~Rastegari, ``Discovering neural wirings,'' in \emph{NIPS}, 2019.

\bibitem{8953212}
Y.~He, P.~Liu, Z.~Wang, Z.~Hu, and Y.~Yang, ``Filter pruning via geometric median for deep convolutional neural networks acceleration,'' in \emph{CVPR}, 2019.

\bibitem{lin2020hrank}
M.~Lin, R.~Ji, Y.~Wang, Y.~Zhang, B.~Zhang, Y.~Tian, and L.~Shao, ``Hrank: Filter pruning using high-rank feature map,'' in \emph{CVPR}, 2020.

\bibitem{Wang2020PruningFS}
Y.~Wang, X.~Zhang, L.~Xie, J.~Zhou, H.~Su, B.~Zhang, and X.~Hu, ``Pruning from scratch,'' in \emph{AAAI}, 2020.

\bibitem{10.1007/978-3-030-58598-3_36}
Y.~Li, S.~Gu, K.~Zhang, L.~Van~Gool, and R.~Timofte, ``Dhp: Differentiable meta pruning via hypernetworks,'' in \emph{ECCV}, 2020.

\bibitem{8953628}
S.~Lin, R.~Ji, C.~Yan, B.~Zhang, L.~Cao, Q.~Ye, F.~Huang, and D.~Doermann, ``Towards optimal structured cnn pruning via generative adversarial learning,'' in \emph{CVPR}, 2019.

\bibitem{CHL00L21}
W.~Wang, M.~Chen, S.~Zhao, L.~Chen, J.~Hu, H.~Liu, D.~Cai, X.~He, and W.~Liu, ``Accelerate cnns from three dimensions: A comprehensive pruning framework,'' in \emph{ICML}, 2021.

\bibitem{lagunas2021block}
F.~Lagunas, E.~Charlaix, V.~Sanh, and A.~M. Rush, ``Block pruning for faster transformers,'' in \emph{EMNLP}, 2021.

\bibitem{zhou2021learning}
A.~Zhou, Y.~Ma, J.~Zhu, J.~Liu, Z.~Zhang, K.~Yuan, W.~Sun, and H.~Li, ``Learning n: m fine-grained structured sparse neural networks from scratch,'' \emph{arXiv preprint arXiv:2102.04010}, 2021.

\bibitem{fang2022algorithm}
C.~Fang, A.~Zhou, and Z.~Wang, ``An algorithm--hardware co-optimized framework for accelerating n: M sparse transformers,'' \emph{IEEE T-VLSI}, 2022.

\bibitem{yang2023global}
H.~Yang, H.~Yin, M.~Shen, P.~Molchanov, H.~Li, and J.~Kautz, ``Global vision transformer pruning with hessian-aware saliency,'' in \emph{CVPR}, 2023.

\bibitem{yu2023boost}
C.~Yu, T.~Chen, Z.~Gan, and J.~Fan, ``Boost vision transformer with gpu-friendly sparsity and quantization,'' in \emph{CVPR}, 2023.

\bibitem{cheng2023survey}
H.~Cheng, M.~Zhang, and J.~Q. Shi, ``A survey on deep neural network pruning-taxonomy, comparison, analysis, and recommendations,'' \emph{arXiv preprint arXiv:2308.06767}, 2023.

\bibitem{yu2018slimmable}
J.~Yu, L.~Yang, N.~Xu, J.~Yang, and T.~Huang, ``Slimmable neural networks,'' in \emph{ICLR}, 2018.

\bibitem{yu2019universally}
J.~Yu and T.~S. Huang, ``Universally slimmable networks and improved training techniques,'' in \emph{ICCV}, 2019.

\bibitem{li2021dynamic}
C.~Li, G.~Wang, B.~Wang, X.~Liang, Z.~Li, and X.~Chang, ``Dynamic slimmable network,'' in \emph{CVPR}, 2021.

\bibitem{cai2019once}
H.~Cai, C.~Gan, T.~Wang, Z.~Zhang, and S.~Han, ``Once-for-all: Train one network and specialize it for efficient deployment,'' \emph{ICLR}, 2020.

\bibitem{wang2017model}
C.~Wang, X.~Lan, and Y.~Zhang, ``Model distillation with knowledge transfer from face classification to alignment and verification,'' \emph{arXiv preprint arXiv:1709.02929}, 2017.

\bibitem{lee2018teacher}
H.~J. Lee, W.~J. Baddar, H.~G. Kim, S.~T. Kim, and Y.~M. Ro, ``Teacher and student joint learning for compact facial landmark detection network,'' in \emph{MMM}, 2018.

\bibitem{zhu2021student}
Y.~Zhu and Y.~Wang, ``Student customized knowledge distillation: Bridging the gap between student and teacher,'' in \emph{ICCV}, 2021.

\bibitem{park2021learning}
D.~Y. Park, M.-H. Cha, D.~Kim, B.~Han \emph{et~al.}, ``Learning student-friendly teacher networks for knowledge distillation,'' in \emph{NIPS}, 2021.

\bibitem{kim2016sequence}
Y.~Kim and A.~M. Rush, ``Sequence-level knowledge distillation,'' \emph{arXiv preprint arXiv:1606.07947}, 2016.

\bibitem{gordon2019explaining}
M.~A. Gordon and K.~Duh, ``Explaining sequence-level knowledge distillation as data-augmentation for neural machine translation,'' \emph{arXiv preprint arXiv:1912.03334}, 2019.

\bibitem{wang2021joint}
Z.-R. Wang and J.~Du, ``Joint architecture and knowledge distillation in cnn for chinese text recognition,'' \emph{Pattern Recognition}, 2021.

\bibitem{chebotar2016distilling}
Y.~Chebotar and A.~Waters, ``Distilling knowledge from ensembles of neural networks for speech recognition.'' in \emph{Interspeech}, 2016.

\bibitem{kwon2020adaptive}
K.~Kwon, H.~Na, H.~Lee, and N.~S. Kim, ``Adaptive knowledge distillation based on entropy,'' in \emph{ICASSP}, 2020.

\bibitem{li2021mutual}
Z.~Li, Y.~Ming, L.~Yang, and J.-H. Xue, ``Mutual-learning sequence-level knowledge distillation for automatic speech recognition,'' \emph{Neurocomputing}, 2021.

\bibitem{xu2021kdnet}
Q.~Xu, Z.~Chen, K.~Wu, C.~Wang, M.~Wu, and X.~Li, ``Kdnet-rul: A knowledge distillation framework to compress deep neural networks for machine remaining useful life prediction,'' \emph{IEEE Transactions on Industrial Electronics}, 2021.

\bibitem{xu2022contrastive}
Q.~Xu, Z.~Chen, M.~Ragab, C.~Wang, M.~Wu, and X.~Li, ``Contrastive adversarial knowledge distillation for deep model compression in time-series regression tasks,'' \emph{Neurocomputing}, 2022.

\bibitem{hinton2015distilling}
G.~Hinton, O.~Vinyals, J.~Dean \emph{et~al.}, ``Distilling the knowledge in a neural network,'' \emph{arXiv preprint arXiv:1503.02531}, 2015.

\bibitem{zhang2018better}
C.~Zhang and Y.~Peng, ``Better and faster: knowledge transfer from multiple self-supervised learning tasks via graph distillation for video classification,'' in \emph{IJCAI}, 2018.

\bibitem{Zhao_2022_CVPR}
B.~Zhao, Q.~Cui, R.~Song, Y.~Qiu, and J.~Liang, ``Decoupled knowledge distillation,'' in \emph{CVPR}, 2022.

\bibitem{chen2017learning}
G.~Chen, W.~Choi, X.~Yu, T.~Han, and M.~Chandraker, ``Learning efficient object detection models with knowledge distillation,'' in \emph{NIPS}, 2017.

\bibitem{yuan2019obtain}
C.~Yuan and R.~Pan, ``Obtain dark knowledge via extended knowledge distillation,'' in \emph{AIAM}, 2019.

\bibitem{hegde2020variational}
S.~Hegde, R.~Prasad, R.~Hebbalaguppe, and V.~Kumar, ``Variational student: Learning compact and sparser networks in knowledge distillation framework,'' in \emph{ICASSP}, 2020.

\bibitem{saputra2019distilling}
M.~R.~U. Saputra, P.~P. De~Gusmao, Y.~Almalioglu, A.~Markham, and N.~Trigoni, ``Distilling knowledge from a deep pose regressor network,'' in \emph{ICCV}, 2019.

\bibitem{zagoruyko2016paying}
S.~Zagoruyko and N.~Komodakis, ``Paying more attention to attention: Improving the performance of convolutional neural networks via attention transfer,'' in \emph{ICLR}, 2017.

\bibitem{yim2017gift}
J.~Yim, D.~Joo, J.~Bae, and J.~Kim, ``A gift from knowledge distillation: Fast optimization, network minimization and transfer learning,'' in \emph{CVPR}, 2017.

\bibitem{kim2018paraphrasing}
J.~Kim, S.~Park, and N.~Kwak, ``Paraphrasing complex network: Network compression via factor transfer,'' in \emph{NIPS}, 2018.

\bibitem{ahn2019variational}
S.~Ahn, S.~X. Hu, A.~Damianou, N.~D. Lawrence, and Z.~Dai, ``Variational information distillation for knowledge transfer,'' in \emph{CVPR}, 2019.

\bibitem{liu2021exploring}
L.~Liu, Q.~Huang, S.~Lin, H.~Xie, B.~Wang, X.~Chang, and X.~Liang, ``Exploring inter-channel correlation for diversity-preserved knowledge distillation,'' in \emph{ICCV}, 2021.

\bibitem{lin2022knowledge}
S.~Lin, H.~Xie, B.~Wang, K.~Yu, X.~Chang, X.~Liang, and G.~Wang, ``Knowledge distillation via the target-aware transformer,'' in \emph{CVPR}, 2022.

\bibitem{shang2021lipschitz}
Y.~Shang, B.~Duan, Z.~Zong, L.~Nie, and Y.~Yan, ``Lipschitz continuity guided knowledge distillation,'' in \emph{ICCV}, 2021.

\bibitem{huang2021revisiting}
Z.~Huang, X.~Shen, J.~Xing, T.~Liu, X.~Tian, H.~Li, B.~Deng, J.~Huang, and X.-S. Hua, ``Revisiting knowledge distillation: An inheritance and exploration framework,'' in \emph{CVPR}, 2021.

\bibitem{ji2021show}
M.~Ji, B.~Heo, and S.~Park, ``Show, attend and distill: Knowledge distillation via attention-based feature matching,'' in \emph{AAAI}, 2021.

\bibitem{passalis2018learning}
N.~Passalis and A.~Tefas, ``Learning deep representations with probabilistic knowledge transfer,'' in \emph{ECCV}, 2018.

\bibitem{park2019relational}
W.~Park, D.~Kim, Y.~Lu, and M.~Cho, ``Relational knowledge distillation,'' in \emph{CVPR}, 2019.

\bibitem{liu2019knowledge}
Y.~Liu, J.~Cao, B.~Li, C.~Yuan, W.~Hu, Y.~Li, and Y.~Duan, ``Knowledge distillation via instance relationship graph,'' in \emph{CVPR}, 2019.

\bibitem{huang2022knowledge}
T.~Huang, S.~You, F.~Wang, C.~Qian, and C.~Xu, ``Knowledge distillation from a stronger teacher,'' in \emph{NIPS}, 2022.

\bibitem{yun2020regularizing}
S.~Yun, J.~Park, K.~Lee, and J.~Shin, ``Regularizing class-wise predictions via self-knowledge distillation,'' in \emph{CVPR}, 2020.

\bibitem{tian2019contrastive}
Y.~Tian, D.~Krishnan, and P.~Isola, ``Contrastive representation distillation,'' in \emph{ICLR}, 2020.

\bibitem{zhu2021complementary}
J.~Zhu, S.~Tang, D.~Chen, S.~Yu, Y.~Liu, M.~Rong, A.~Yang, and X.~Wang, ``Complementary relation contrastive distillation,'' in \emph{CVPR}, 2021.

\bibitem{peng2019correlation}
B.~Peng, X.~Jin, J.~Liu, D.~Li, Y.~Wu, Y.~Liu, S.~Zhou, and Z.~Zhang, ``Correlation congruence for knowledge distillation,'' in \emph{Proceedings of the IEEE/CVF International Conference on Computer Vision}, 2019, pp. 5007--5016.

\bibitem{geng2018hardware}
X.~Geng, J.~Lin, B.~Zhao, A.~Kong, M.~M.~S. Aly, and V.~Chandrasekhar, ``Hardware-aware softmax approximation for deep neural networks,'' in \emph{ACCV}, 2018.

\bibitem{greedynms}
N.~Dalal and B.~Triggs, ``Histograms of oriented gradients for human detection,'' in \emph{CVPR}, 2005.

\bibitem{bodla2017soft}
N.~Bodla, B.~Singh, R.~Chellappa, and L.~S. Davis, ``Soft-nms--improving object detection with one line of code,'' in \emph{ICCV}, 2017.

\bibitem{tychsen2018improving}
L.~Tychsen-Smith and L.~Petersson, ``Improving object localization with fitness nms and bounded iou loss,'' in \emph{CVPR}, 2018.

\bibitem{jiang2018acquisition}
B.~Jiang, R.~Luo, J.~Mao, T.~Xiao, and Y.~Jiang, ``Acquisition of localization confidence for accurate object detection,'' in \emph{ECCV}, 2018.

\bibitem{gahlert2020visibility}
N.~G{\"a}hlert, N.~Hanselmann, U.~Franke, and J.~Denzler, ``Visibility guided nms: Efficient boosting of amodal object detection in crowded traffic scenes,'' \emph{arXiv preprint arXiv:2006.08547}, 2020.

\bibitem{he2019bounding}
Y.~He, C.~Zhu, J.~Wang, M.~Savvides, and X.~Zhang, ``Bounding box regression with uncertainty for accurate object detection,'' in \emph{CVPR}, 2019.

\bibitem{liu2019adaptive}
S.~Liu, D.~Huang, and Y.~Wang, ``Adaptive nms: Refining pedestrian detection in a crowd,'' in \emph{CVPR}, 2019.

\bibitem{salscheider2021featurenms}
N.~O. Salscheider, ``Featurenms: Non-maximum suppression by learning feature embeddings,'' in \emph{ICPR}, 2021.

\bibitem{hosang2016convnet}
J.~Hosang, R.~Benenson, and B.~Schiele, ``A convnet for non-maximum suppression,'' in \emph{German Conference on Pattern Recognition}, 2016.

\bibitem{hosang2017learning}
{J. Hosang}, {R. Benenson}, and {B. Schiele}, ``Learning non-maximum suppression,'' in \emph{CVPR}, 2017.

\bibitem{hu2018relation}
H.~Hu, J.~Gu, Z.~Zhang, J.~Dai, and Y.~Wei, ``Relation networks for object detection,'' in \emph{CVPR}, 2018.

\bibitem{rothe2014non}
R.~Rothe, M.~Guillaumin, and L.~V. Gool, ``Non-maximum suppression for object detection by passing messages between windows,'' in \emph{ACCV}, 2014.

\bibitem{gpunms2}
D.~Oro \emph{et~al.}, ``{Work-Efficient Parallel Non-Maximum Suppression Kernels},'' \emph{The Computer Journal}, 08 2020.

\bibitem{shapoolnms}
C.~Chen, T.~Zhang, Z.~Yu, A.~Raghuraman, S.~Udayan, J.~Lin, and M.~M.~S. Aly, ``Scalable hardware acceleration of non-maximum suppression,'' in \emph{DATE}, 2022.

\bibitem{maxpoolnms}
L.~Cai \emph{et~al.}, ``Maxpoolnms: getting rid of nms bottlenecks in two-stage object detectors,'' in \emph{CVPR}, 2019.

\bibitem{psrrmaxpool}
T.~Zhang \emph{et~al.}, ``Psrr-maxpoolnms: Pyramid shifted maxpoolnms with relationship recovery,'' in \emph{CVPR}, 2021.

\bibitem{joulin2017efficient}
A.~Joulin, M.~Ciss{\'e}, D.~Grangier, H.~J{\'e}gou \emph{et~al.}, ``Efficient softmax approximation for gpus,'' in \emph{ICML}, 2017.

\bibitem{shim2017svd}
K.~Shim, M.~Lee, I.~Choi, Y.~Boo, and W.~Sung, ``Svd-softmax: Fast softmax approximation on large vocabulary neural networks,'' in \emph{NIPS}, 2017.

\bibitem{blanc2018adaptive}
G.~Blanc and S.~Rendle, ``Adaptive sampled softmax with kernel based sampling,'' in \emph{ICML}, 2018.

\bibitem{lu2021soft}
J.~Lu, J.~Yao, J.~Zhang, X.~Zhu, H.~Xu, W.~Gao, C.~Xu, T.~Xiang, and L.~Zhang, ``Soft: Softmax-free transformer with linear complexity,'' in \emph{NIPS}, 2021.

\bibitem{zhao2021ann}
K.~Zhao, L.~Song, Y.~Zhang, P.~Pan, Y.~Xu, and R.~Jin, ``Ann softmax: acceleration of extreme classification training,'' \emph{Proceedings of the VLDB Endowment}, 2021.

\bibitem{chen2018learning}
P.~H. Chen, S.~Si, S.~Kumar, Y.~Li, and C.-J. Hsieh, ``Learning to screen for fast softmax inference on large vocabulary neural networks,'' \emph{arXiv preprint arXiv:1810.12406}, 2018.

\bibitem{liao2019doubly}
S.~Liao, T.~Chen, T.~Lin, D.~Zhou, and C.~Wang, ``Doubly sparse: Sparse mixture of sparse experts for efficient softmax inference,'' \emph{arXiv preprint arXiv:1901.10668}, 2019.

\bibitem{rawat2019sampled}
A.~S. Rawat, J.~Chen, F.~X.~X. Yu, A.~T. Suresh, and S.~Kumar, ``Sampled softmax with random fourier features,'' in \emph{NIPS}, 2019.

\bibitem{andreas2015accuracy}
J.~Andreas, M.~Rabinovich, M.~I. Jordan, and D.~Klein, ``On the accuracy of self-normalized log-linear models,'' \emph{NIPS}, 2015.

\bibitem{krizhevsky2012imagenet}
A.~Krizhevsky, I.~Sutskever, and G.~E. Hinton, ``Imagenet classification with deep convolutional neural networks,'' \emph{NIPS}, 2012.

\bibitem{he2015delving}
K.~He, X.~Zhang, S.~Ren, and J.~Sun, ``Delving deep into rectifiers: Surpassing human-level performance on imagenet classification,'' in \emph{ICCV}, 2015.

\bibitem{tan2019efficientnet}
M.~Tan and Q.~Le, ``Efficientnet: Rethinking model scaling for convolutional neural networks,'' in \emph{ICML}, 2019.

\bibitem{cai2018proxylessnas}
H.~Cai, L.~Zhu, and S.~Han, ``Proxylessnas: Direct neural architecture search on target task and hardware,'' \emph{arXiv preprint arXiv:1812.00332}, 2018.

\bibitem{tan2019mnasnet}
M.~Tan, B.~Chen, R.~Pang, V.~Vasudevan, M.~Sandler, A.~Howard, and Q.~V. Le, ``Mnasnet: Platform-aware neural architecture search for mobile,'' in \emph{CVPR}, 2019.

\bibitem{wu2019fbnet}
B.~Wu, X.~Dai, P.~Zhang, Y.~Wang, F.~Sun, Y.~Wu, Y.~Tian, P.~Vajda, Y.~Jia, and K.~Keutzer, ``Fbnet: Hardware-aware efficient convnet design via differentiable neural architecture search,'' in \emph{CVPR}, 2019.

\bibitem{liu2018darts}
H.~Liu, K.~Simonyan, and Y.~Yang, ``Darts: Differentiable architecture search,'' \emph{ICLR}, 2019.

\bibitem{pham2018efficient}
H.~Pham, M.~Guan, B.~Zoph, Q.~Le, and J.~Dean, ``Efficient neural architecture search via parameters sharing,'' in \emph{ICML}, 2018.

\bibitem{ancilotto2023xinet}
A.~Ancilotto, F.~Paissan, and E.~Farella, ``Xinet: Efficient neural networks for tinyml,'' in \emph{ICCV}, 2023.

\bibitem{lin2021mcunetv2}
J.~Lin, W.-M. Chen, H.~Cai, C.~Gan, and S.~Han, ``Mcunetv2: Memory-efficient patch-based inference for tiny deep learning,'' \emph{arXiv preprint arXiv:2110.15352}, 2021.

\bibitem{cai2020tinytl}
H.~Cai, C.~Gan, L.~Zhu, and S.~Han, ``Tinytl: Reduce memory, not parameters for efficient on-device learning,'' \emph{NIPS}, 2020.

\bibitem{chen2014diannao}
T.~Chen, Z.~Du, N.~Sun, J.~Wang, C.~Wu, Y.~Chen, and O.~Temam, ``Diannao: A small-footprint high-throughput accelerator for ubiquitous machine-learning,'' \emph{ACM SIGARCH Computer Architecture News}, 2014.

\bibitem{chen2014dadiannao}
Y.~Chen, T.~Luo, S.~Liu, S.~Zhang, L.~He, J.~Wang, L.~Li, T.~Chen, Z.~Xu, N.~Sun \emph{et~al.}, ``Dadiannao: A machine-learning supercomputer,'' in \emph{MICRO}, 2014.

\bibitem{liu2015pudiannao}
D.~Liu, T.~Chen, S.~Liu, J.~Zhou, S.~Zhou, O.~Teman, X.~Feng, X.~Zhou, and Y.~Chen, ``Pudiannao: A polyvalent machine learning accelerator,'' \emph{ACM SIGARCH Computer Architecture News}, 2015.

\bibitem{nvdla}
\BIBentryALTinterwordspacing
{Nvidia}. (2018) {NVIDIA Deep Learning Accelerator}. [Online]. Available: \url{http://nvdla.org/primer.html}
\BIBentrySTDinterwordspacing

\bibitem{kwon2018maeri}
H.~Kwon, A.~Samajdar, and T.~Krishna, ``Maeri: Enabling flexible dataflow mapping over dnn accelerators via reconfigurable interconnects,'' \emph{ACM SIGPLAN Notices}, 2018.

\bibitem{strassen1969gaussian}
V.~Strassen, ``Gaussian elimination is not optimal,'' \emph{Numerische mathematik}, 1969.

\bibitem{winograd1980arithmetic}
S.~Winograd, \emph{Arithmetic complexity of computations}.\hskip 1em plus 0.5em minus 0.4em\relax Siam, 1980.

\bibitem{brigham1967fast}
E.~O. Brigham and R.~Morrow, ``The fast fourier transform,'' \emph{IEEE spectrum}, 1967.

\bibitem{cong2014minimizing}
J.~Cong and B.~Xiao, ``Minimizing computation in convolutional neural networks,'' in \emph{ICANN}, 2014.

\bibitem{lavin2016fast}
A.~Lavin and S.~Gray, ``Fast algorithms for convolutional neural networks,'' in \emph{CVPR}, 2016.

\bibitem{ben1997fast}
S.~Ben-Yacoub, ``Fast object detection using mlp and fft,'' IDIAP, Tech. Rep., 1997.

\bibitem{mathieu2013fast}
M.~Mathieu, M.~Henaff, and Y.~LeCun, ``Fast training of convolutional networks through ffts,'' \emph{arXiv preprint arXiv:1312.5851}, 2013.

\bibitem{vasilache2014fast}
N.~Vasilache, J.~Johnson, M.~Mathieu, S.~Chintala, S.~Piantino, and Y.~LeCun, ``Fast convolutional nets with fbfft: A gpu performance evaluation,'' \emph{arXiv preprint arXiv:1412.7580}, 2014.

\bibitem{liang2019evaluating}
Y.~Liang, L.~Lu, Q.~Xiao, and S.~Yan, ``Evaluating fast algorithms for convolutional neural networks on fpgas,'' \emph{IEEE TCAD}, 2019.

\bibitem{lu2020hardware}
S.~Lu, M.~Wang, S.~Liang, J.~Lin, and Z.~Wang, ``Hardware accelerator for multi-head attention and position-wise feed-forward in the transformer,'' in \emph{SOCC}, 2020.

\bibitem{ham20203}
T.~J. Ham, S.~J. Jung, S.~Kim, Y.~H. Oh, Y.~Park, Y.~Song, J.-H. Park, S.~Lee, K.~Park, J.~W. Lee \emph{et~al.}, ``A\^{} 3: Accelerating attention mechanisms in neural networks with approximation,'' in \emph{HPCA}, 2020.

\bibitem{hu2021vis}
W.~Hu, D.~Xu, Z.~Fan, F.~Liu, and Y.~He, ``Vis-top: Visual transformer overlay processor,'' \emph{arXiv preprint arXiv:2110.10957}, 2021.

\bibitem{sun2022vaqf}
M.~Sun, H.~Ma, G.~Kang, Y.~Jiang, T.~Chen, X.~Ma, Z.~Wang, and Y.~Wang, ``Vaqf: Fully automatic software-hardware co-design framework for low-bit vision transformer,'' \emph{arXiv preprint arXiv:2201.06618}, 2022.

\bibitem{yu2022nn}
J.~Yu, J.~Park, S.~Park, M.~Kim, S.~Lee, D.~H. Lee, and J.~Choi, ``Nn-lut: neural approximation of non-linear operations for efficient transformer inference,'' in \emph{DAC}, 2022.

\bibitem{marchisio2023swifttron}
A.~Marchisio, D.~Dura, M.~Capra, M.~Martina, G.~Masera, and M.~Shafique, ``Swifttron: An efficient hardware accelerator for quantized transformers,'' \emph{arXiv preprint arXiv:2304.03986}, 2023.

\bibitem{nag2023vita}
S.~Nag, G.~Datta, S.~Kundu, N.~Chandrachoodan, and P.~A. Beerel, ``Vita: A vision transformer inference accelerator for edge applications,'' \emph{arXiv preprint arXiv:2302.09108}, 2023.

\bibitem{yin20171}
S.~Yin, P.~Ouyang, S.~Tang, F.~Tu, X.~Li, L.~Liu, and S.~Wei, ``A 1.06-to-5.09 tops/w reconfigurable hybrid-neural-network processor for deep learning applications,'' in \emph{Symposium on VLSI Circuits}, 2017.

\bibitem{zhang2019new}
H.~Zhang, D.~Chen, and S.-B. Ko, ``New flexible multiple-precision multiply-accumulate unit for deep neural network training and inference,'' \emph{IEEE Transactions on Computers}, 2019.

\bibitem{lee20197}
J.~Lee, J.~Lee, D.~Han, J.~Lee, G.~Park, and H.-J. Yoo, ``7.7 lnpu: A 25.3 tflops/w sparse deep-neural-network learning processor with fine-grained mixed precision of fp8-fp16,'' in \emph{ISSCC}.\hskip 1em plus 0.5em minus 0.4em\relax IEEE, 2019.

\bibitem{zhou2020convolutional}
X.~Zhou, L.~Zhang, C.~Guo, X.~Yin, and C.~Zhuo, ``A convolutional neural network accelerator architecture with fine-granular mixed precision configurability,'' in \emph{ISCAS}, 2020.

\bibitem{fleischer2018scalable}
B.~Fleischer, S.~Shukla, M.~Ziegler, J.~Silberman, J.~Oh, V.~Srinivasan, J.~Choi, S.~Mueller, A.~Agrawal, T.~Babinsky \emph{et~al.}, ``A scalable multi-teraops deep learning processor core for ai trainina and inference,'' in \emph{IEEE Symposium on VLSI Circuits}, 2018.

\bibitem{wang2018design}
J.~Wang, Q.~Lou, X.~Zhang, C.~Zhu, Y.~Lin, and D.~Chen, ``Design flow of accelerating hybrid extremely low bit-width neural network in embedded fpga,'' in \emph{FPL}, 2018.

\bibitem{agrawal20219}
A.~Agrawal, S.~K. Lee, J.~Silberman, M.~Ziegler, M.~Kang, S.~Venkataramani, N.~Cao, B.~Fleischer, M.~Guillorn, M.~Cohen \emph{et~al.}, ``9.1 a 7nm 4-core ai chip with 25.6 tflops hybrid fp8 training, 102.4 tops int4 inference and workload-aware throttling,'' in \emph{ISSCC}, 2021.

\bibitem{recanatesi2019dimensionality}
S.~Recanatesi \emph{et~al.}, ``Dimensionality compression and expansion in deep neural networks,'' \emph{arXiv preprint arXiv:1906.00443}, 2019.

\bibitem{oktay2019scalable}
D.~Oktay \emph{et~al.}, ``Scalable model compression by entropy penalized reparameterization,'' \emph{arXiv preprint arXiv:1906.06624}, 2019.

\bibitem{wiedemann2019deepcabac}
S.~Wiedemann \emph{et~al.}, ``Deepcabac: Context-adaptive binary arithmetic coding for deep neural network compression,'' \emph{arXiv preprint arXiv:1905.08318}, 2019.

\bibitem{dubey2018coreset}
A.~Dubey \emph{et~al.}, ``Coreset-based neural network compression,'' in \emph{ECCV}, 2018.

\bibitem{han2015deep}
S.~Han, H.~Mao, and W.~J. Dally, ``Deep compression: Compressing deep neural networks with pruning, trained quantization and huffman coding,'' in \emph{ICLR}, 2016.

\bibitem{chen2021efficient}
C.~Chen, Z.~Wang, X.~Chen, J.~Lin, and M.~M.~S. Aly, ``Efficient tunstall decoder for deep neural network compression,'' in \emph{DAC}, 2021.

\bibitem{cormen2009introduction}
T.~H. Cormen, C.~E. Leiserson, R.~L. Rivest, and C.~Stein, \emph{Introduction to algorithms}.\hskip 1em plus 0.5em minus 0.4em\relax MIT press, 2009.

\bibitem{rabbani2002jpeg2000}
M.~Rabbani, ``Jpeg2000: Image compression fundamentals, standards and practice,'' \emph{Journal of Electronic Imaging}, 2002.

\bibitem{witten1987arithmetic}
I.~H. Witten \emph{et~al.}, ``Arithmetic coding for data compression,'' \emph{Communications of the ACM}, 1987.

\bibitem{chen2016eyeriss}
Y.-H. Chen, T.~Krishna, J.~S. Emer, and V.~Sze, ``Eyeriss: An energy-efficient reconfigurable accelerator for deep convolutional neural networks,'' \emph{IEEE Journal of Solid-State Circuits}, 2016.

\bibitem{han2016eie}
S.~Han, X.~Liu, H.~Mao, J.~Pu, A.~Pedram, M.~A. Horowitz, and W.~J. Dally, ``Eie: efficient inference engine on compressed deep neural network,'' in \emph{ISCA}, 2016.

\bibitem{parashar2017scnn}
A.~Parashar, M.~Rhu, A.~Mukkara, A.~Puglielli, R.~Venkatesan, B.~Khailany, J.~Emer, S.~W. Keckler, and W.~J. Dally, ``Scnn: An accelerator for compressed-sparse convolutional neural networks,'' in \emph{ISCA}, 2017.

\bibitem{chen2019eyeriss}
Y.-H. Chen, T.-J. Yang, J.~Emer, and V.~Sze, ``Eyeriss v2: A flexible accelerator for emerging deep neural networks on mobile devices,'' \emph{IEEE Journal on Emerging and Selected Topics in Circuits and Systems}, 2019.

\bibitem{dave2021hardware}
S.~Dave, R.~Baghdadi, T.~Nowatzki, S.~Avancha, A.~Shrivastava, and B.~Li, ``Hardware acceleration of sparse and irregular tensor computations of ml models: A survey and insights,'' \emph{Proceedings of the IEEE}, 2021.

\bibitem{zhang2016cambricon}
S.~Zhang, Z.~Du, L.~Zhang, H.~Lan, S.~Liu, L.~Li, Q.~Guo, T.~Chen, and Y.~Chen, ``Cambricon-x: An accelerator for sparse neural networks,'' in \emph{MICRO}, 2016.

\bibitem{gustavson1972some}
F.~G. Gustavson, ``Some basic techniques for solving sparse systems of linear equations,'' in \emph{Sparse matrices and their applications}, 1972.

\bibitem{smith2015splatt}
S.~Smith, N.~Ravindran, N.~D. Sidiropoulos, and G.~Karypis, ``Splatt: Efficient and parallel sparse tensor-matrix multiplication,'' in \emph{IPDPS}, 2015.

\bibitem{hegde2019extensor}
K.~Hegde, H.~Asghari-Moghaddam, M.~Pellauer, N.~Crago, A.~Jaleel, E.~Solomonik, J.~Emer, and C.~W. Fletcher, ``Extensor: An accelerator for sparse tensor algebra,'' in \emph{MICRO}, 2019.

\bibitem{yuan2018sticker}
Z.~Yuan, J.~Yue, H.~Yang, Z.~Wang, J.~Li, Y.~Yang, Q.~Guo, X.~Li, M.-F. Chang, H.~Yang \emph{et~al.}, ``Sticker: A 0.41-62.1 tops/w 8bit neural network processor with multi-sparsity compatible convolution arrays and online tuning acceleration for fully connected layers,'' in \emph{IEEE symposium on VLSI circuits}, 2018.

\bibitem{zhang2019snap}
J.-F. Zhang, C.-E. Lee, C.~Liu, Y.~S. Shao, S.~W. Keckler, and Z.~Zhang, ``Snap: A 1.67—21.55 tops/w sparse neural acceleration processor for unstructured sparse deep neural network inference in 16nm cmos,'' in \emph{VLSIC}.\hskip 1em plus 0.5em minus 0.4em\relax IEEE, 2019.

\bibitem{kang2019accelerator}
H.-J. Kang, ``Accelerator-aware pruning for convolutional neural networks,'' \emph{IEEE TCSVT}, 2019.

\bibitem{lu2019efficient}
L.~Lu, J.~Xie, R.~Huang, J.~Zhang, W.~Lin, and Y.~Liang, ``An efficient hardware accelerator for sparse convolutional neural networks on fpgas,'' in \emph{FCCM}, 2019.

\bibitem{zhang2019compact}
J.~J. Zhang, P.~Raj, S.~Zarar, A.~Ambardekar, and S.~Garg, ``Compact: On-chip compression of activations for low power systolic array based cnn acceleration,'' \emph{TECS}, 2019.

\bibitem{zhou2018cambricon}
X.~Zhou, Z.~Du, Q.~Guo, S.~Liu, C.~Liu, C.~Wang, X.~Zhou, L.~Li, T.~Chen, and Y.~Chen, ``Cambricon-s: Addressing irregularity in sparse neural networks through a cooperative software/hardware approach,'' in \emph{MICRO}, 2018.

\bibitem{mishra2017fine}
A.~K. Mishra, E.~Nurvitadhi, G.~Venkatesh, J.~Pearce, and D.~Marr, ``Fine-grained accelerators for sparse machine learning workloads,'' in \emph{ASP-DAC}, 2017.

\bibitem{han2017ese}
S.~Han, J.~Kang, H.~Mao, Y.~Hu, X.~Li, Y.~Li, D.~Xie, H.~Luo, S.~Yao, Y.~Wang \emph{et~al.}, ``Ese: Efficient speech recognition engine with sparse lstm on fpga,'' in \emph{ACM/SIGDA FPGA}, 2017.

\bibitem{moons201714}
B.~Moons, R.~Uytterhoeven, W.~Dehaene, and M.~Verhelst, ``14.5 envision: A 0.26-to-10tops/w subword-parallel dynamic-voltage-accuracy-frequency-scalable convolutional neural network processor in 28nm fdsoi,'' in \emph{ISSCC}, 2017.

\bibitem{gpunms}
D.~Oro \emph{et~al.}, ``Work-efficient parallel non-maximum suppression for embedded gpu architectures,'' in \emph{ICASSP}, 2016.

\bibitem{powernms}
M.~Shi \emph{et~al.}, ``A fast and power-efficient hardware architecture for non-maximum suppression,'' \emph{TCAS-II}, 2019.

\bibitem{yuan2016efficient}
B.~Yuan, ``Efficient hardware architecture of softmax layer in deep neural network,'' in \emph{SOCC}, 2016.

\bibitem{li2018efficient}
Z.~Li, H.~Li, X.~Jiang, B.~Chen, Y.~Zhang, and G.~Du, ``Efficient fpga implementation of softmax function for dnn applications,'' in \emph{ASID}, 2018.

\bibitem{wang2018high}
M.~Wang, S.~Lu, D.~Zhu, J.~Lin, and Z.~Wang, ``A high-speed and low-complexity architecture for softmax function in deep learning,'' in \emph{APCCAS}.\hskip 1em plus 0.5em minus 0.4em\relax IEEE, 2018.

\bibitem{sun2018high}
Q.~Sun, Z.~Di, Z.~Lv, F.~Song, Q.~Xiang, Q.~Feng, Y.~Fan, X.~Yu, and W.~Wang, ``A high speed softmax vlsi architecture based on basic-split,'' in \emph{ICSICT}, 2018.

\bibitem{hu2018efficient}
R.~Hu, B.~Tian, S.~Yin, and S.~Wei, ``Efficient hardware architecture of softmax layer in deep neural network,'' in \emph{ICDSP}, 2018.

\bibitem{du2019efficient}
G.~Du, C.~Tian, Z.~Li, D.~Zhang, Y.~Yin, and Y.~Ouyang, ``Efficient softmax hardware architecture for deep neural networks,'' in \emph{Proceedings of the 2019 on Great Lakes Symposium on VLSI}, 2019.

\bibitem{kouretas2020hardware}
I.~Kouretas and V.~Paliouras, ``Hardware implementation of a softmax-like function for deep learning,'' \emph{Technologies}, 2020.

\bibitem{wang2019customized}
K.-Y. Wang, Y.-D. Huang, Y.-L. Ho, and W.-C. Fang, ``A customized convolutional neural network design using improved softmax layer for real-time human emotion recognition,'' in \emph{AICAS}, 2019.

\bibitem{cardarilli2021pseudo}
G.~C. Cardarilli, L.~Di~Nunzio, R.~Fazzolari, D.~Giardino, A.~Nannarelli, M.~Re, and S.~Span{\`o}, ``A pseudo-softmax function for hardware-based high speed image classification,'' \emph{Scientific reports}, 2021.

\bibitem{spagnolo2021aggressive}
F.~Spagnolo, S.~Perri, and P.~Corsonello, ``Aggressive approximation of the softmax function for power-efficient hardware implementations,'' \emph{IEEE Transactions on Circuits and Systems II: Express Briefs}, 2021.

\bibitem{stevens2021softermax}
J.~R. Stevens, R.~Venkatesan, S.~Dai, B.~Khailany, and A.~Raghunathan, ``Softermax: Hardware/software co-design of an efficient softmax for transformers,'' in \emph{DAC}, 2021.

\bibitem{gaines1969stochastic}
B.~R. Gaines, ``Stochastic computing systems,'' \emph{Advances in Information Systems Science}, 1969.

\bibitem{liu2020survey}
Y.~Liu, S.~Liu, Y.~Wang, F.~Lombardi, and J.~Han, ``A survey of stochastic computing neural networks for machine learning applications,'' \emph{IEEE TNNLS}, 2020.

\bibitem{salehi2020low}
S.~A. Salehi, ``Low-cost stochastic number generators for stochastic computing,'' \emph{VLSI}, 2020.

\bibitem{frasser2022fully}
C.~F. Frasser, P.~Linares-Serrano, I.~D. de~Los~Rios, A.~Moran, E.~S. Skibinsky-Gitlin, J.~Font-Rossello, V.~Canals, M.~Roca, T.~Serrano-Gotarredona, and J.~L. Rossello, ``Fully parallel stochastic computing hardware implementation of convolutional neural networks for edge computing applications,'' \emph{IEEE TNNLS}, 2022.

\bibitem{moran2022digital}
A.~Mor{\'a}n, L.~Parrilla, M.~Roca, J.~Font-Rossello, E.~Isern, and V.~Canals, ``Digital implementation of radial basis function neural networks based on stochastic computing,'' \emph{IEEE Journal on Emerging and Selected Topics in Circuits and Systems}, 2022.

\bibitem{paillier1999public}
P.~Paillier, ``Public-key cryptosystems based on composite degree residuosity classes,'' in \emph{Eurocrypt}, 1999.

\bibitem{rivest1978method}
R.~L. Rivest, A.~Shamir, and L.~Adleman, ``A method for obtaining digital signatures and public-key cryptosystems,'' \emph{Communications of the ACM}, 1978.

\bibitem{gentry2009fully}
C.~Gentry, ``Fully homomorphic encryption using ideal lattices,'' in \emph{STOC}, 2009.

\bibitem{isakov2019survey}
M.~Isakov, V.~Gadepally, K.~M. Gettings, and M.~A. Kinsy, ``Survey of attacks and defenses on edge-deployed neural networks,'' in \emph{HPEC}, 2019.

\bibitem{yagisawa2015fully}
M.~Yagisawa, ``Fully homomorphic encryption without bootstrapping,'' \emph{Cryptology ePrint Archive}, 2015.

\bibitem{fan2012somewhat}
J.~Fan and F.~Vercauteren, ``Somewhat practical fully homomorphic encryption,'' \emph{Cryptology ePrint Archive}, 2012.

\bibitem{aikata2023reed}
A.~Aikata, A.~C. Mert, S.~Kwon, M.~Deryabin, and S.~S. Roy, ``Reed: Chiplet-based scalable hardware accelerator for fully homomorphic encryption,'' \emph{arXiv preprint arXiv:2308.02885}, 2023.

\bibitem{zhang2019encrypted}
S.-X. Zhang, Y.~Gong, and D.~Yu, ``Encrypted speech recognition using deep polynomial networks,'' in \emph{ICASSP}, 2019.

\bibitem{lou2019she}
Q.~Lou and L.~Jiang, ``She: A fast and accurate deep neural network for encrypted data,'' \emph{NIPS}, 2019.

\bibitem{bourse2018fast}
F.~Bourse, M.~Minelli, M.~Minihold, and P.~Paillier, ``Fast homomorphic evaluation of deep discretized neural networks,'' in \emph{Crypto}, 2018.

\bibitem{pmlr-v80-sanyal18a}
A.~Sanyal, M.~Kusner, A.~Gascon, and V.~Kanade, ``{TAPAS}: Tricks to accelerate (encrypted) prediction as a service,'' in \emph{ICML}, 2018.

\bibitem{chou2018faster}
E.~Chou, J.~Beal, D.~Levy, S.~Yeung, A.~Haque, and L.~Fei-Fei, ``Faster cryptonets: Leveraging sparsity for real-world encrypted inference,'' \emph{arXiv preprint arXiv:1811.09953}, 2018.

\bibitem{lu2021ffconv}
Y.~Lu, J.~Lin, C.~Jin, Z.~Wang, M.~Wu, K.~M.~M. Aung, and X.~Li, ``Ffconv: Fast factorized convolutional neural network inference on encrypted data,'' \emph{arXiv preprint arXiv:2102.03494}, 2021.

\bibitem{khalil2010hardware}
M.~Khalil-Hani, V.~P. Nambiar, and M.~Marsono, ``Hardware acceleration of openssl cryptographic functions for high-performance internet security,'' in \emph{International Conference on Intelligent Systems, Modelling and Simulation}, 2010.

\bibitem{cousins2016designing}
D.~B. Cousins, K.~Rohloff, and D.~Sumorok, ``Designing an fpga-accelerated homomorphic encryption co-processor,'' \emph{IEEE Transactions on Emerging Topics in Computing}, 2016.

\bibitem{lee2020optimizing}
D.~Lee, W.~Lee, H.~Oh, and K.~Yi, ``Optimizing homomorphic evaluation circuits by program synthesis and term rewriting,'' in \emph{PLDI}, 2020.

\bibitem{kim2023sharp}
J.~Kim, S.~Kim, J.~Choi, J.~Park, D.~Kim, and J.~H. Ahn, ``Sharp: A short-word hierarchical accelerator for robust and practical fully homomorphic encryption,'' in \emph{ISCA}, 2023.

\bibitem{kim2022bts}
S.~Kim, J.~Kim, M.~J. Kim, W.~Jung, J.~Kim, M.~Rhu, and J.~H. Ahn, ``Bts: An accelerator for bootstrappable fully homomorphic encryption,'' in \emph{ISCA}, 2022.

\bibitem{doroz2014accelerating}
Y.~Dor{\"o}z, E.~{\"O}zt{\"u}rk, and B.~Sunar, ``Accelerating fully homomorphic encryption in hardware,'' \emph{IEEE Transactions on Computers}, 2014.

\bibitem{reagen2021cheetah}
B.~Reagen, W.-S. Choi, Y.~Ko, V.~T. Lee, H.-H.~S. Lee, G.-Y. Wei, and D.~Brooks, ``Cheetah: Optimizing and accelerating homomorphic encryption for private inference,'' in \emph{HPCA}, 2021.

\bibitem{chillotti2020tfhe}
I.~Chillotti, N.~Gama, M.~Georgieva, and M.~Izabach{\`e}ne, ``Tfhe: fast fully homomorphic encryption over the torus,'' \emph{Journal of Cryptology}, 2020.

\bibitem{cheon2017homomorphic}
J.~H. Cheon, A.~Kim, M.~Kim, and Y.~Song, ``Homomorphic encryption for arithmetic of approximate numbers,'' in \emph{ASIACRYPT}, 2017.

\bibitem{brown2020language}
T.~Brown, B.~Mann, N.~Ryder, M.~Subbiah, J.~D. Kaplan, P.~Dhariwal, A.~Neelakantan, P.~Shyam, G.~Sastry, A.~Askell \emph{et~al.}, ``Language models are few-shot learners,'' \emph{NIPS}, 2020.

\bibitem{touvron2023llama}
H.~Touvron, T.~Lavril, G.~Izacard, X.~Martinet, M.-A. Lachaux, T.~Lacroix, B.~Rozi{\`e}re, N.~Goyal, E.~Hambro, F.~Azhar \emph{et~al.}, ``Llama: Open and efficient foundation language models,'' \emph{arXiv preprint arXiv:2302.13971}, 2023.

\bibitem{yao2022zeroquant}
Z.~Yao, R.~Yazdani~Aminabadi, M.~Zhang, X.~Wu, C.~Li, and Y.~He, ``Zeroquant: Efficient and affordable post-training quantization for large-scale transformers,'' \emph{NIPS}, 2022.

\bibitem{xu2022etinynet}
K.~Xu, Y.~Li, H.~Zhang, R.~Lai, and L.~Gu, ``Etinynet: Extremely tiny network for tinyml,'' in \emph{AAAI}, 2022.

\bibitem{yang2020mobileda}
J.~Yang, H.~Zou, S.~Cao, Z.~Chen, and L.~Xie, ``Mobileda: Toward edge-domain adaptation,'' \emph{IEEE Internet of Things Journal}, 2020.

\bibitem{ryu2022knowledge}
M.~Ryu, G.~Lee, and K.~Lee, ``Knowledge distillation for bert unsupervised domain adaptation,'' \emph{Knowledge and Information Systems}, 2022.

\end{thebibliography}


\begin{thebibliography}{10}
\providecommand{\url}[1]{#1}
\csname url@samestyle\endcsname
\providecommand{\newblock}{\relax}
\providecommand{\bibinfo}[2]{#2}
\providecommand{\BIBentrySTDinterwordspacing}{\spaceskip=0pt\relax}
\providecommand{\BIBentryALTinterwordstretchfactor}{4}
\providecommand{\BIBentryALTinterwordspacing}{\spaceskip=\fontdimen2\font plus
\BIBentryALTinterwordstretchfactor\fontdimen3\font minus \fontdimen4\font\relax}
\providecommand{\BIBforeignlanguage}[2]{{%
\expandafter\ifx\csname l@#1\endcsname\relax
\typeout{** WARNING: IEEEtran.bst: No hyphenation pattern has been}%
\typeout{** loaded for the language `#1'. Using the pattern for}%
\typeout{** the default language instead.}%
\else
\language=\csname l@#1\endcsname
\fi
#2}}
\providecommand{\BIBdecl}{\relax}
\BIBdecl

\bibitem{liu2017survey}
W.~Liu, Z.~Wang, X.~Liu, N.~Zeng, Y.~Liu, and F.~E. Alsaadi, ``A survey of deep neural network architectures and their applications,'' \emph{Neurocomputing}, 2017.

\bibitem{thakkar2021comprehensive}
A.~Thakkar and K.~Chaudhari, ``A comprehensive survey on deep neural networks for stock market: The need, challenges, and future directions,'' \emph{Expert Systems with Applications}, 2021.

\bibitem{Krizhevsky2012nips}
A.~Krizhevsky, I.~Sutskever, and G.~Hinton, ``Image{N}et classification with deep convolutional neural networks,'' in \emph{NIPS}, 2012.

\bibitem{Szegedy2015CVPR}
C.~Szegedy, W.~Liu, Y.~Jia, P.~Sermanet, S.~Anguelov, D.~Erhan, V.~Vanhoucke, and A.~Rabinovich, ``Going deeper with convolutions,'' in \emph{CVPR}, 2015.

\bibitem{Simonyan2015ICLR}
K.~Simonyan and A.~Zisserman, ``Very deep convolutional networks for large-scale image recognition,'' in \emph{ICLR}, 2015.

\bibitem{he2016CVPR}
K.~He, X.~Zhang, S.~Ren, and J.~Sun, ``Deep residual learning for image recognition,'' in \emph{CVPR}, 2016.

\bibitem{i2016squeezenet}
F.~N. Iandola, S.~Han, M.~W. Moskewicz, K.~Ashraf, W.~J. Dally, and K.~Keutzer, ``Squeezenet: Alexnet-level accuracy with 50x fewer parameters and< 0.5 mb model size,'' \emph{arXiv preprint arXiv:1602.07360}, 2016.

\bibitem{huang2017densely}
G.~Huang, Z.~Liu, L.~Van Der~Maaten, and K.~Q. Weinberger, ``Densely connected convolutional networks,'' in \emph{CVPR}, 2017.

\bibitem{Sandler2018arXiv}
M.~Sandler, A.~Howard, M.~Zhu, A.~Zhmoginov, and L.-C. Chen, ``Mobilenetv2: Inverted residuals and linear bottlenecks,'' in \emph{arXiv}, 2018.

\bibitem{vaswani2017attention}
A.~Vaswani, N.~Shazeer, N.~Parmar, J.~Uszkoreit, L.~Jones, A.~N. Gomez, {\L}.~Kaiser, and I.~Polosukhin, ``Attention is all you need,'' \emph{NIPS}, 2017.

\bibitem{brown2020language}
T.~Brown, B.~Mann, N.~Ryder, M.~Subbiah, J.~D. Kaplan, P.~Dhariwal, A.~Neelakantan, P.~Shyam, G.~Sastry, A.~Askell \emph{et~al.}, ``Language models are few-shot learners,'' \emph{NIPS}, 2020.

\bibitem{bwn}
I.~Hubara, M.~Courbariaux, D.~Soudry, R.~El-Yaniv, and Y.~Bengio, ``Quantized neural networks: Training neural networks with low precision weights and activations,'' in \emph{arXiv:1609.07061}, 2016.

\bibitem{xnor}
M.~Rastegari, V.~Ordonez, J.~Redmon, and A.~Farhadi, ``{XNOR}-{NET}: {I}mage{n}et classification using binary convolutional neural networks,'' in \emph{arXiv preprint arXiv:1603.05279}, 2016.

\bibitem{Li16nips}
F.~Li, B.~Zhang, and B.~Liu, ``Ternary weight networks,'' in \emph{NIPS Workshop}, 2016.

\bibitem{zhu2016trained}
C.~Zhu, S.~Han, H.~Mao, and W.~J. Dally, ``Trained ternary quantization,'' \emph{arXiv preprint arXiv:1612.01064}, 2016.

\bibitem{Zhang18eccv}
D.~Zhang, J.~Yang, D.~Ye, and G.~Hua, ``Lq-nets: learned quantization for highly accurate and compact deep neural networks,'' in \emph{ECCV}, 2018.

\bibitem{inq}
A.~Zhou, A.~Yao, Y.~Guo, L.~Xu, and Y.~Chen, ``Incremental network quantization: {T}owards lossless {cnn}s with low-precision weights,'' in \emph{arXiv preprint arXiv:1702.03044}, 2017.

\bibitem{dorefa}
S.~Zhou, Y.~Wu, Z.~Ni, X.~Zhou, H.~Wen, and Y.~Zou, ``Dorefa-net: {T}raining low bitwidth convolutional neural networks with low bitwidth gradients,'' in \emph{arXiv preprint arXiv:1606.06160}, 2016.

\bibitem{datafree}
M.~Nagel, M.~Baalen, T.~Blankevoort, and M.~Welling, ``Datafree quantization through weight equalization and bias correction,'' \emph{CVPR}, 2019.

\bibitem{DSQ}
R.~Gong, X.~Liu, S.~Jiang, T.~Li, P.~Hu, J.~Lin, F.~Yu, and J.~Yan, ``Differentiable soft quantization: Bridging full-precision and low-bit neu- ral networks,'' in \emph{ICCV}, 2019.

\bibitem{layerwise}
S.~Chen, W.~Wang, and S.~J. Pan, ``Deep neural network quantization via layer-wise optimization using limited training data,'' in \emph{AAAI}, 2019.

\bibitem{Zhuang2018cvpr}
B.~Zhuang, C.~Shen, M.~Tan, L.~Liu, and I.~Reid, ``Towards effective low-bitwidth convolutional neural networks,'' in \emph{CVPR}, 2018.

\bibitem{QIL}
S.~Jung, C.~Son, S.~Lee, J.~Son, J.-J. Han, Y.~Kwak, S.~J. Hwang, and C.~Choi, ``Learning to quantize deep networks by optimizing quantization intervals with task loss,'' in \emph{CVPR}, 2019.

\bibitem{SAT}
Q.~Jin, L.~Yang, Z.~Liao, and X.~Qian, ``Neural network quantization with scale-adjusted training,'' \emph{BMVC}, 2020.

\bibitem{esser2019learned}
S.~K. Esser, J.~L. McKinstry, D.~Bablani, R.~Appuswamy, and D.~S. Modha, ``Learned step size quantization,'' \emph{arXiv preprint arXiv:1902.08153}, 2019.

\bibitem{Auxi}
B.~Zhuang, L.~Liu, M.~Tan, C.~Shen, and I.~Reid, ``Training quantized neural networks with a full-precision auxiliary module,'' in \emph{CVPR}, 2020.

\bibitem{pq}
Y.~Gong, L.~Liu, M.~Yang, and L.~Bourdev, ``Compressing deep convolutional networks using vector quantization,'' in \emph{ICLR}, 2015.

\bibitem{hash}
W.~Chen, J.~T. Wilson, S.~Tyree, K.~Q. Weinberger, and Y.~Chen, ``Compressing neural networks with the hashing trick,'' in \emph{ICML}, 2015.

\bibitem{frequency}
W.~Chen, J.~Wilson, S.~Tyree, K.~Q. Weinberger, and Y.~Chen, ``Compressing convolutional neural networks in the frequency domain,'' in \emph{SIGKDD}, 2016.

\bibitem{revisit}
P.~Stock, A.~Joulin, R.~Gribonval, B.~Graham, and H.~Jegou, ``And the bit goes down: revisiting the quantization of neural networks,'' in \emph{ICLR}, 2020.

\bibitem{APoT}
Y.~Li, X.~Dong, and W.~Wang, ``Additive powers-of-two quantization: An efficient non-uniform discretization for neural networks,'' in \emph{ICLR}, 2020.

\bibitem{Faraone2018arxiv}
J.~Faraone, N.~Fraser, M.~Blott, and P.~H. Leong, ``Syq: Learning symmetric quantization for efficient deep neural networks,'' in \emph{arXiv}, 2018.

\bibitem{hinton2015distilling}
G.~Hinton, O.~Vinyals, J.~Dean \emph{et~al.}, ``Distilling the knowledge in a neural network,'' \emph{arXiv preprint arXiv:1503.02531}, 2015.

\bibitem{romero2014fitnets}
A.~Romero, N.~Ballas, S.~E. Kahou, A.~Chassang, C.~Gatta, and Y.~Bengio, ``Fitnets: Hints for thin deep nets,'' \emph{arXiv preprint arXiv:1412.6550}, 2014.

\bibitem{chen2022knowledge}
D.~Chen, J.-P. Mei, H.~Zhang, C.~Wang, Y.~Feng, and C.~Chen, ``Knowledge distillation with the reused teacher classifier,'' in \emph{CVPR}, 2022.

\bibitem{wang2017model}
C.~Wang, X.~Lan, and Y.~Zhang, ``Model distillation with knowledge transfer from face classification to alignment and verification,'' \emph{arXiv preprint arXiv:1709.02929}, 2017.

\bibitem{wang2019distilling}
T.~Wang, L.~Yuan, X.~Zhang, and J.~Feng, ``Distilling object detectors with fine-grained feature imitation,'' in \emph{CVPR}, 2019.

\bibitem{xu2021kdnet}
Q.~Xu, Z.~Chen, K.~Wu, C.~Wang, M.~Wu, and X.~Li, ``Kdnet-rul: A knowledge distillation framework to compress deep neural networks for machine remaining useful life prediction,'' \emph{IEEE Transactions on Industrial Electronics}, 2021.

\bibitem{xu2022contrastive}
Q.~Xu, Z.~Chen, M.~Ragab, C.~Wang, M.~Wu, and X.~Li, ``Contrastive adversarial knowledge distillation for deep model compression in time-series regression tasks,'' \emph{Neurocomputing}, 2022.

\bibitem{xu2017training}
Z.~Xu, Y.-C. Hsu, and J.~Huang, ``Training shallow and thin networks for acceleration via knowledge distillation with conditional adversarial networks,'' in \emph{ICLR Workshop}, 2018.

\bibitem{gao2019adversarial}
L.~Gao, H.~Mi, B.~Zhu, D.~Feng, Y.~Li, and Y.~Peng, ``An adversarial feature distillation method for audio classification,'' \emph{IEEE Access}, 2019.

\bibitem{chung2020feature}
I.~Chung, S.~Park, J.~Kim, and N.~Kwak, ``Feature-map-level online adversarial knowledge distillation,'' in \emph{ICML}, 2020.

\bibitem{park2020feature}
S.~Park and N.~Kwak, ``Feature-level ensemble knowledge distillation for aggregating knowledge from multiple networks,'' in \emph{ECAI 2020}, 2020.

\bibitem{liu2021semantics}
Y.~Liu, K.~Wang, G.~Li, and L.~Lin, ``Semantics-aware adaptive knowledge distillation for sensor-to-vision action recognition,'' \emph{IEEE TIP}, 2021.

\bibitem{chebotar2016distilling}
Y.~Chebotar and A.~Waters, ``Distilling knowledge from ensembles of neural networks for speech recognition.'' in \emph{Interspeech}, 2016.

\bibitem{wu2019multi}
M.-C. Wu, C.-T. Chiu, and K.-H. Wu, ``Multi-teacher knowledge distillation for compressed video action recognition on deep neural networks,'' in \emph{ICASSP}, 2019.

\bibitem{fukuda2017efficient}
T.~Fukuda, M.~Suzuki, G.~Kurata, S.~Thomas, J.~Cui, and B.~Ramabhadran, ``Efficient knowledge distillation from an ensemble of teachers.'' in \emph{Interspeech}, 2017.

\bibitem{yuan2021reinforced}
F.~Yuan, L.~Shou, J.~Pei, W.~Lin, M.~Gong, Y.~Fu, and D.~Jiang, ``Reinforced multi-teacher selection for knowledge distillation,'' in \emph{AAAI}, 2021.

\bibitem{furlanello2018born}
T.~Furlanello, Z.~Lipton, M.~Tschannen, L.~Itti, and A.~Anandkumar, ``Born again neural networks,'' in \emph{ICML}, 2018.

\bibitem{tian2020rethinking}
Y.~Tian, Y.~Wang, D.~Krishnan, J.~B. Tenenbaum, and P.~Isola, ``Rethinking few-shot image classification: a good embedding is all you need?'' in \emph{ECCV}, 2020.

\bibitem{zhang2018deep}
Y.~Zhang, T.~Xiang, T.~M. Hospedales, and H.~Lu, ``Deep mutual learning,'' in \emph{CVPR}, 2018.

\bibitem{ji2021refine}
M.~Ji, S.~Shin, S.~Hwang, G.~Park, and I.-C. Moon, ``Refine myself by teaching myself: Feature refinement via self-knowledge distillation,'' in \emph{CVPR}, 2021.

\bibitem{lopes2017data}
R.~G. Lopes, S.~Fenu, and T.~Starner, ``Data-free knowledge distillation for deep neural networks,'' \emph{NIPS LLD Workshop}, 2017.

\bibitem{yin2020dreaming}
H.~Yin, P.~Molchanov, J.~M. Alvarez, Z.~Li, A.~Mallya, D.~Hoiem, N.~K. Jha, and J.~Kautz, ``Dreaming to distill: Data-free knowledge transfer via deepinversion,'' in \emph{CVPR}, 2020.

\bibitem{chen2019data}
H.~Chen, Y.~Wang, C.~Xu, Z.~Yang, C.~Liu, B.~Shi, C.~Xu, C.~Xu, and Q.~Tian, ``Data-free learning of student networks,'' in \emph{ICCV}, 2019.

\bibitem{zhang2021data}
Y.~Zhang, H.~Chen, X.~Chen, Y.~Deng, C.~Xu, and Y.~Wang, ``Data-free knowledge distillation for image super-resolution,'' in \emph{CVPR}, 2021.

\bibitem{nayak2019zero}
G.~K. Nayak, K.~R. Mopuri, V.~Shaj, V.~B. Radhakrishnan, and A.~Chakraborty, ``Zero-shot knowledge distillation in deep networks,'' in \emph{ICML}, 2019.

\bibitem{ma2021undistillable}
H.~Ma and T.~Chen, ``Undistillable: Making a nasty teacher that cannot teach students,'' in \emph{ICLR}, 2021.

\bibitem{jandial2022distilling}
S.~Jandial, Y.~Khasbage, A.~Pal, V.~N. Balasubramanian, and B.~Krishnamurthy, ``Distilling the undistillable: Learning from a nasty teacher,'' in \emph{ECCV}, 2022.

\bibitem{horowitz20141}
M.~Horowitz, ``1.1 computing's energy problem (and what we can do about it),'' in \emph{ISSCC}, 2014.

\bibitem{vivienne2019efficient}
V.~Sze, ``Efficient processing of deep neural networks: from algorithms to hardware architectures,'' in \emph{NIPS}, 2019.

\bibitem{chen2021efficient}
C.~Chen, Z.~Wang, X.~Chen, J.~Lin, and M.~M.~S. Aly, ``Efficient tunstall decoder for deep neural network compression,'' in \emph{DAC}, 2021.

\bibitem{chen2019eyeriss}
Y.-H. Chen, T.-J. Yang, J.~Emer, and V.~Sze, ``Eyeriss v2: A flexible accelerator for emerging deep neural networks on mobile devices,'' \emph{IEEE Journal on Emerging and Selected Topics in Circuits and Systems}, 2019.

\bibitem{zhang2019snap}
J.-F. Zhang, C.-E. Lee, C.~Liu, Y.~S. Shao, S.~W. Keckler, and Z.~Zhang, ``Snap: A 1.67—21.55 tops/w sparse neural acceleration processor for unstructured sparse deep neural network inference in 16nm cmos,'' in \emph{VLSIC}.\hskip 1em plus 0.5em minus 0.4em\relax IEEE, 2019.

\bibitem{zhou2018cambricon}
X.~Zhou, Z.~Du, Q.~Guo, S.~Liu, C.~Liu, C.~Wang, X.~Zhou, L.~Li, T.~Chen, and Y.~Chen, ``Cambricon-s: Addressing irregularity in sparse neural networks through a cooperative software/hardware approach,'' in \emph{MICRO}, 2018.

\bibitem{psrrmaxpool}
T.~Zhang \emph{et~al.}, ``Psrr-maxpoolnms: Pyramid shifted maxpoolnms with relationship recovery,'' in \emph{CVPR}, 2021.

\bibitem{gpunms2}
D.~Oro \emph{et~al.}, ``{Work-Efficient Parallel Non-Maximum Suppression Kernels},'' \emph{The Computer Journal}, 08 2020.

\bibitem{powernms}
M.~Shi \emph{et~al.}, ``A fast and power-efficient hardware architecture for non-maximum suppression,'' \emph{TCAS-II}, 2019.

\bibitem{shapoolnms}
C.~Chen, T.~Zhang, Z.~Yu, A.~Raghuraman, S.~Udayan, J.~Lin, and M.~M.~S. Aly, ``Scalable hardware acceleration of non-maximum suppression,'' in \emph{DATE}, 2022.

\end{thebibliography}

\vfill

\end{document}


\subsection{Deep Neural Network}
\label{subsec:DNNs}
The computation of a neuron of the neural network usually involves a weighted sum of input values, followed by a non-linear function including Sigmoid, ReLU, and Softmax function. Furthermore, the outputs of neurons are often referred to as \emph{activations}, while the synapses are \emph{weights}. Mathematically, the computation at each layer is: 
\begin{equation}
    y_{j}=f(\sum\limits_{i=1}^{K} W_{ij} \times x_{i} + b)
\end{equation}
where $K$, $W_{ij}$, $x_{i}$ and $y_{j}$ are the number of neurons and weights, input activations and output activations, respectively, and \emph{$f(\cdot)$} is a non-linear function. We generally use Deep Neural Networks (DNNs)~\cite{liu2017survey, thakkar2021comprehensive} to refer to the neural networks with more than three layers, \textit{i.e.}, more than one hidden layer.
DNNs are capable of learning high-level features with more complexity and abstraction than shallower neural networks. The deep feature hierarchy enables DNNs to use various kinds of features to accurately represent the input data.

To date, DNNs have evolved significantly, introducing innovative architectures that set new performance standards. 
As a pioneering work, AlexNet~\cite{Krizhevsky2012nips} won the ImageNet Large Scale Visual Recognition Challenge (ILSVRC)  in 2012 and demonstrated the power of deep neural networks in image recognition. GoogleNet~\cite{Szegedy2015CVPR} and VGGNet~\cite{Simonyan2015ICLR} introduced innovative architectures with various network depths and layer configurations, contributing to improved accuracy in image recognition tasks.
ResNet~\cite{he2016CVPR} introduced the innovative techniques of residual learning and skip connections with bottleneck architecture, allowing for the training of exceptionally deep neural networks by facilitating the flow of gradients and mitigating the degradation problem.
SqueezeNet~\cite{i2016squeezenet} optimized model size and computational efficiency by incorporating $1\times1$ convolutions, squeeze-and-excitation modules for channel-wise attention, and aggressive down-sampling, enabling high-performance deep neural networks with significantly reduced parameters for resource-constrained environments.
DenseNet~\cite{huang2017densely} utilized densely connected blocks where each layer receives direct inputs from all preceding layers, promoting feature reuse, alleviating the vanishing gradient problem, and enhancing parameter efficiency in deep neural network training.
MobileNet~\cite{Sandler2018arXiv} employed depthwise separable convolutions, a lightweight architecture, and efficient design choices, enabling the development of highly efficient and low-latency neural networks suitable for mobile and edge devices. 
The Transformer architecture~\cite{vaswani2017attention} introduced self-attention mechanisms and parallelization, dispensing with recurrent or convolutional layers, thereby revolutionizing natural language processing tasks and enabling effective modeling of sequential data.
GPT-3 (Generative Pre-trained Transformer 3)~\cite{brown2020language} leveraged a massive neural network with 175 billion parameters, enabling it to perform a wide array of natural language processing tasks through pre-training on diverse datasets. These architectures are propelling AI into uncharted territories, demonstrating the relentless pursuit of excellence in deep learning.


\subsection{Review on Uniform Quantizers}
For instance, Hubara \textit{et al.} \cite{bwn} proposed a training approach to train quantized weights and activations with binary format. Rastegari \textit{et al.} \cite{xnor} presented XNOR-Nets which approximate convolutions using primarily binary operations.
Li \textit{et al.} \cite{Li16nips} introduced Ternary Weight Networks (TWNs) with weights constrained to 1, 0, and -1 (2-bits).
Finally, Zhu \textit{et al.} \cite{zhu2016trained} proposed Trained Ternary Quantization (TTQ) which trains quantized networks with low precision (2-bits) from scratch.

Many quantization approaches with uniform quantizers are proposed to quantize neural networks to any precision.
For example, LQ-Nets \cite{Zhang18eccv} proposed jointly training a quantized, bit-operation-compatible DNN and its associated quantizers.
INQ \cite{inq} targeted to convert pre-trained neural networks into low-precision versions whose weights are constrained to be either power of two or zero.
DOREFA-NET \cite{dorefa} proposed to train neural networks with low bit-width weights, activations and gradients.
Nagel \textit{et al.} \cite{datafree} proposed data-free quantization through weight equalization and bias correction.
Gong \textit{et al.} \cite{DSQ} proposed Differentiable Soft Quantization (DSQ) to bridge the gap between the full-precision and low-bit networks.
Chen \textit{et al.} \cite{layerwise} presented a layer-wise quantization technique for deep neural networks, which requires very limited training data (only 1\% of the original dataset).
Zhuang \textit{et al.} \cite{Zhuang2018cvpr} proposed two-stage optimization, namely progressive optimization and joint optimization, to train low-precision weights and activations.
Jung \textit{et al.} \cite{QIL} proposed learning to quantize activations and weights via a trainable quantizer that transforms and discretizes them. 
SAT \cite{SAT} directly scaled down weights in the last layer to alleviate over-fitting.
Esser \textit{et al.} \cite{esser2019learned} involved estimating and scaling the task loss gradient at the quantizer step size of each weight and activation layer, enabling it to be learned simultaneously with other network parameters. 
Finally,
Zhuang \textit{et al.} \cite{Auxi} proposed a method  involves training a network with low-precision using an auxiliary module that operates at full-precision. 

\begin{table*}[!htbp]
\footnotesize
\centering
\begin{threeparttable}
\caption{A summary of quantization approaches with uniform or non-uniform quantizers.  
}
\begin{tabular}{ c | c | c | c | c | c | c }
\hline
Method & Type of Quantizer & Model & Weight & Activation & Accuracy (Original) & Dataset \\
\hline
\multirow{2}{*}{Hubara \textit{et al.} \cite{bwn}} & \multirow{2}{*}{Uniform} & AlexNet \cite{Krizhevsky2012nips}  & 1 Bit & 1 Bit & 41.8\% (56.5\%) & ImageNet \\
&  & GoogleNet \cite{Szegedy2015CVPR} & 1 Bit & 1 Bit & 47.1\% (71.6\%) & ImageNet \\
\hline
\multirow{2}{*}{XNOR-Nets \cite{xnor}} & \multirow{2}{*}{Uniform} & AlexNet & 1 Bit & 1 Bit & 44.2\% (56.6\%) & ImageNet \\
 & & ResNet-18 \cite{he2016CVPR} & 1 Bit & 1 Bit & 51.2\% (69.3\%) & ImageNet \\
\hline
TWN \cite{Li16nips} & Uniform & ResNet-18 & 2 Bits & 32 Bits & 61.8\% (65.4\%) & ImageNet \\
\hline
\multirow{2}{*}{TTQ \cite{zhu2016trained}} & \multirow{2}{*}{Uniform} & AlexNet &  2 Bits & 32 Bits & 57.5\% (57.2\%) & ImageNet \\
 & & ResNet-18 &  2 Bits & 32 Bits & 66.6\% (69.6\%) & ImageNet \\
\hline
\multirow{5}{*}{LQ-Nets \cite{Zhang18eccv}} & \multirow{5}{*}{Uniform} 
& AlexNet &  1 Bit & 2 Bits & 55.7\% (61.8\%) & ImageNet \\
& & GoogleNet &  1 Bit & 2 Bits & 65.6\% (72.9\%) & ImageNet \\
& & VGGNet \cite{Simonyan2015ICLR} & 1 Bit & 2 Bits & 67.1\% (72.0\%) & ImageNet \\
& & ResNet-50 \cite{he2016CVPR} &  1 Bit & 2 Bits & 68.7\% (76.4\%) & ImageNet \\
& & DenseNet \cite{huang2017densely} &  2 Bits & 2 Bits & 69.6\% (75.3\%) & ImageNet \\
\hline
\multirow{4}{*}{INQ \cite{inq}} & \multirow{4}{*}{Uniform} & AlexNet &  5 Bits & 32 Bits & 57.4\% (57.2\%) & ImageNet \\
& & VGGNet &  5 Bits & 32 Bits & 70.8\% (69.0\%) & ImageNet \\
& & GoogleNet &  5 Bits & 32 Bits & 69.0\% (68.9\%) & ImageNet \\
& & ResNet-50 &  5 Bits & 32 Bits & 74.8\% (73.2\%) & ImageNet \\
\hline
DOREFA \cite{dorefa} & Uniform & AlexNet &  1 Bit & 1 Bit & 43.6\% (55.9\%) & ImageNet \\
\hline
\multirow{2}{*}{Nagel \textit{et al.} \cite{datafree}} & \multirow{2}{*}{Uniform} 
& ResNet-18 &  6 Bits & 6 Bits & 66.3\% (69.7\%) & ImageNet \\
& & MobileNet-V2 \cite{Sandler2018arXiv} &  8 Bits & 8 Bits & 71.2\% (71.7\%) & ImageNet \\
\hline
\multirow{2}{*}{DSQ \cite{DSQ}} & \multirow{2}{*}{Uniform} 
& ResNet-34 &  4 Bits & 4 Bits & 72.8\% (73.8\%) & ImageNet \\
& & MobileNet-V2 &  4 Bits & 4 Bits & 64.8\% (71.9\%) & ImageNet \\
\hline
\multirow{2}{*}{Chen \textit{et al.} \cite{layerwise}} & \multirow{2}{*}{Uniform} 
& AlexNet & 3 Bits & 3 Bits & 40.5\% (58.3\%) & ImageNet \\
& & ResNet-18 &  3 Bits & 3 Bits & 53.4\% (69.8\%) & ImageNet \\
\hline
\multirow{2}{*}{Zhuang \textit{et al.} \cite{Zhuang2018cvpr}} & \multirow{2}{*}{Uniform} & AlexNet &  2 Bits & 2 Bits & 51.6\% (57.2\%) & ImageNet \\
& & ResNet-50 &  2 Bits & 2 Bits & 70.0\% (75.6\%) & ImageNet \\
\hline
\multirow{2}{*}{Jung \textit{et al.} \cite{QIL}} & \multirow{2}{*}{Uniform} 
& AlexNet &  2 Bits & 2 Bits & 58.1\% (61.8\%) & ImageNet \\
& & ResNet-18 &  2 Bits & 2 Bits & 65.7\% (70.2\%) & ImageNet \\
\hline
\multirow{2}{*}{SAT \cite{SAT}} & \multirow{2}{*}{Uniform} 
& ResNet-50 &  2 Bits & 2 Bits & 73.3\% (75.9\%) & ImageNet \\
& & MobileNet-V2  & 4 Bits & 4 Bits & 71.1\% (71.8\%) & ImageNet \\
\hline
\multirow{3}{*}{Esser \textit{et al.} \cite{esser2019learned}} & \multirow{3}{*}{Uniform} 
& VGGNet &  2 Bits & 2 Bits & 71.4\% (73.4\%) & ImageNet \\
& & ResNet-50 &  2 Bits & 2 Bits & 73.7\% (76.9\%) & ImageNet \\
& & SqueezeNet~\cite{i2016squeezenet} & 2 Bits & 2 Bits & 53.3\% (67.3\%) & ImageNet \\
\hline
Zhuang \textit{et al.} \cite{Auxi} & Uniform & ResNet-50 &  2 Bits & 2 Bits & 73.8\% (---\tnote{b} ) & ImageNet \\
\hline
Gong \textit{et al.} \cite{pq} & Non-Uniform & CNN\tnote{a} &  1 Bit & 32 Bits & 64.6\% (66.4\%) & Holidays \\
\hline
\multirow{2}{*}{HashNets \cite{hash}} & \multirow{2}{*}{Non-Uniform} & 3-Layer CNN\tnote{a} &  0.5 Bits & 32 Bits & 97.3\% (---\tnote{b} ) & MNIST \\
& & 5-Layer CNN\tnote{a} &  0.5 Bits & 32 Bits & 98.1\% (---\tnote{b} ) & MNIST \\
\hline
\multirow{2}{*}{Chen \textit{et al.} \cite{frequency}} & \multirow{2}{*}{Non-Uniform} & CNN\tnote{a} &  2 Bits & 32 Bits & 78.6\% (85.1\%) & CIFAR10 \\
& & CNN\tnote{a} &  0.5 Bits & 32 Bits & 69.2\% (85.6\%) & CIFAR10 \\
\hline
Stock \textit{et al.} \cite{revisit} & Non-Uniform & ResNet-50 &  1 Bit & 32 bits & 68.2\% (76.2\%) & ImageNet \\ 
\hline
APoT \cite{APoT} &  Non-Uniform & ResNet-18 &  3 Bits & 3 Bits & 68.5\% (70.2\%) & ImageNet \\
\hline
\multirow{3}{*}{SYQ \cite{Faraone2018arxiv}} & \multirow{3}{*}{Non-Uniform} & ResNet-18 &  3 Bits & 3 Bits & 68.2\% (70.3\%) & ImageNet \\
& & AlexNet &  3 Bits & 3 Bits & 54.3\% (56.4\%) & ImageNet \\
& & MobileNet-V2 &  4 Bits & 4 Bits & 67.4\% (71.8\%) & ImageNet \\
\hline
\end{tabular}\label{table:quantizer}
\begin{tablenotes}
    \item[a] The network architecture is unconventional and is specified in the quantization paper.
    \item[b] The corresponding paper does not report the accuracy of original models.
\end{tablenotes}
\end{threeparttable}
\end{table*}
In Table~\ref{table:quantizer}, the bit width of non-uniform quantization approaches are decimal numbers. 
It is calculated by the size of parameters after quantization divided by the number of parameters.

 \subsection{Distillation Schemes}
\label{sec:distillation_schemes}
\textbf{Metric-specified vs. Metric-free Distillation}
The conventional KD methods minimize the discrepancy between teacher and student's outputs (either logits or feature representations) using various explicit-defined distance metrics \cite{hinton2015distilling, romero2014fitnets,chen2022knowledge, wang2017model,wang2019distilling}. 
An alternative approach is to adopt the adversarial learning scheme. In particular, instead of manually selecting specific distance metrics in advance, a discriminator is employed to distinguish the source of logits or the feature maps (\textit{i.e.,} from teacher or student). By adversarially training the discriminator and student, the student is able to generate similar feature maps as its teacher. Previous works \cite{xu2021kdnet, xu2022contrastive, xu2017training, gao2019adversarial, chung2020feature} have demonstrated the effectiveness of adversarial learning in enhancing student's education with teacher's knowledge. 


\textbf{Teacher-specified vs. Teacher-free Distillation}
The distillation knowledge can be extracted from single teachers or multi-teacher. For the single-teacher KD scenario, the knowledge source is straightforward. For multi-teacher KD scenario, simply averaging over the outputs of all teachers  \cite{hinton2015distilling,park2020feature, liu2021semantics} or using fixed weights for different teachers are two commonly-used strategies \cite{chebotar2016distilling, wu2019multi}. Apart from using fixed weights, Fukuda \textit{et al.} \cite{fukuda2017efficient} presented two additional strategies for multi-teacher KD: 
randomly selecting a teacher from an ensemble of teachers for each training batch or iteratively using each teacher from the ensemble to train the student for each batch.
In contrast to using static weights for all training instances, Yuan \textit{et al.} \cite{yuan2021reinforced} presented a reinforcement learning-based KD approach to dynamically assign different weights to teachers during student's training process. 

The above-mentioned teacher-specified methods typically involve two stages: first, pre-train one or multiple teacher models, and then transferring the knowledge to the student via various techniques. However, training a single teacher with a large-scale dataset or multiple teachers could be very time-consuming. Hence, some teacher-free KD approaches (online distillation \cite{furlanello2018born,tian2020rethinking}, mutual learning \cite{zhang2018deep}, or self distillation \cite{ji2021refine,ji2021refine} in previous works) have been proposed to eliminate the need for pre-training cumbersome teachers. Approaches mentioned assume a compact student model deployable in resource-limited environments and aim to enhance its performance via knowledge distillation.

\textbf{Data-available vs. Data-free Distillation} 
Existing KD works assume data availability for training teacher and student. However, data privacy or safety concerns may prevent access to the original dataset.
To address this issue, several data-free KD approaches have been proposed. For instance,
Lopes \textit{et al.} \cite{lopes2017data} utilized the metadata (the activations collected at the teacher's training process) to reconstruct the synthesis data for student training. Similarly, Yin \textit{et al.} \cite{yin2020dreaming} leveraged information from the teacher's batch normalization layers to synthesize images. Chen \textit{et al.} \cite{chen2019data}, and Zhang \textit{et al.} \cite{zhang2021data} exploited the generative adversarial networks (GANs) to regenerate training data where the pre-trained teacher served as the discriminator.  Nayak \textit{et al.} \cite{nayak2019zero} modeled the data distribution in teacher's softmax spaces and proposed a sample extraction approach for data synthesis. 
These data-free methods are useful when dataset access is limited, enabling distillation techniques in sensitive or resource-constrained environments. More intriguingly, some recent studies \cite{ma2021undistillable, jandial2022distilling} suggest that these data-free approaches could result in an unintended server-side consequence of KD, potentially jeopardizing the protection of intellectual property in machine learning. It may create an opening for unauthorized individuals to replicate the functionality of an intellectual property teacher model by mimicking its input and output behaviors as a black box. As the countermeasures to mitigate such risk, an undistillable model called \textit{Nasty Teacher} and an extension of it have been introduced in \cite{ma2021undistillable, jandial2022distilling}, respectively. Although less explored, data privacy and protection of intellectual property hold significant importance and could be another potential research hotspot in KD.

In summary, the metric- and teacher-specific distillation schemes remain the mainstreams among the existing KD works, in which most of them focus on data-available configuration. On the one hand, the metric-specific KD methods are more task-specified and have demonstrated their effectiveness on various applications, whereas they also have some drawbacks such as poor generalization capability on other tasks. Metric-free distillation schemes can partially alleviate this issue via the adversarial learning techniques. But it still suffers from the difficulty in the aspect of model convergence. On the other hand, although training cumbersome models as teachers are very laborious, most of KD works employ the specified teacher during distillation due to the superior generalization ability derived from complex network architecture. However, how to select a proper teacher, especially considering the network capability gap or disparate network architectures between teacher and student, is still very challenging. Besides, data-free distillation has shown great potential in terms of intellectual propetry protection, how to generate synthetic data with high quality and diversity to improve model generalization capability needs further exploration.

\subsection{NMS algorithms}
The NMS algorithm is shown in~\ref{alg:nms}. 
 \begin{algorithm}
 \footnotesize
  \caption{The classical greedy NMS.}
 \SetKwData{Left}{left}\SetKwData{This}{this}\SetKwData{Up}{up}
 \SetKwFunction{Union}{Union}\SetKwFunction{FindCompress}{FindCompress}
 \SetKwInOut{Input}{Input}\SetKwInOut{Output}{Output}
 \Input{$\mathcal{B} = \{b_1, ..,b_N\} $, $\mathcal{S} = \{s_1, ..,s_N\}$, $\tau$ \;
 ~~~~~~~~~~~~~~~$\mathcal{B}$ is the list of initial detection boxes\; 
 ~~~~~~~~~~~~~~~$\mathcal{S}$ contains corresponding detection scores\;
 ~~~~~~~~~~~~~~~$\tau$ is the NMS threshold
 }
 \Begin{
 $\mathcal{D} \leftarrow \{\}$ \;
 \While{$\mathcal{B} \ne$ empty}{
 \DontPrintSemicolon 
 \small{{\Comment*[l]{1) greedy: sorting}}}
 $m \leftarrow$ argmax $\mathcal{S} $\;
 $\mathcal{M} \leftarrow$ $b_m$\;
 \DontPrintSemicolon 
 \small{{\Comment*[l]{2) greedy: finalizing box}}}
 $\mathcal{D} \leftarrow \mathcal{D} \bigcup \mathcal{M }$; $ \mathcal{B} \leftarrow \mathcal{B} -  \mathcal{M}$\; 
 \For{$b_i ~ in ~ \mathcal{B}$} { 
\DontPrintSemicolon 
\small{{\Comment*[l]{3) greedy: duplicate check}}}
  \If{~$iou(\mathcal{M}, b_i) \ge \tau$~}{ 
  $ \mathcal{B} \leftarrow \mathcal{B} -  b_i$;~$ \mathcal{S} \leftarrow \mathcal{S} -  s_i$\;
  }
  }
 }
 \Return{$\mathcal{D}, \mathcal{S}$}
 }
 \label{alg:nms}
 \end{algorithm}\DecMargin{0em}
 
\subsection{When NN compression meets Hardware}
\begin{table}[t!]
\centering
\footnotesize
\caption{Rough energy consumption for various operations in 45 nm fabrication technology with 0.9V supply voltage. Table adopted from~\cite{horowitz20141,vivienne2019efficient}.}
\label{tab:hw-energy-cost-horowitz20141}
\resizebox{\columnwidth}{!}{%
\begin{tabular}{c|c|c|c|c}
\hline
Operations &
  \begin{tabular}[c]{@{}c@{}}Energy \\ Cost (pJ)\end{tabular} &
  \begin{tabular}[c]{@{}c@{}}Relative \\ Energy Cost\end{tabular} &
  \begin{tabular}[c]{@{}c@{}}Area\\ Cost ($\mu$m$^2$)\end{tabular} &
  \begin{tabular}[c]{@{}c@{}}Relative \\ Area Cost\end{tabular} \\ \hline
8-bit Add                                                              & 0.03 & 1$\times$        & 36   & 1$\times$     \\ \hline
32-bit Add                                                             & 0.1  & 3.3$\times$      & 137  & 3.8$\times$   \\ \hline
8-bit Mult                                                             & 0.2  & 6.7$\times$      & 282  & 7.8$\times$   \\ \hline
16-bit FP Add                                                          & 0.4  & 13.3$\times$     & 1360 & 37.8$\times$  \\ \hline
32-bit FP Add                                                          & 0.9  & 30.0$\times$     & 4184 & 116.2$\times$ \\ \hline
16-bit FP Mult                                                       & 1.1  & 36.7$\times$     & 1640 & 45.6$\times$  \\ \hline
32-bit Mult                                                            & 3.1  & 103.3$\times$    & 3495 & 97.1$\times$  \\ \hline
32-bit FP Mult                                                       & 3.7  & 123.3$\times$    & 7700 & 213.9$\times$ \\ \hline
\begin{tabular}[c]{@{}c@{}}32-bit SRAM \\      read (8KB)\end{tabular} & 5    & 166.7$\times$    & --   & --            \\ \hline
\begin{tabular}[c]{@{}c@{}}32-bit SRAM \\      read (1MB)\end{tabular} & 50   & 1,666.7$\times$  & --   & --            \\ \hline
32-bit DRAM   Read                                                     & 640  & 21,333.3$\times$ & --   & --            \\ \hline
\end{tabular}%
}
\end{table}

\begin{table}[!t]
\centering
\footnotesize
\caption{Tunstall decoder and Huffman decoder performance of ResNet-50. (Table adapted from~\cite{chen2021efficient})}
\label{tab:tun_huf_cmp}
\resizebox{\columnwidth}{!}{%
\begin{tabular}{c|c|c|c|c}
\hline
Hardware Decoder &
  \begin{tabular}[c]{@{}c@{}}Resources\\ Utilizations\\ (\# LUT)\end{tabular} &
  \begin{tabular}[c]{@{}c@{}}Decoding \\ Cycles\end{tabular} &
  \begin{tabular}[c]{@{}c@{}}Memory \\ Capacity \\ (MB)\end{tabular} &
  \begin{tabular}[c]{@{}c@{}}Memory Access \\ Consumption \\ (mJ)\end{tabular} \\ \hline
\begin{tabular}[c]{@{}c@{}}7-stage 24 memory banks \\ LO Tunstall decoder\end{tabular} & 6,322 & 8,653,482  & 5.85 & 373.5 \\ \hline
\begin{tabular}[c]{@{}c@{}}1024 entries \\ MO Tunstall decoder\end{tabular}            & 1,601 & 4,565,527  & 6.13 & 130.1 \\ \hline
\begin{tabular}[c]{@{}c@{}}256 entries \\ Huffman decoder\end{tabular}                 & 5,021 & 28,460,750 & 5.59 & 139.3 \\ \hline
\end{tabular}%
}
\end{table}
Table~\ref{tab:tun_huf_cmp}~\cite{chen2021efficient} summaries the detailed theoretical hardware performance of the 7-stage 24 memory banks Logic-Oriented Tunstall decoder and the 1024 entries Memory-Oriented Tunstall decoder with the 256 entries Huffman decoder. The Huffman decoder decodes one weight per clock cycle. 
All implementations are synthesized and mapped to FPGA.
As summarised, the 256 entries Huffman hardware decoder consumes $3.14\times$ more hardware resources and costs $6.23\times$ more time to decode the same amount of weights.

\begin{figure}[t!]
    \centering
    \small
    \includegraphics[width=0.8\columnwidth]{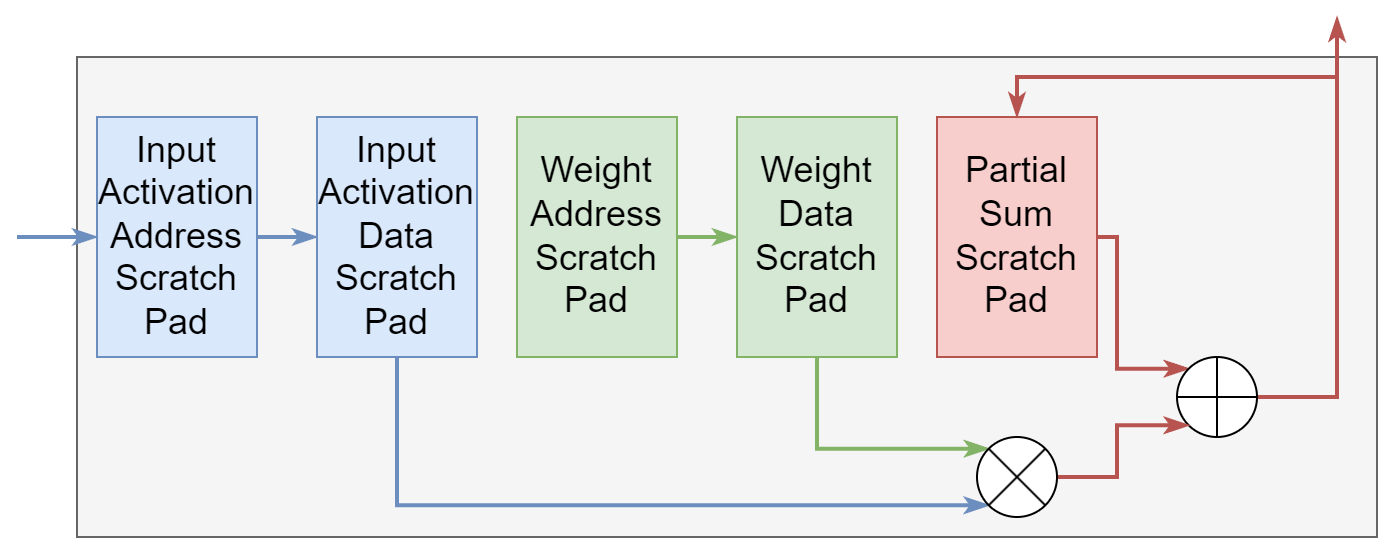}
    \caption{Simplified Eyeriss v2 PE Architecture. The PE is designed for CSC data format; the address scratchpad for both input activation and weight is used to store the address vector in the CSC compressed data, while the data scratchpad stores the data and count vectors. Figure adapted and simplified from~\cite{chen2019eyeriss}}
    \vspace{-0.1cm}
    \label{fig:eyeriss_v2_pe}
\end{figure}

\begin{figure}[!t]
    \centering
    \captionsetup[subfloat]{labelfont=footnotesize,textfont=footnotesize}
    \subfloat[AIM in SNAP]{
        \label{fig:hw-zhang2019snap}
        \centering
        \includegraphics[width=0.35\columnwidth]{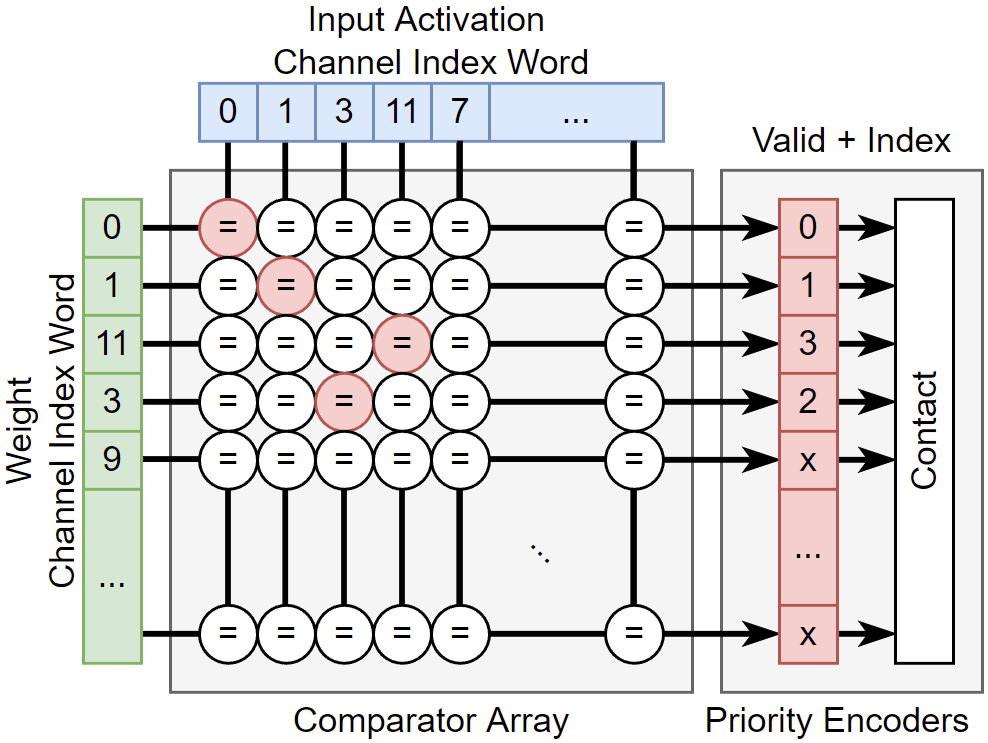}
    }
    \subfloat[Neuron selector module in Cambricon-S]{
        \label{fig:hw-zhou2018cambricon}
        \centering
        \includegraphics[width=0.6\columnwidth]{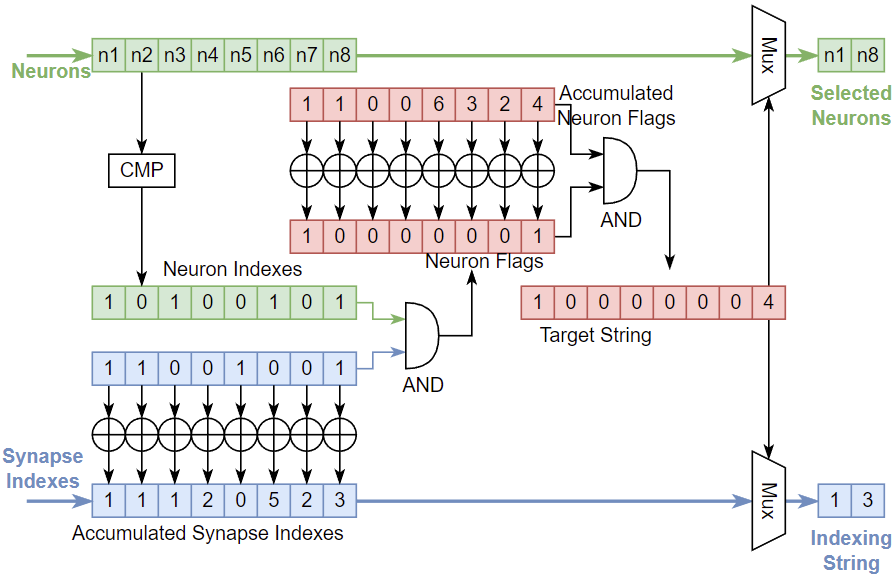}
    }
    \caption{a) The decoding modules in SNAP, where the comparator array is used to find the matched valid input activation and corresponding weight. Figure adopted and from~\cite{zhang2019snap}. b) The Neuron selector module in Cambrion-S, where AND gates decode the input activation and weight. Figure adopted from~\cite{zhou2018cambricon}.}
    \label{fig:zhang2019snap-zhou2018cambricon}
\end{figure}

\begin{table}[!t]
\centering
\caption{Execution time and speedup of clustering 1000 and 8000 bounding boxes in different methods compared with GreedyNMS} 
\label{tab:nms-hw-perf-cmp} 
\resizebox{\columnwidth}{!}{%
\begin{threeparttable} 
\begin{tabular}{c|c|c|c|c}
\hline
\multirow{2}{*}{Method} & \multicolumn{2}{c|}{NMS Execution Time ($\mu$s)}  & \multicolumn{2}{c}{Speedup} \\ \cline{2-5} 
                                 & n=1,000  & n=8,000 & n=1,000  & n=8,000 \\ \hline
GreedyNMS\tnote{a}          & 35,000                   & 512,000                 & 1                        & 1                       \\ \hline
PSRR-MaxpoolNMS~\cite{psrrmaxpool}\tnote{a}        & 18,000                   & 89,000      & 1.944                    & 5.753                   \\ \hline
GPU-NMS V2~\cite{gpunms2}\tnote{b}             & 324                      & ---                      & 108.025                  & ---                      \\ \hline
Shi \etal~\cite{powernms}\tnote{c}              & 12.79                    & 102.32                  & 2,736.513                & 5,003.909               \\ \hline
ShapoolNMS~\cite{shapoolnms}\tnote{d}              & 5.69                   & 11.99                & 6,154.495                & 42,713.830              \\ \hline
\end{tabular}

\begin{tablenotes}
\footnotesize
\item[a] The data is reported in~\cite{psrrmaxpool}.
\item[b] Executed on GeForce GTX 1060. The execution time of $n=8,000$ is not reported.
\item[c] The operation frequency is 400MHz. The speedup for $n=8,000$ is projected from the reported execution time of $n=1,000$ and $n=10,000$ in a linear trend.
\item[d] The operation frequency is 400MHz. The shown values are the execution time of a 4-way ShapoolNMS to process the 1-stage or the second stage of the 2-stage detectors.
\end{tablenotes}
\end{threeparttable}
}
\end{table}

\begin{figure}[t!]
    \centering
    \includegraphics[width=0.9\columnwidth]{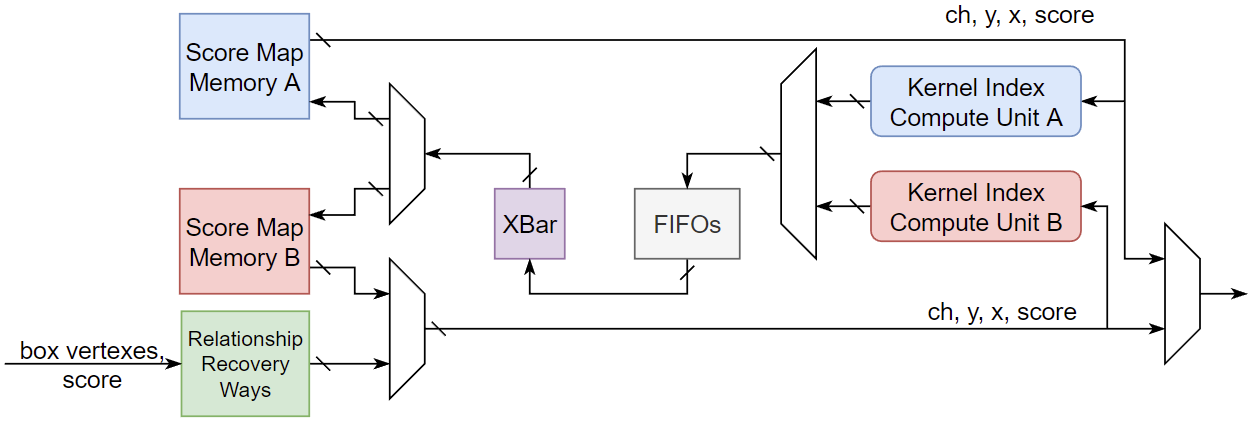}
    \caption{System architecture which is illustrated with 4 RRWs ($N=4$) and 8 memory banks in SMMs ($M=8$). The depth of the FIFOs is 4.}
    \label{fig:hw-shapoolnms-top}
\end{figure}

\bibliographystyle{IEEEtran}
\bibliography{IEEEabrv, reference.bib}